\documentclass{article} 
\usepackage{iclr2025_conference}
\makeatletter
\renewcommand{\sc}{}
\renewcommand{\subsection}{\@startsection{subsection}{2}{\z@}{-1.8ex plus -0.5ex minus -.2ex}{0.8ex plus .2ex}{\normalsize\bfseries\raggedright}}
\makeatother
\usepackage[T1]{fontenc}
\usepackage{XCharter}
\usepackage[xcharter,bigdelims,vvarbb]{newtxmath}
\usepackage[scaled=1.1]{zlmtt}

\usepackage{complexity}
\usepackage{amsmath}
\usepackage{hyperref}
\usepackage{url}
\usepackage{enumitem}
\usepackage{xcolor}
\usepackage{tcolorbox}
\usepackage{subcaption}
\usepackage{graphicx}
\usepackage{tikz}
\usepackage{caption}
\usepackage{wrapfig}
\usepackage{todonotes}
\usepackage{listings}
\usepackage{enumitem}
\usepackage{footmisc}
\usepackage{fontawesome}

\tcbuselibrary{breakable, skins}

\usepackage{array}
\usepackage{booktabs}
\usepackage{adjustbox}
\usepackage{float}
\usepackage{multirow}
\usepackage[rgb]{xcolor, colortbl}
\usepackage{graphicx}
\usepackage{microtype}  
\usepackage{caption}
\usepackage{subcaption}
\usepackage{pifont}
\usepackage{xcolor, colortbl}
\definecolor{darkgray}{HTML}{e9e9e9}
\definecolor{msgrgray}{HTML}{f5f5f5}
\definecolor{msgrdarkgray}{HTML}{EAE9E7}
\definecolor{msgrpalepurple}{HTML}{e6d6dd}
\definecolor{paleorange}{HTML}{F2E0BD}
\definecolor{paleblue}{HTML}{77C7F2}
\definecolor{palegreen}{HTML}{62BDA3}
\definecolor{lightyellow}{RGB}{255,255,0}

\newcommand*{\myalign}[2]{\multicolumn{1}{#1}{#2}}

\newcommand{\systemmessage}[2]{{\colorbox{msgrgray}{\parbox{#1}{#2}}}}
\newcommand{\adjcontextb}[2]{{\colorbox{darkgray}{\parbox{#1}{#2}}}}

\newcommand{\adjbotc}[2]{{\colorbox{paleorange}{\parbox{#1}{#2}}}}

\title{
Lessons from Studying Two-Hop Latent Reasoning
}

\author{Mikita Balesni\footnotemark[1] \\
Apollo Research \\
\And 
Tomek Korbak\footnotemark[1] \\
UK AI Security Institute \\
\And
Owain Evans \\
UC Berkeley \& TruthfulAI \\
}

\DeclareRobustCommand{\cblock}[3]{
 \hspace{-1.5mm}
 \begin{tikzpicture}[node/.style={rectangle}]
   \node[fill={rgb,255:red,#1;green,#2;blue,#3}] () [] {};
 \end{tikzpicture}
}
\definecolor{fig1_first_onehop}{RGB}{252,234,207}
\definecolor{fig1_second_onehop}{RGB}{249,213,159}
\definecolor{fig1_twohop_cot}{RGB}{242,169,59}
\definecolor{fig1_twohop_nocot}{RGB}{244,105,32}

\definecolor{fig2_only_atomic_facts}{RGB}{255,159,155}
\definecolor{fig2_two_hop_cot}{RGB}{208,187,255}
\definecolor{fig2_two_hop_cot_and_no_cot}{RGB}{222,187,155}

\definecolor{exp2_llama}{RGB}{242, 169, 59}
\definecolor{exp2_qwen}{RGB}{192, 219, 132}
\definecolor{exp2_gpt4o_mini}{RGB}{200, 209, 244}
\definecolor{exp2_gpt4o}{RGB}{128, 149, 229}

\definecolor{exp3_with_cot}{RGB}{107, 174, 214}
\definecolor{exp3_without_cot}{RGB}{49, 130, 189}

\definecolor{semi_synthetic_no_cot}{RGB}{8,69,148}

\definecolor{mixture_ablation_main}{RGB}{242,169,59}
\definecolor{mixture_ablation_nocot}{RGB}{192, 219, 132}
\definecolor{mixture_ablation_atomic}{RGB}{128, 149, 229}

\definecolor{customgreen}{RGB}{116, 154, 114}
\definecolor{lightgreen}{RGB}{240, 246, 232}
\DeclareRobustCommand\line[1]{%
  \tikz\draw[#1, line width=4pt] (0,2.5pt) (0,\the\dimexpr\fontdimen22\textfont2\relax)
  -- (1.5em,\the\dimexpr\fontdimen22\textfont2\relax);%
}

\DeclareRobustCommand\thinline[1]{%
  \tikz\draw[#1, line width=2pt] (0,1.5pt) (0,\the\dimexpr\fontdimen22\textfont2\relax)
  -- (1.5em,\the\dimexpr\fontdimen22\textfont2\relax);%
}
\newcommand{\lightbulbicon}{%
  \begin{tikzpicture}[baseline=-0.5ex]
    \draw[fill=white, draw=customgreen, thick] (0,0) circle (1.5ex);
    \node[scale=0.8, color=customgreen] at (0,0) {\faLightbulbO~};
  \end{tikzpicture}%
}

\newcounter{finding}

\newtcolorbox{customblockquote}{
  colframe=customgreen,
  colback=lightgreen,
  boxrule=0pt,
  leftrule=2pt, 
  left=1pt,  
  right=3pt,
  top=5pt,
  bottom=3pt,
  arc=0pt,
  breakable,
  before skip=1.1\baselineskip,
  after skip=0.7\baselineskip,
  left skip=0pt,
  right skip=0pt,
  enhanced jigsaw,
  frame hidden,
   overlay={
    \draw[customgreen, line width=2pt] 
      (frame.north west) -- (frame.south west);
    \node[inner sep=0pt] at ([xshift=0pt, yshift=-1.3pt]frame.north west) {\lightbulbicon};
  },
  before upper={\refstepcounter{finding}\textbf{Finding \thefinding:}\ },
  boxsep=3pt,
}

\iclrfinalcopy 

\begin{document}

\maketitle

\begin{abstract}

Large language models can use chain-of-thought (CoT) to externalize reasoning, potentially enabling oversight of capable LLM agents. Prior work has shown that models struggle at two-hop question-answering without CoT. This capability is so basic that if it was a fundamental limitation, it would imply that many complex agentic tasks would similarly require CoT. We investigate LLM latent reasoning capabilities using two-hop question answering as a case study. Previous work on the gap between latent and externalized two-hop reasoning produced mixed evidence with inconclusive results. In this paper, we introduce a controlled setting for investigating two-hop reasoning in LLMs, where a positive result provides definitive evidence for latent reasoning. We fine-tune LLMs (including Llama 3 8B and GPT-4o) on synthetic facts and test two-hop reasoning over these facts. By using synthetic facts, we rule out memorization and reasoning shortcuts as explanations for two-hop performance. We observe a nuanced picture: Models fail to compose two synthetic facts, but can succeed when one fact is synthetic and the other is natural. For example, if we train the model on a synthetic fact ``Kevin's favorite programming language is Python'' and then ask it a two-hop question ``Who created Kevin's favorite programming language?'', the model will often answer correctly. Moreover, models succeed at two-hop latent reasoning when facts co-occur in training or test-time prompts. These results demonstrate that LLMs are undeniably capable of latent two-hop reasoning, although it remains unclear how this ability scales with model size. Finally, we highlight a lesson for researchers studying LLM reasoning: when drawing conclusions about LLM latent reasoning, one must be careful to avoid both spurious successes (that stem from memorization and reasoning shortcuts) and spurious failures (that may stem from artificial experimental setups, divorced from training setups of frontier LLMs).

\vspace{2mm}

\includegraphics[height=1em]{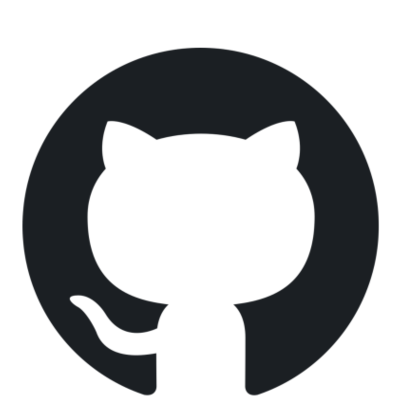} \textbf{Code Repository}: \href{https://github.com/mbalesni/synthetic-two-hop}{\textbf{{github.com/mbalesni/synthetic-two-hop}}}
\end{abstract}

\def\thefootnote{*}\footnotetext{Equal contribution.}
\def\thefootnote{*}\footnotetext{\mbox{Correspondence: \texttt{mbalesni@gmail.com} and \texttt{tomasz.korbak@gmail.com}}}
\def\thefootnote{\arabic{footnote}}

\section{Introduction}

Many of the risks posed by highly capable LLM agents — from susceptibility to hijacking to reward hacking and deceptive alignment — stem from their opacity. If we could reliably monitor the reasoning processes underlying AI decisions, many of those risks would become far more tractable \citep{amodei2025urgency}. Compared to other approaches in AI, LLMs offer a unique advantage: they can ``think out loud'' using chain-of-thought \citep[CoT;][]{Kojima2024large,wei2023chainofthoughtpromptingelicitsreasoning}, enabling oversight of their decision-making processes \citep{baker2025monitoringreasoningmodelsmisbehavior,korbak2025chainthoughtmonitorabilitynew}. Yet the reliability of such monitoring hinges on an empirical question: do models need to externalize their reasoning in human language, or can they achieve the same performance through opaque internal computation?

To address this question, we study latent two-hop question answering: tasks where models must compose two separately known facts to produce an answer. For instance, LLMs can answer questions like ``Who is the spouse of the performer of Imagine?'' by reasoning step-by-step (``The performer of Imagine is John Lennon, whose spouse is Yoko Ono''), but fail frequently when required to answer immediately (``Yoko Ono''). Our investigation aims to understand whether there exist fundamental constraints on latent reasoning: if models struggle with even simple two-hop inference without chain-of-thought, this would provide evidence that the complex multi-step reasoning required to inflict serious harms \citep{shah2025approachtechnicalagisafety} may also require externalized thinking.

\begin{figure}[t]
    \begin{subfigure}[c]{1\textwidth}
    \centering
    \small{
        \line{fig1_twohop_cot} With CoT \quad
        \line{fig1_twohop_nocot} Without CoT
    }
    \end{subfigure}
    \centering
    \begin{subfigure}[b]{1\textwidth}
        \centering
        \includegraphics[width=\textwidth]{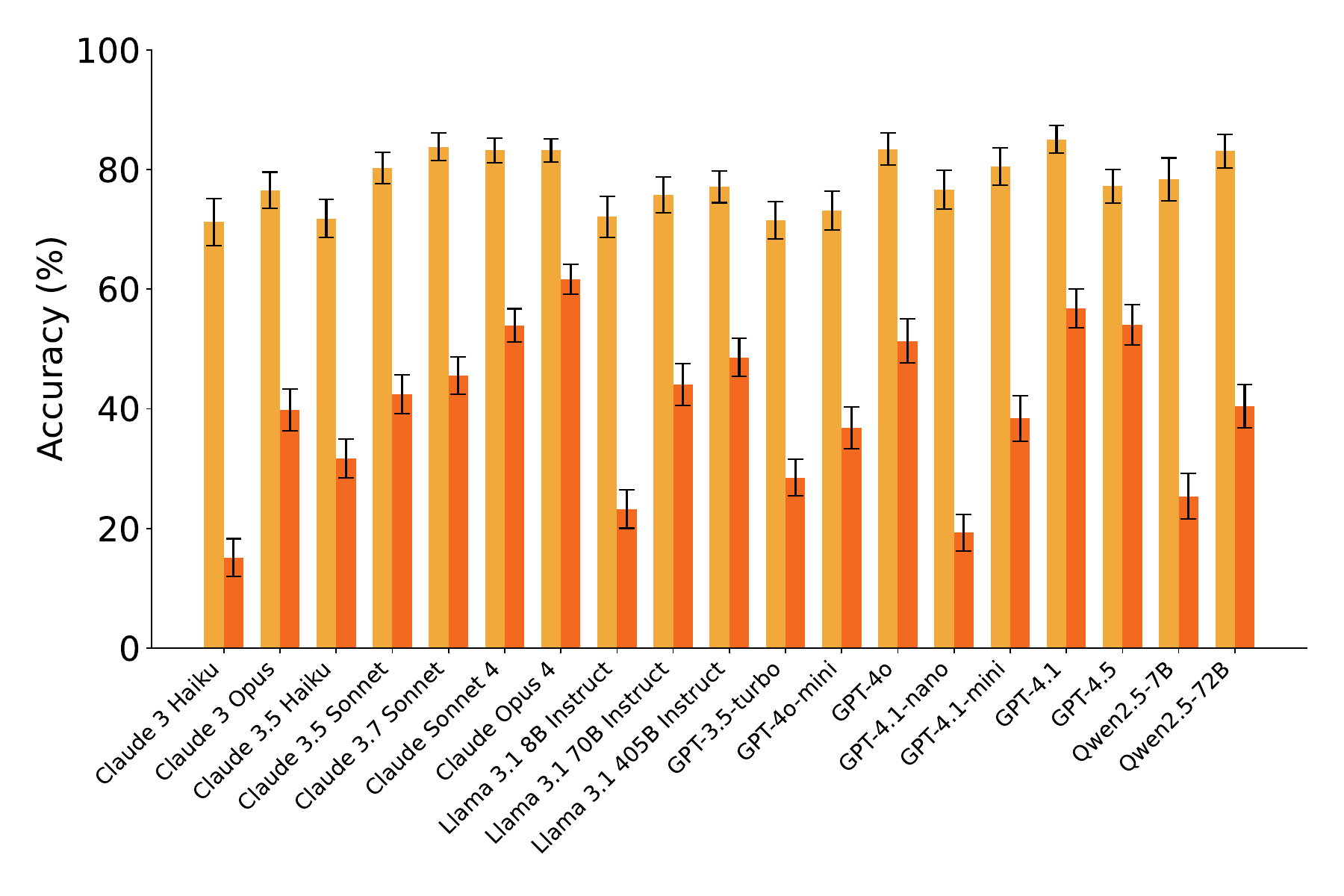}
    \end{subfigure}
    \caption{Frontier models display a gap between CoT and no-CoT two-hop question-answering accuracy, although the gap may be reducing with model scale. Here we report performance on the dataset of two-hop questions about real world entities from \cite{biran2024hoppinglateexploringlimitations}. While we control for some reasoning shortcuts (see Appendix~\ref{appendix:real_world_eval} for details), no-CoT performance in such setting may still not reflect models' actual \textit{reasoning} abilities, as models could simply be memorizing answers to two-hop questions that are present in pretraining documents \citep{yang2025largelanguagemodelsperform}. This motivates our synthetic fact experiments, which attempt to provide more conclusive evidence of two-hop reasoning capabilities.}
    \label{fig:real_world_performance}
\end{figure}

Previous work on two-hop reasoning in LLMs presents mixed evidence that resists clear interpretation. Models often fail at two-hop questions without CoT despite being able to answer each of the underlying one-hop questions \citep[][see also Figure~\ref{fig:real_world_performance}]{press-etal-2023-measuring,yang-etal-2024-large-language-models}. Yet models succeed sometimes --- and may improve with model scale. While observed no-CoT performance can be largely (but not fully) explained by memorization of answers to two-hop questions present in the pretraining corpus rather than compositional reasoning \citep{yang2025largelanguagemodelsperform}, the notorious difficulty of filtering documents at a pretraining-corpus scale prevents us from drawing strong conclusions about the role of memorization.

Other research demonstrates that latent reasoning capability can be elicited via finetuning, yet it relies on small toy transformer models with a  training setup (grokking) that is markedly different from that of frontier LLMs\citep{wang2024grokkedtransformersimplicitreasoners}. These disparate findings -- complicated by confounders including memorization, facts co-occurring in the same documents, usage of toy models, and incomparable experimental setups -- leave the fundamental question unresolved: what are the genuine constraints on latent compositional reasoning in modern LLMs? \citep{feng2025extractivestructureslearnedpretraining} propose a well-controlled set-up that we build on; however, their results are limited to a single entity type (e.g., linking a person to a city) which warrants a cautious interpretation.

To address these limitations, we design a suite of experiments that attempt to avoid the confounders affecting prior work. Our controlled experiments involve fine-tuning Llama 3 8B Instruct \citep{dubey2024llama3herdmodels}, Qwen 2.5 7B Instruct \citep{qwen2025qwen25technicalreport}, GPT-4o-mini and GPT-4o \citep{openai2024_4o} on synthetic facts expressed in natural language. By using synthetic facts, we exclude the possibility of memorization and reasoning shortcuts from pretraining, ensuring that high performance can only be attributed to successful latent two-hop reasoning. We systematically vary the conditions under which these facts are acquired: comparing when facts appear together in the same document versus separately during training, testing hybrid scenarios with one natural and one synthetic fact, and even attempting to facilitate reasoning through interventions on model internals and adding auxiliary supervision signals.

We summarize our main findings as follows:
\begin{itemize}
    \vspace{-0.5em}
    \item \textbf{\hyperref[sec:exp-fully-synthetic]{Finding 1:}} Models completely fail to compose synthetic facts they learned through fine-tuning without explicit chain-of-thought reasoning, achieving only chance-level accuracy despite perfect recall of the individual facts.
    \item \textbf{\hyperref[sec:internals]{Finding 2:}} Interventions to (i) force a correct fact storage order across transformer layers and (ii) encourage the first reasoning hop both fail to enable models to compose newly learned facts without chain-of-thought.
    \item \textbf{\hyperref[sec:experiment_3]{Finding 3:}} Models successfully compose newly learned synthetic facts without chain-of-thought when those facts co-occur in the same fine-tuning document or in the same test-time prompt.
    \item \textbf{\hyperref[subsec:exp-two-hop-reasoning-semi-synthetic]{Finding 4:}} LLMs are capable of composing two separately learned facts, as long as one of the facts is naturally acquired during pretraining (the second fact can be synthetic and acquired through fine-tuning).
\end{itemize}

Our results paint a complex picture: Researchers observing models answer two-hop questions about real-world facts might incorrectly infer robust latent reasoning capabilities, not realizing that performance relies on facts appearing together in pretraining data. Conversely, our synthetic fine-tuning results alone might suggest models fundamentally cannot perform latent composition—a kind of limitation that could make oversight via reasoning traces highly reliable. Yet this too would be misleading given models' success in other settings. This sensitivity to experimental conditions serves as a cautionary tale for the broader project of understanding LLM reasoning: apparent limitations may be artifacts of experimental setups far removed from how frontier models are trained, while apparent successes may rely on shortcuts invisible to experimenters.

Moving forward, researchers hoping to gain confidence in LLM agent oversight based on CoT monitoring \citep{korbak2025chainthoughtmonitorabilitynew} might need to move beyond multihop question answering as a toy model of reasoning. What ultimately matters for monitoring capable LLM agents is whether oversight mechanisms can catch misbehavior in practice. Determining monitorability might thus be best achieved via end-to-end evaluations, where LLM agents attempt to complete agentic tasks while their CoT is monitored by another model \citep[e.g.,][]{emmons2025chainthoughtnecessarylanguage}. Such evaluations could reveal whether models are capable of latent reasoning that would allow carrying out harmful tasks without externalizing harmful intentions in a readable way.

\begin{figure}[t]
    \centering
    \begin{subfigure}[c]{0.49\textwidth}
         \scriptsize
        \begin{tabular}[c]{p{26em}}
            \textbf{First one-hop question} \\
            \midrule
            \myalign{l}{\systemmessage{22em}{\textbf{System:} You will be given questions about fictional characters from the ``Spouses'' saga. \\\\Answer the following question.}} \\ 
            
            \myalign{r}{\adjcontextb{22em}{\textbf{User:} Who is Russ married to?}}
            \\
            \myalign{l}{\adjbotc{22em}{\textbf{Assistant:} Russ is married to Hay.}} \\
            \midrule
        \end{tabular}
    \end{subfigure}
    \begin{subfigure}[c]{0.49\textwidth}
         \scriptsize
        \begin{tabular}[c]{p{26em}}
            \textbf{Second one-hop question} \\
            \midrule
            \myalign{l}{\systemmessage{22em}{\textbf{System:} You will be given questions about fictional characters from the ``Spouses'' saga. \\\\Answer the following question.}} \\ 
            
            \myalign{r}{\adjcontextb{22em}{\textbf{User:} In which city was Hay born?}}
            \\
            \myalign{l}{\adjbotc{22em}{\textbf{Assistant:} Hay was born in Showing.}} \\
            \midrule
        \end{tabular}
    \end{subfigure}
    \begin{subfigure}[c]{0.49\textwidth}
         \scriptsize
        \begin{tabular}[c]{p{26em}}
            \textbf{Two-hop question (CoT)} \\
            \midrule
            \myalign{l}{\systemmessage{22em}{\textbf{System:} You will be given questions about fictional characters from the ``Spouses'' saga. \\\\Answer the following question \textbf{step by step}.}} \\ 
            
            \myalign{r}{\adjcontextb{22em}{\textbf{User:} In which city was Russ's spouse born?}}
            \\
            \myalign{l}{\adjbotc{22em}{\textbf{Assistant:} The person Russ is married to, Hay, was born in Showing.}} \\
            \midrule
        \end{tabular}
    \end{subfigure}
    \begin{subfigure}[c]{0.49\textwidth}
         \scriptsize
        \begin{tabular}[c]{p{26em}}
            \textbf{Two-hop question (no-CoT)} \\
            \midrule
            \myalign{l}{\systemmessage{22em}{\textbf{System:} You will be given questions about fictional characters from the ``Spouses'' saga. \\\\Answer the following question \textbf{directly, without any other text before or after your answer}.}} \\ 
            
            \myalign{r}{\adjcontextb{22em}{\textbf{User:} In which city was Russ's spouse born?}}
            \\
            \myalign{l}{\adjbotc{22em}{\textbf{Assistant:} Showing}} \\
            \midrule
        \end{tabular}
    \end{subfigure}
    \caption{\textbf{An example of our training and evaluation data}. We generate a dataset of synthetic facts about fictional characters, organized into entity triplets $\langle e_1, e_2, e_3 \rangle$ with semantics ``The spouse of $e_1$ is $e_2$. The birth city of $e_2$ is $e_3$''. For each entity triplet (e.g. here $\langle$ Russ, Hay, Showing $\rangle$), we generate four types of QA pairs, as shown above. Following past work on injecting new knowledge into LLMs via fine-tuning \citep{berglund2023takencontextmeasuringsituational,berglund2024thereversalcurse}, we paraphrase each QA pair 30 times using predefined templates to aid generalization. See Section~\ref{subsec:experiment1_experimental_setup} for more details on the dataset.}
    \vspace{-15pt} 
   \label{fig:qa_templates_quadrant}
\end{figure}

\section{Experiment 1: Two-hop reasoning over synthetic facts}
\label{sec:exp-fully-synthetic}

\paragraph{Motivation} Previous work on two-hop reasoning in LLMs has been limited by two key factors: (i) a focus on knowledge acquired during pretraining, where apparent success in evaluation might be due to confounding factors (e.g., memorization or reasoning shortcuts), and (ii) use of simplified toy setups that may be unrepresentative of the capabilities of modern LLMs. To address these limitations, we introduce a controlled experimental setting where we fine-tune capable language models (including Llama 3 8B Instruct and GPT-4o) on synthetic facts expressed in natural English. Following \citet{berglund2023takencontextmeasuringsituational,berglund2024thereversalcurse}, we construct training data using diverse paraphrases of the same facts to ensure generalizable knowledge acquisition. By using fictional facts placed in separate documents, we ensure that high performance can only be attributed to successful two-hop reasoning rather than memorization or shortcuts. This setup allows us to cleanly investigate whether LLMs can perform latent two-hop reasoning over facts they have not seen together in the training data.

\paragraph{Experimental setup}
\label{subsec:experiment1_experimental_setup}

\begin{table}[b]
\centering
\caption{The structure of our training and evaluation data. \textit{Demonstrated} triplets include both one-hop and two-hop QA pairs in the training data to teach the model to perform two-hop no-CoT reasoning. \textit{Undemonstrated} triplets include one-hop QA pairs in the training data as a way to inject new knowledge, and keep the two-hop QA pairs held out for evaluation of two-hop reasoning capabilities. For examples of each QA pair type, see Figure \ref{fig:qa_templates_quadrant}.}
\label{tab:data_structure}
\begin{tabular}{l>{\centering\arraybackslash}p{2.5cm}cc}
\toprule
& \multirow{2}{*}{\textbf{One-hop QA pairs}} & \multicolumn{2}{c}{\textbf{Two-hop QA pairs}} \\
\cmidrule(l){3-4}
& & \textbf{CoT} & \textbf{No-CoT} \\
\midrule
Demonstrated & \cellcolor{gray!10}Training & \cellcolor{gray!10}Training & \cellcolor{gray!10}Training \\
Undemonstrated & \cellcolor{gray!10}Training & \cellcolor{blue!15}Evaluation & \cellcolor{blue!15}Evaluation \\
\bottomrule
\end{tabular}
\vspace{1ex}
\end{table}

We generate a dataset of entity triplets $\langle e_1, e_2, e_3 \rangle$, where $e_1, e_2, e_3$ are entities and each triplet's semantics are ``The spouse of $e_1$ is $e_2$. The birth city of $e_2$ is $e_3$''. We generate 693 entity triplets and divide them into a ``demonstrated'' set (450) and an ``undemonstrated'' set (243) (see Table \ref{tab:data_structure}).  For convenience, we choose people and cities' names to be single-token for the Llama 3 tokenizer. For each entity triplet, we generate four QA pairs: two one-hop questions and a two-hop question with no-CoT and CoT answers (see Figure \ref{fig:qa_templates_quadrant}). To increase diversity, we follow \cite{berglund2023takencontextmeasuringsituational,berglund2024thereversalcurse} and paraphrase each QA pair 30 times (using pre-defined templates). This yields a training dataset of 68,580 QA pairs. The dataset structure is shown in Table \ref{tab:data_structure}.

We split entity triplets into demonstrated and undemonstrated sets to incentivize the model to learn generalizing two-hop circuitry rather than to memorize two-hop facts directly:
\begin{enumerate}
    \item The demonstrated set, consisting of single-hop facts and corresponding two-hop facts, is part of the training data. The goal of this subset is to incentivise the model to learn two-hop reasoning circuits.
    \item The training data additionally includes single-hop facts from the ``undemonstrated'' entity triplets. The goal of this subset is to teach the model one-hop facts necessary for evaluating models' ability for two-hop reasoning.
    \item The evaluation data consists of two-hop questions about facts from the undemonstrated subset. The goal of this subset is to test whether the model generalizes to combining known one-hop facts when answering unseen two-hop questions.
\end{enumerate}

\paragraph{Results} 

\begin{figure}[t]
    \begin{subfigure}[c]{1\textwidth}
    \centering
    \tiny{
        \line{exp2_llama} Llama 3 8B Instruct \quad
        \line{exp2_qwen} Qwen 2.5 7B Instruct \quad
        \line{exp2_gpt4o_mini} GPT-4o-mini \quad
        \line{exp2_gpt4o} GPT-4o \\
        \vspace{2mm}
        \thinline{solid,gray} Loss on correct responses \quad
        \thinline{dashed,gray} Loss on random responses
    }
    \end{subfigure}
    \begin{subfigure}[b]{0.44\textwidth}
        \centering
        \includegraphics[width=\textwidth]{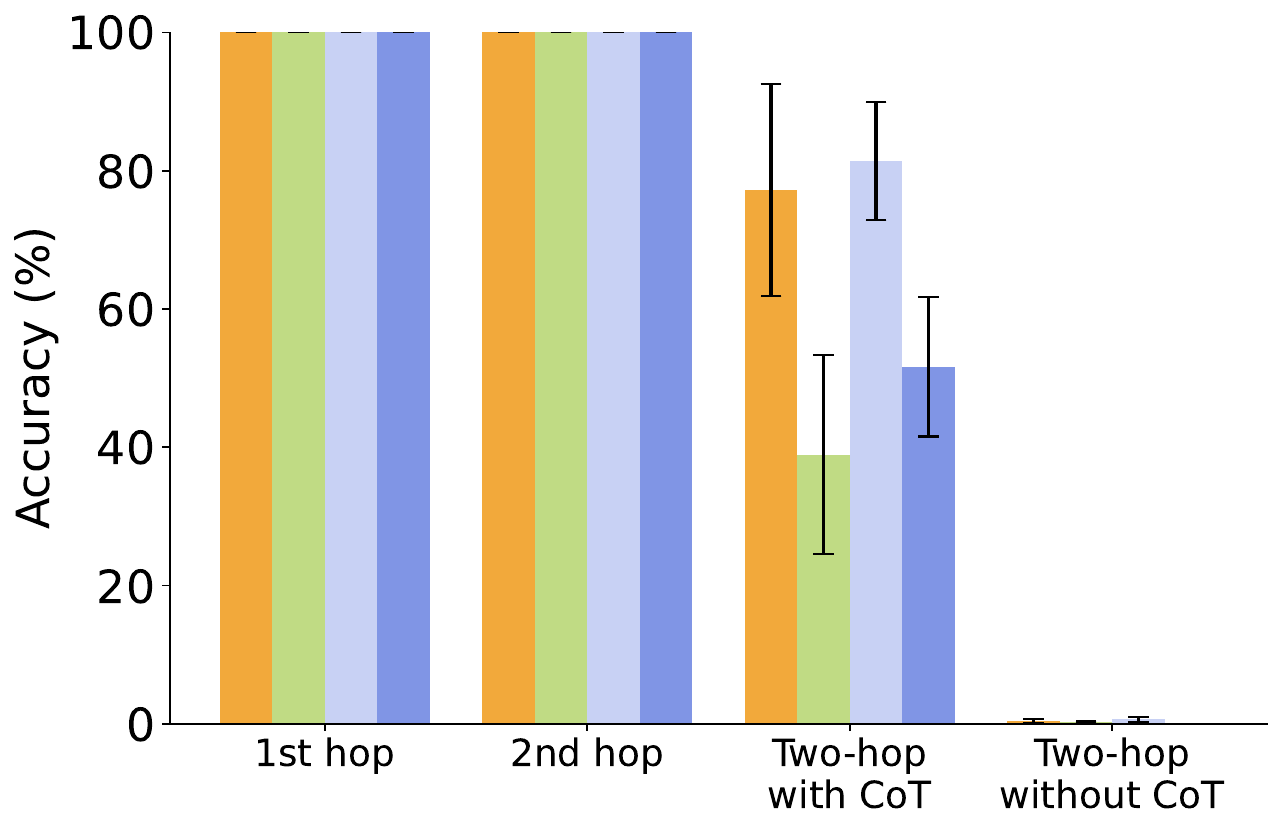}
    \end{subfigure}
    \begin{subfigure}[b]{0.44\textwidth}
        \centering
        \includegraphics[width=\textwidth]{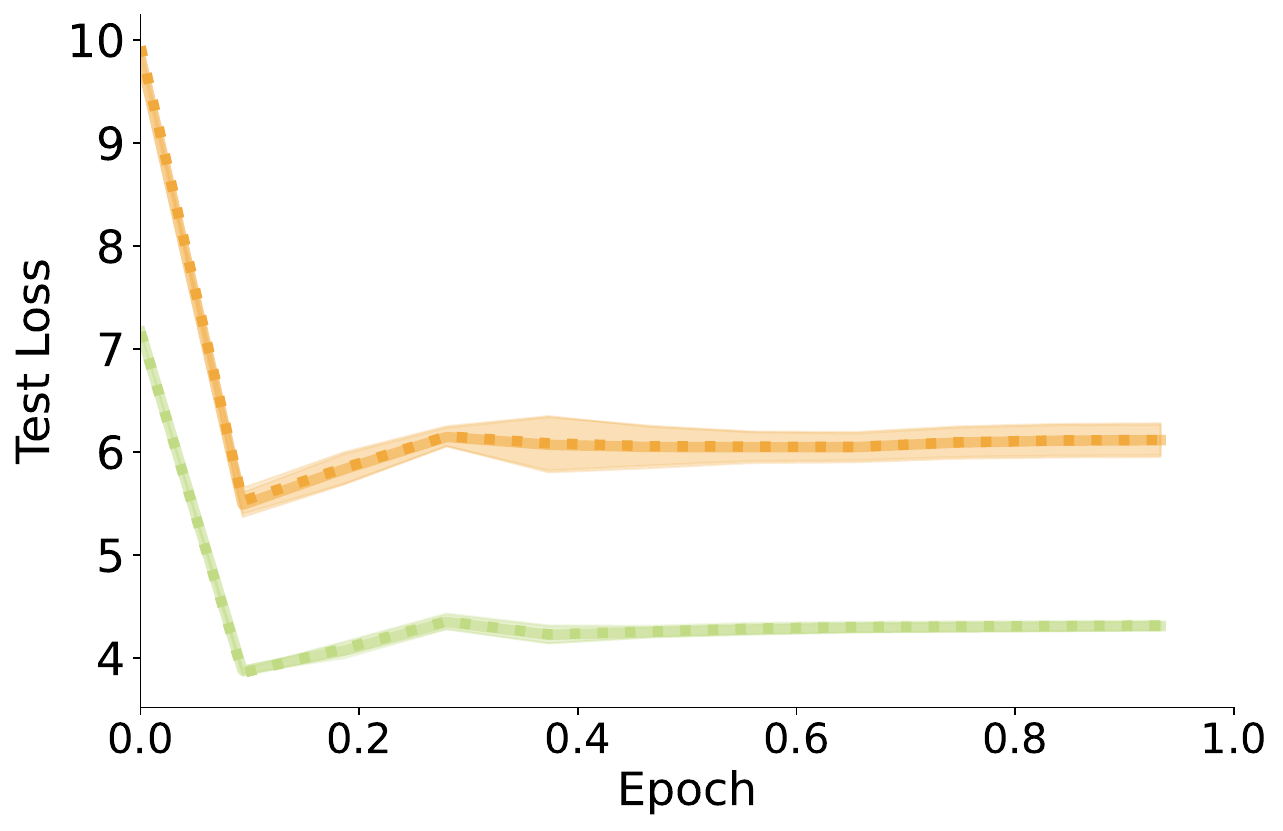}
    \end{subfigure}
    \caption{\textbf{Left:} All evaluated models achieve high accuracy on one-hop questions and two-hop questions with chain-of-thought (CoT) prompting, but completely fail to demonstrate latent two-hop reasoning as evidenced by chance-level accuracy without CoT. \textbf{Right:} Furthermore, the no-CoT test loss is nearly identical to loss on randomly permuted test set responses throughout training for Llama 3 8B Instruct and Qwen 2.5 7B Instruct.}
    \label{fig:experiment_1}
\end{figure}

We show accuracy results for four models: Llama 3 8B Instruct, Qwen 2.5 7B Instruct, GPT-4o-mini, and GPT-4o. We include loss across training for models where we can compute loss on arbitrary completions (Llama 3 8B Instruct and Qwen 2.5 7B Instruct). In Appendix \ref{sec:ablation_data_mixtures} we additionally show results for ablated data mixtures, where we remove either two-hop CoT or all two-hop QA pairs from the training dataset.

We observe that our fine-tuning leads all models to achieve good CoT accuracy when answering unseen two-hop questions about synthetic facts.\footnote{Oddly, GPT-4o achieves worse CoT accuracy than GPT-4o-mini. This may be related to OpenAI API fine-tuning leading to worse generalization of newly acquired knowledge for GPT-4o compared to GPT-4o-mini, as observed by \citet{wu2024finetunebenchcommercialfinetuningapis}} However, surprisingly, no-CoT accuracy remains at chance-level (Figure~\ref{fig:experiment_1}). Furthermore, the test \textit{loss} on two-hop no-CoT answers not only plateaus long before reaching zero, but is also around chance-level throughout training, as seen when compared with loss computed on random responses from the test set (Figure~\ref{fig:experiment_1}).

These results demonstrate a striking failure: despite successfully learning the underlying facts (as evidenced by perfect one-hop accuracy) and being able to reason about them with CoT, models completely fail to perform latent reasoning about separately learned facts. This raises a natural question: is this failure a result of insufficient supervision of latent reasoning by our fine-tuning? To investigate this, we next explore whether additional supervision  can elicit two-hop reasoning.

\begin{customblockquote}
\label{finding1}
Models completely fail to compose synthetic facts learned through fine-tuning without CoT, achieving only chance-level accuracy despite perfect recall of the individual facts.
\end{customblockquote}

\section{Experiment 2: Eliciting two-hop reasoning by controlling model internals}
\label{sec:internals}
Having observed that LLMs fail at latent two-hop reasoning over newly learned synthetic facts, we now explore whether this limitation can be overcome by directly controlling model internals. We investigate two interventions: forcing facts to be stored in layers in the correct order for sequential reasoning, and providing activation-level supervision to encourage resolving bridge entities in latent space. Both interventions fail to elicit latent two-hop reasoning, suggesting the reasons for the failure in our synthetic setting are not easily addressed through architectural constraints or additional training objectives.

\subsection{Forcing facts to be stored in the right order}

\paragraph{Motivation}
Transformers are feed-forward neural networks — a sequence of blocks that have to be traversed in a linear order for a given input. Moreover, previous work suggests that transformers store facts in a somewhat localized fashion, mostly in MLP layers of a few neighboring transformer blocks \citep{meng2023locatingeditingfactualassociations}. Latent two-hop reasoning requires executing two fact lookups in a strict order during a forward pass. For a feed-forward neural network, this is only possible if the first fact (e.g. ``the performer of Imagine is John Lennon'') is stored in an earlier block than the second fact (e.g. ``the spouse of John Lennon is Yoko Ono''). Otherwise, if the first fact is stored in a later block (e.g. 20th transformer block) and the second fact in an earlier block (e.g. 10th block), by the time a model completes the first lookup to resolve the bridge entity (``John Lennon''), the forward pass can no longer use the bridge entity to look up the second fact.

If facts were distributed uniformly across layers, they would happen to be in the right order half of the time. Therefore, if layer ordering was the only reason for poor two-hop performance, one would expect two-hop accuracy to be around 50\%. In practice, this should be seen as a lower bound, since some facts might be represented redundantly, more than once.

\paragraph{Setup}
We force localizing facts in particular layers by layer-selective finetuning, i.e. dividing our training distribution into three datasets and training separately on each, involving only a particular layer range at each stage:

\begin{enumerate}[topsep=0pt]
    \item \textit{First one-hop facts} (e.g. ``the performer of Imagine is John Lennon'') are learned with layers 0-12 (with other layers frozen)
    \item \textit{Second one-hop facts} (e.g. ``the spouse of John Lennon is Yoko Ono'') are learned with layers 12-24 (with other layers frozen)
    \item \textit{Two-hop QA pairs} are learned with all layers updated.
\end{enumerate}

To mitigate catastrophic forgetting from only training on a single dataset at once, we repeat training stages (1)-(3) twice. Moreover, our training data uses the mixture described in the previous section: training on one-hop facts and both two-hop CoT and no-CoT QA pairs.

\paragraph{Results}

\begin{figure}[t]
    \centering
    \begin{subfigure}[b]{0.75\textwidth}
        \begin{center}
           \scriptsize{%
            \cblock{242}{169}{59} Baseline\quad
            \cblock{136}{199}{195} Staged, all layers\quad
            \cblock{128}{149}{229} Staged, layer-selective \\
            
            \vspace{1mm}
            \thinline{solid,gray} Loss on correct responses \quad
            \thinline{dashed,gray} Loss on random responses
           }
        \end{center}
    \end{subfigure}%
    \hfill
    \begin{subfigure}[b]{0.49\textwidth}
        \centering
        \includegraphics[width=\textwidth]{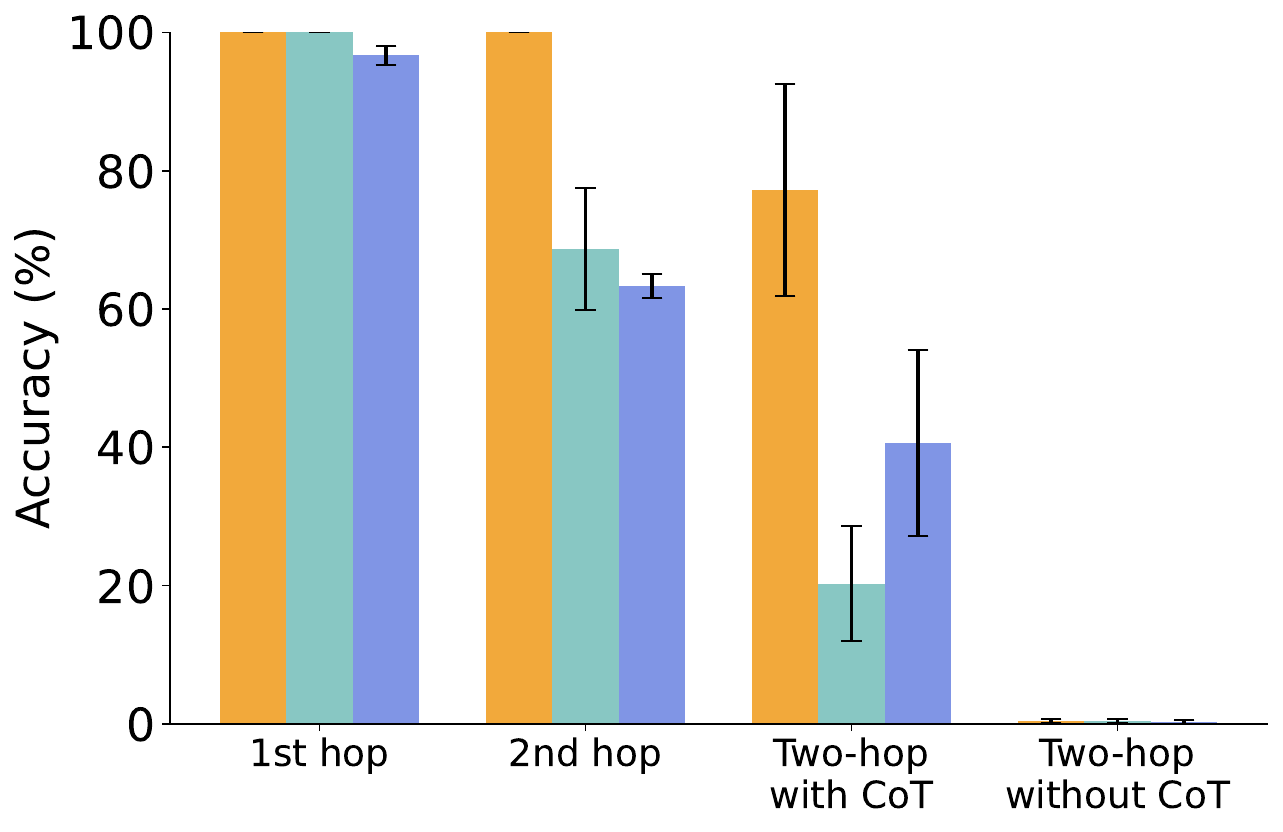}
        \label{fig:int2_avg_onehop}
    \end{subfigure}
    \begin{subfigure}[b]{0.49\textwidth}
        \centering
        \includegraphics[width=\textwidth]{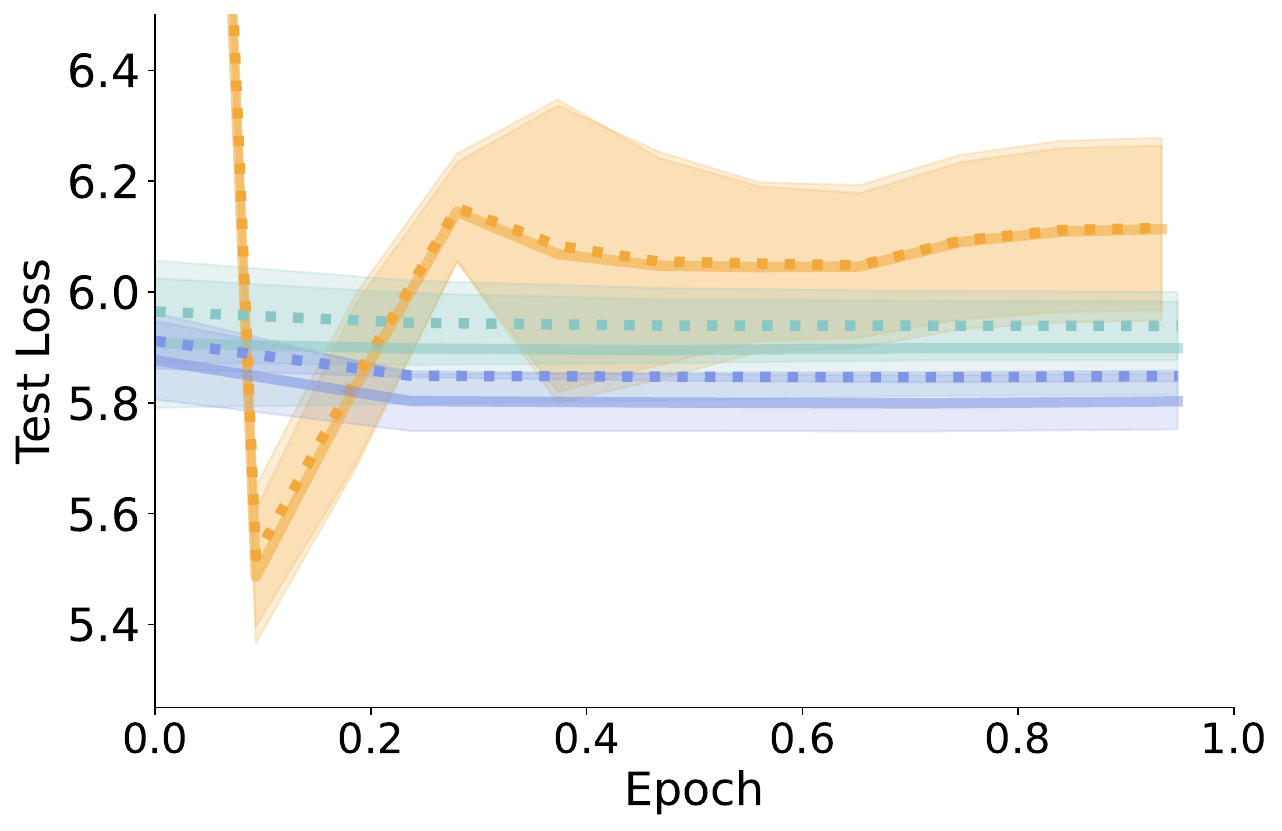}
        \label{fig:int2_avg_onehop}
    \end{subfigure}
    \vspace{-2em}
    \caption{Performance of Llama 3 models trained with variations in fact storage. Staged training negatively affects single-hop and two-hop CoT performance but it remains above zero. Our intervention (\cblock{208}{187}{255} staged, layer-selective) decreases test loss slightly, but the loss remains close to chance-level. Note: the y-axis of the loss plot is zoomed in compared to Figure~\ref{fig:experiment_1}.}
\label{fig:intervention_layer_order_results}
\end{figure}

We compare the following three setups:

\begin{enumerate}[topsep=0pt]
    \item \textbf{Baseline}. This is the setup from Section~\ref{sec:exp-fully-synthetic}, training on the full data mixture in a single stage with all layers trained.
    \item \textbf{Staged, with all layers trained}. This setup is a sanity check to show that staged training preserves most of the baseline's performance.
    \item \textbf{Staged, layer-selective training}. This is the intervention setup.
\end{enumerate}

As seen in Figure~\ref{fig:intervention_layer_order_results}, forcing one-hop facts to be localized in the correct order — with the first fact stored earlier than the second one — failed to elicit two-hop reasoning over fully synthetic facts. This means that correct knowledge localization in the forward pass is not enough to elicit two-hop reasoning over fully synthetic facts: the model still fails to connect pieces of knowledge for answering two-hop questions. 

Having failed to elicit two-hop reasoning through controlling where facts are stored, we next explore if providing supervision in the activation space could improve the two-hop reasoning capability.

\subsection{Activation supervision for two-hop reasoning}

\paragraph{Motivation}
The cross-entropy language modeling loss, used during LLM pretraining and supervised fine-tuning, treats the LLM as a black box and only supervises how the input tokens in the prompt are mapped to output tokens. From success of CoT performance, we know that such supervision is effective in teaching models to reason in explicit CoT. Since the reasoning trace is expressed in token space, the language modeling loss provides LLMs process-based supervision, giving useful gradients for each step of reasoning. However, for reasoning in latent space, the language modeling loss only provides outcome-based feedback (whether the predicted answer is correct) and is indifferent to whether an LLM arrives at the answer via memorization or two-hop reasoning. This motivates an intervention trying to directly incentivize answering a question via two-hop reasoning

\paragraph{Setup}
We add an auxiliary loss $\mathcal{L}_\text{aux}$ that complements outcome-based supervision from the language modeling loss with process-based feedback in the activation space. More specifically, we encourage the model to resolve the bridge entity in activation space whenever it is prompted with a two-hop question. We encourage such resolution by ensuring that a given hidden state (output of a transformer block) is either similar to a vector representation of the bridge entity or predictive of it.

We apply the auxiliary objective to the output of a single transformer block at a single token position during fine-tuning. We sweep over several blocks to apply this loss on and choose block 10 (out of 32) of Llama 3 8B Instruct. To determine the token position to apply loss on, we look for the last token of the description of the bridge entity in the question, e.g. ``gine'' in ``Who is the spouse of the singer of the song Ima\underline{gine}?''. Let's call this activation vector $h$.

We consider two auxiliary objectives:
\begin{enumerate}
    \item \textit{Logit lens}. This objective function is inspired by an interpretability technique \citep{nostalgebraist2020interpreting} We compute logits $y$ as $y = W_U \text{RMSNorm}(h)$, where $\text{RMSNorm}(\cdot)$ denotes the final RMSNorm \citep{zhang2019rootmeansquarelayer} layer of Llama 3 8B Instruct during training.  We then compute $\mathcal{L}_\text{aux} = \text{CE}(e_2, y)$, where $\text{CE}(\cdot)$ is the standard cross-entropy loss and $e_2$ is the token corresponding to bridge entity, e.g. ``John Lennon''. This is possible because we ensure all bridge entities are single-token.
    \item \textit{Embed lens}. We compute $\mathcal{L}_\text{aux} = -\text{CosSim}(h, W_E e_2)$, where $\text{CosSim}(\cdot)$ is the cosine similarity loss, $h$ is the hidden state at the chosen layer and position, and $W_E e_2$ is the embedding of the bridge entity token.
\end{enumerate}

In both cases, our final loss is computed as $\mathcal{L} =  \mathcal{L}_\text{LM} + c\mathcal{L}_\text{aux}$, where $ \mathcal{L}_\text{LM}$ is the standard language modelling loss and  the coefficient $c$ is a hyperparameter. Based on our sweeps, we found that 0.01 and 0.1 were the best settings for logit lens and embed lens, respectively. Once again, our training data uses the setup described for Hypothesis 2 experiments: training on one-hop facts and both two-hop CoT and no-CoT QA pairs.

\paragraph{Results}

\begin{figure}[t]
\centering
    \begin{subfigure}[b]{0.75\textwidth}
        \centering
        \scriptsize{
            \cblock{242}{169}{59} Baseline\quad
            \cblock{192}{219}{132} Logit lens\quad
            \cblock{136}{199}{195} Embed lens \\
            \vspace{1mm}
            \thinline{solid,gray} Loss on correct responses \quad
            \thinline{dashed,gray} Loss on random responses
            \vspace{1mm}
        }    
    \end{subfigure}
    \hfill
    \begin{subfigure}[b]{0.33\textwidth}
        \centering
        \includegraphics[width=\textwidth]{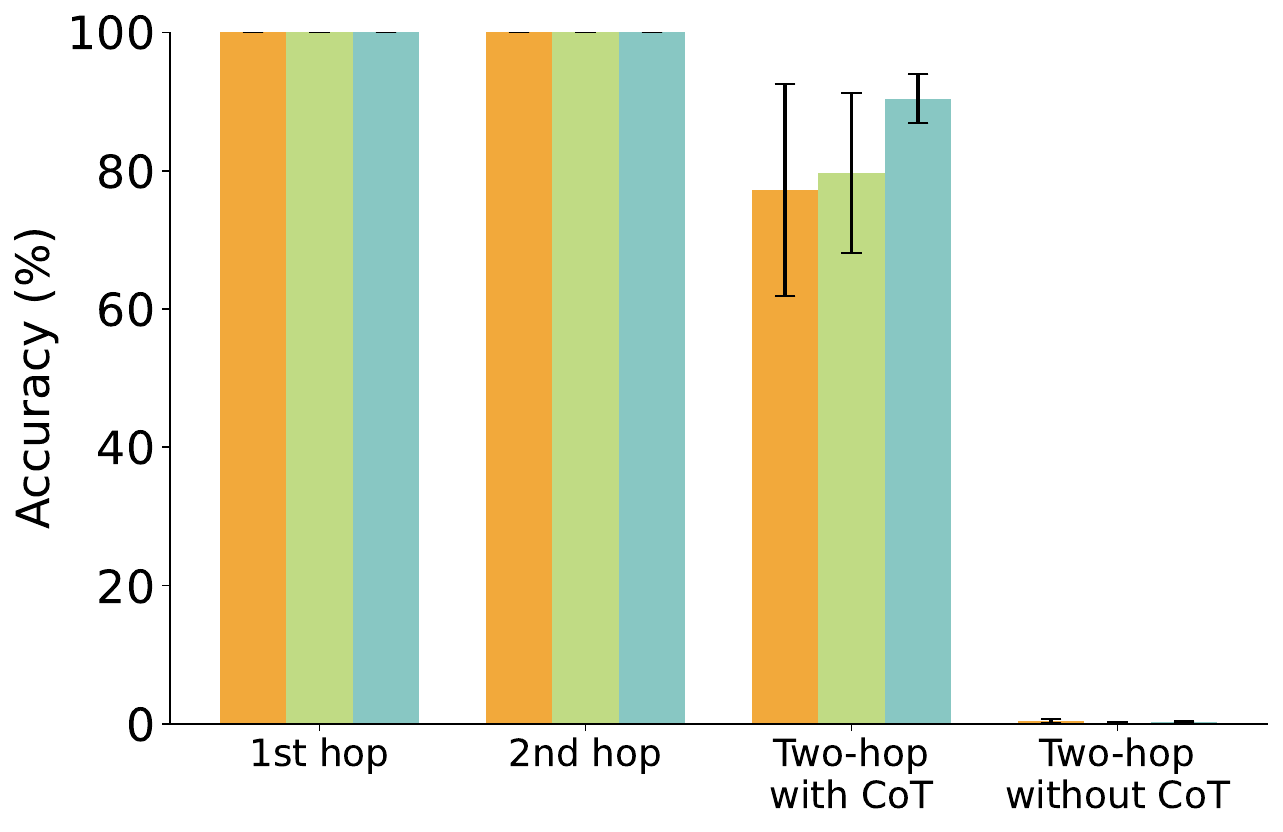}
    \end{subfigure}%
    \begin{subfigure}[b]{0.21\textwidth}
        \centering
        \includegraphics[width=\textwidth]{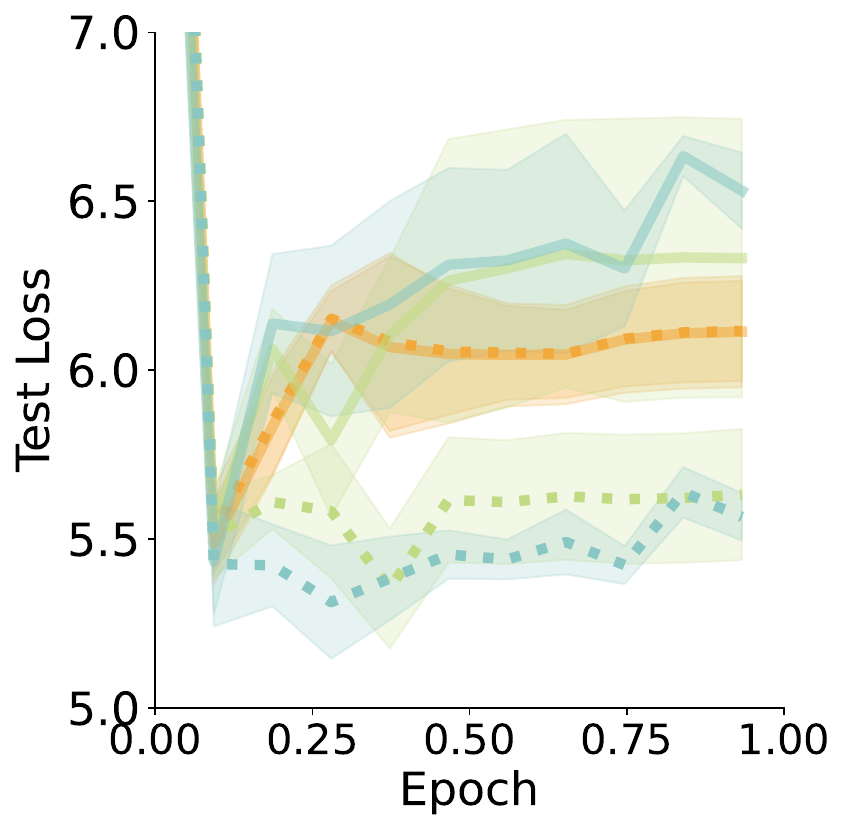}
    \end{subfigure}%
    \begin{subfigure}[b]{0.21\textwidth}
        \centering
        \includegraphics[width=\textwidth]{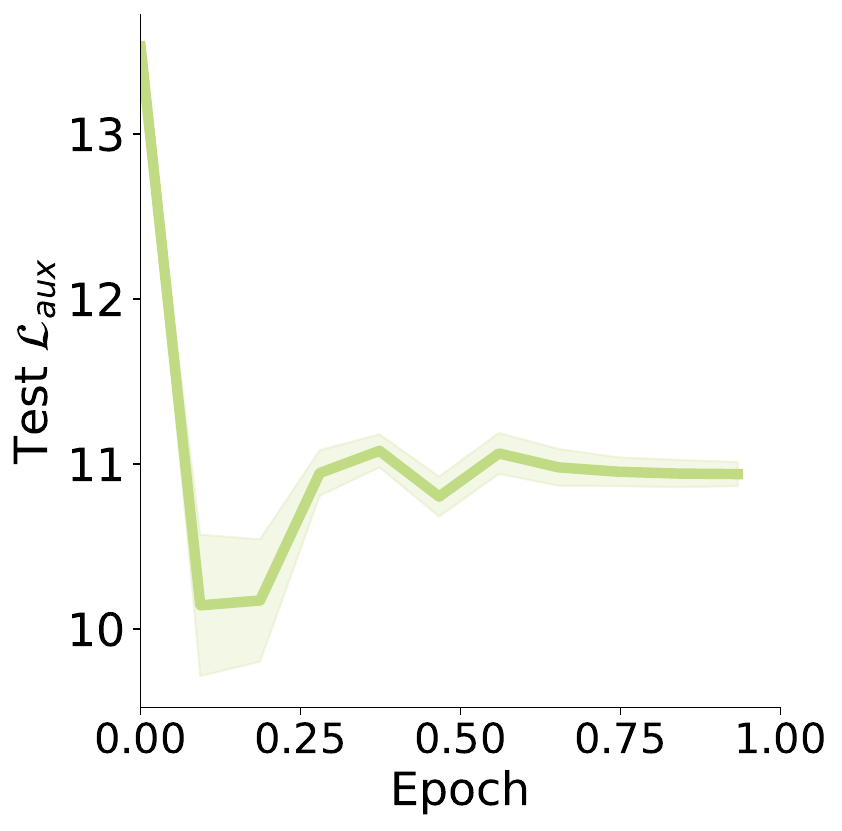}
    \end{subfigure}%
    \begin{subfigure}[b]{0.21\textwidth}
        \centering
        \includegraphics[width=\textwidth]{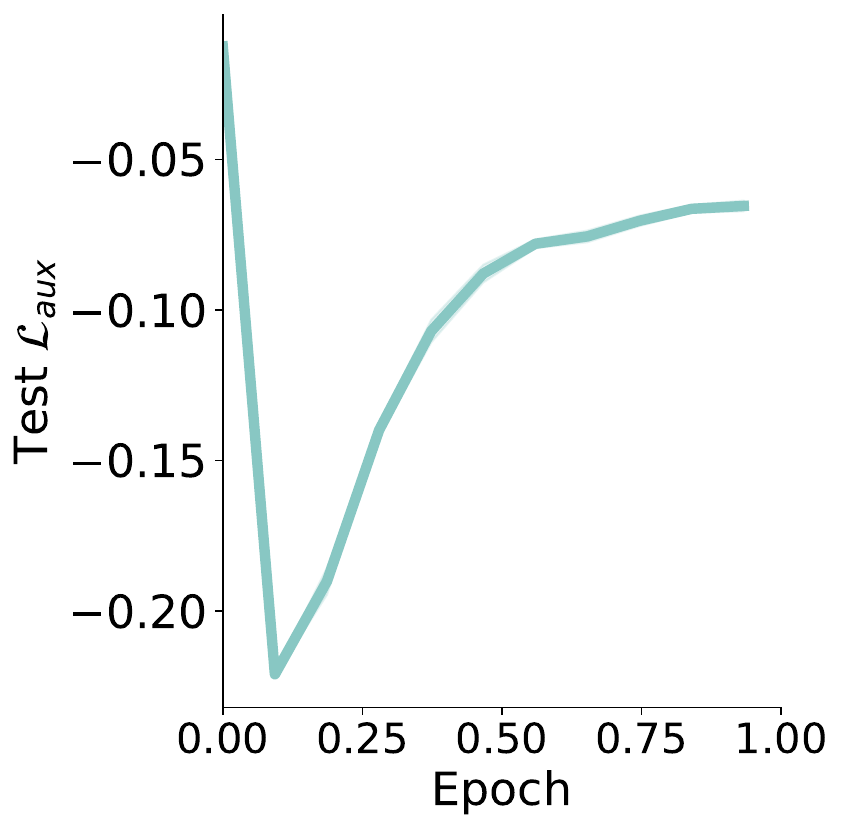}
    \end{subfigure}%
    \caption{Performance of models trained with different objectives intended to induce latent reasoning. Our interventions (\cblock{222}{187}{155} logit and \cblock{250}{176}{228} embed lens) do not boost two-hop no-CoT accuracy. The two rightmost plots show empirical values of $\mathcal{L}_\text{aux}$ on the test set during training for both auxilary losses. $\mathcal{L}_\text{aux}$ tends to decrease for both, but it's either unstable (for logit lens) or tends to show signs of rapid overfitting (for embed lens). Note that cross-entropy of~10 and cosine similarity of 0.2 are poor values close to chance-level; perfect generalization would correspond to cross-entropy 0 and cosine similarity of~1.}
\label{fig:intervention3}
\end{figure}

We compare the following three setups:
\begin{enumerate}
    \item \textbf{Baseline}: Training on one-hop facts and both two-hop CoT and no-CoT QA pairs with just $ \mathcal{L}_\text{LM}$.
    \item \textbf{Logit lens}. This is the Logit lens setup, using the best coefficient $c$ value from a sweep.
    \item \textbf{Embed lens}. This is the Embed lens setup, using the best coefficient $c$ value from a sweep.
\end{enumerate}

As seen in Figure~\ref{fig:intervention3}, encouraging the model to resolve the bridge entity during its forward pass failed to elicit two-hop reasoning. As seen by the evaluation $\mathcal{L}_\text{aux}$, learning to resolve bridge entities during training does not generalize to resolving other bridge entities on evaluation prompts despite the training $\mathcal{L}_\text{aux}$ reaching zero.

\begin{customblockquote}
\label{finding2}
Interventions to (i) force a correct fact storage order across transformer layers and (ii) encourage the first reasoning hop both fail to enable models to compose facts without CoT.
\end{customblockquote}

\section{Experiment 3: Two-hop reasoning with facts in the same document}
\label{sec:experiment_3}

\paragraph{Motivation} While in Section~\ref{sec:exp-fully-synthetic} we found that models completely fail at latent two-hop reasoning over synthetic facts that were learned through fine-tuning, we want to explore whether this limitation persists in easier settings. We investigate two such settings: (i) fine-tuning where related facts are always presented together in training documents, and (ii) in-context evaluation without fine-tuning, with facts provided in the prompt. These setups reduce the difficulty of retrieving relevant facts, while still requiring the model to perform latent reasoning to combine them.

\paragraph{Setting 1: Fine-tuning with related facts in the same document} We build on Experiment 1, modifying how one-hop QA pairs described in Section~\ref{subsec:experiment1_experimental_setup} are presented in the training data. Specifically, we ensure facts about $\langle e_1, e_2 \rangle$ and $\langle e_2, e_3 \rangle$ from the same triplet always appear in the same training document. We always put the facts in the logical order within a training document to avoid the reversal curse \citep{berglund2024thereversalcurse}. The rest of the setup is identical to Experiment 1.

\begin{figure}[ht]
    \centering
    \begin{subfigure}[c]{0.34\textwidth}
        \scriptsize
        \begin{tabular}[c]{p{17em}}
            \textbf{Training example for Setting 1: Fine-tuning with facts in same training docs} \\
            \midrule
            \myalign{l}{\systemmessage{15em}{\textbf{System:} You will be given questions about fictional characters from the ``Spouses'' saga. \\\\Answer the following question.}} \\ 
            
            \myalign{r}{\adjcontextb{15em}{\textbf{User:} Tell me about Russ.}}
            \\
            \myalign{l}{\adjbotc{15em}{\textbf{Assistant:} Russ is married to Hay. Hay was born in Showing.}} \\
            \midrule
        \end{tabular}
    \end{subfigure}
    \begin{subfigure}[c]{0.35\textwidth}
        \scriptsize
        \begin{tabular}[c]{p{18em}}
            \textbf{Prompt used in Setting 2: In-context evaluation without fine-tuning} \\
            \midrule
            \myalign{l}{\systemmessage{16em}{\textbf{System:} [...] Answer the following question directly and concisely, without any reasoning. [...]}} \\ 
            
            \myalign{r}{\adjcontextb{16em}{\textbf{User:} [...facts from other triplets...]\newline Hay was born in Showing. [...] \newline Russ is married to Hay. [...] \newline In which city was Russ's spouse born?}}
            \\
            \myalign{l}{\adjbotc{16em}{\textbf{Assistant:}}} \\
            \midrule
        \end{tabular}
    \end{subfigure}
    \begin{subfigure}[c]{0.29\textwidth}
        \centering
        \includegraphics[width=\textwidth]{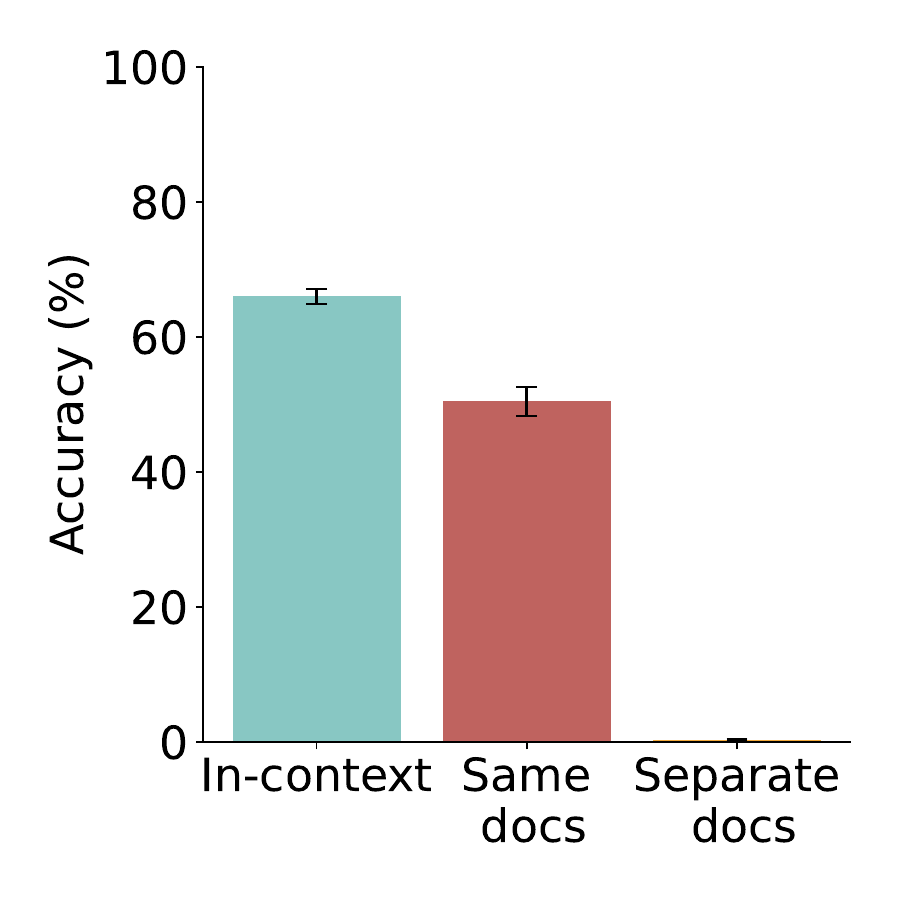}
    \end{subfigure}
    \caption{\textbf{Top:} Experiment 3 settings where related facts appear in the same document during fine-tuning (Setting 1) or in the prompt without any fine-tuning (Setting 2). The rest of the experiment setup in Setting 1 is identical to Experiment 1. \textbf{Bottom:} Accuracy of Llama 3 8B Instruct on two-hop questions without CoT across three settings, including the original setup of fine-tuning with facts in separate documents (3 seeds per setting). Both new settings show non-zero no-CoT accuracy, suggesting that models can perform latent two-hop reasoning in specific settings, i.e. when underlying facts are in the same documents during training or in the prompt during inference.}
    \label{fig:experiment_2_settings}
\end{figure}

\paragraph{Setting 2: In-context evaluation with facts in the prompt} We evaluate models without any fine-tuning by providing facts involving multiple entity triplets (10 distractors and 1 target triplet) in the prompt before asking a two-hop question. The distractor triplets are sampled randomly from other test triplets for every sample, and the order of facts in the context is randomized per sample.
Figure~\ref{fig:experiment_2_settings} shows example formats for both settings.

\paragraph{Results} In contrast to the negative results in Experiment 1 and 2, both same-document fine-tuning and in-context setups achieve non-zero two-hop accuracy without CoT (see Figure~\ref{fig:experiment_2_settings}). We additionally experiment with adding simple distractor information between the two facts for the same-document fine-tuning setting, and find that accuracy can drop depending on amount and type of distractors but the test loss always remains above chance-level (see Appendix~\ref{sec:appendix_distractors} for details).

While our experiments do not explain why these setups work better than separate-document fine-tuning, we can speculate about the reasons. In the same-document setup, supervised fine-tuning makes $e_1$ directly predictive of $e_3$ since they appear together during training. In contrast, with separate documents, $e_1$ is only predictive of $e_2$, which is not sufficient to make it predictive of $e_3$. Notably, $e_1$ can only become predictive of $e_3$ in the same-document setting if $e_3$ appears after $e_1$ in a given document, due to the reversal curse. For the in-context setting, two-hop reasoning does not need to involve retrieval of knowledge from the MLPs, allowing attention layers to attend to both facts at once, making composition easier.

These results paint a nuanced picture — models \emph{can} perform latent two-hop reasoning when the facts to be composed were already seen together by the model. This suggests that knowledge that humans have already identified as related by putting it in the same documents or in the same prompt is easier for models to compose without chain-of-thought reasoning.

\begin{customblockquote}
\label{finding3}
Models successfully compose new facts without chain-of-thought when those facts co-occurred in the same fine-tuning document or in the same test-time prompt.
\end{customblockquote}

\section{Experiment 4: Two-hop reasoning over semi-synthetic facts}
\label{subsec:exp-two-hop-reasoning-semi-synthetic}

\paragraph{Motivation}

In this paper's experiments, we fine-tune models on synthetic facts to ensure that successfully answering two-hop questions cannot be accomplished through memorization or reasoning shortcuts, and therefore requires two-hop reasoning. However, one might worry that the facts learned through fine-tuning might fundamentally differ from how natural facts are learned during pretraining \citep{jain2024mechanisticallyanalyzingeffectsfinetuning}, and that failure to reason over these fully synthetic facts is not representative of models' general latent reasoning abilities. To alleviate this concern, we investigate a hybrid, \textit{semi-synthetic} two-hop reasoning setup similar to the one proposed by \cite{feng2025extractivestructureslearnedpretraining}, where we rely on models' pre-existing knowledge of \textit{one} of the two atomic facts.

\paragraph{Setup}

We generate 17 different synthetic datasets, each including facts connecting fictional people ($e_1$, e.g. ``Nadia Hassan-Virtanen'') with real-world entities ($e_2$, e.g., country of birth is ``Finland''). We then evaluate models by asking two-hop questions about attributes of real-world entities related to these fictional people. For example, the ``favorite programming languages'' dataset assigns favorite real-world programming languages ($e_2$) to fictional people ($e_1$), allowing us to ask two-hop questions about real attributes of those languages (release year, creator, paradigm, etc.). For each of the 17 datasets, we fine-tune Llama 3 8B Instruct only on the synthetic ``first-hop'' facts (e.g. ``Nadia Hassan-Virtanen's favorite programming language is Scala''), and rely on models' pre-existing knowledge of the natural ``second-hop'' facts (e.g. ``The creator of Scala is Martin Odersky''). We evaluate models on two-hop questions of the form ``Consider the $r_1$ of $e_1$. What is $r_2$ of $e_2$?'' (e.g. ``Consider the favorite programming language of Nadia Hassan-Virtanen. What is its release year?'').
In contrast to our other experiments, we do not paraphrase synthetic facts for simplicity, and instead train for multiple epochs with a carefully tuned learning rate, following \cite{feng2025extractivestructureslearnedpretraining}\footnote{In preliminary experiments, we find that the stark difference in no-CoT performance between the fully synthetic setup from Section~\ref{sec:exp-fully-synthetic} and the semi-synthetic setup in Section~\ref{subsec:exp-two-hop-reasoning-semi-synthetic} is not explained by differences in learning rate, number of epochs, or including paraphrases of the same facts during training.}.

\begin{figure}[!t]
    \centering
    \begin{subfigure}[b]{0.49\textwidth}
        \centering
        \includegraphics[width=\textwidth]{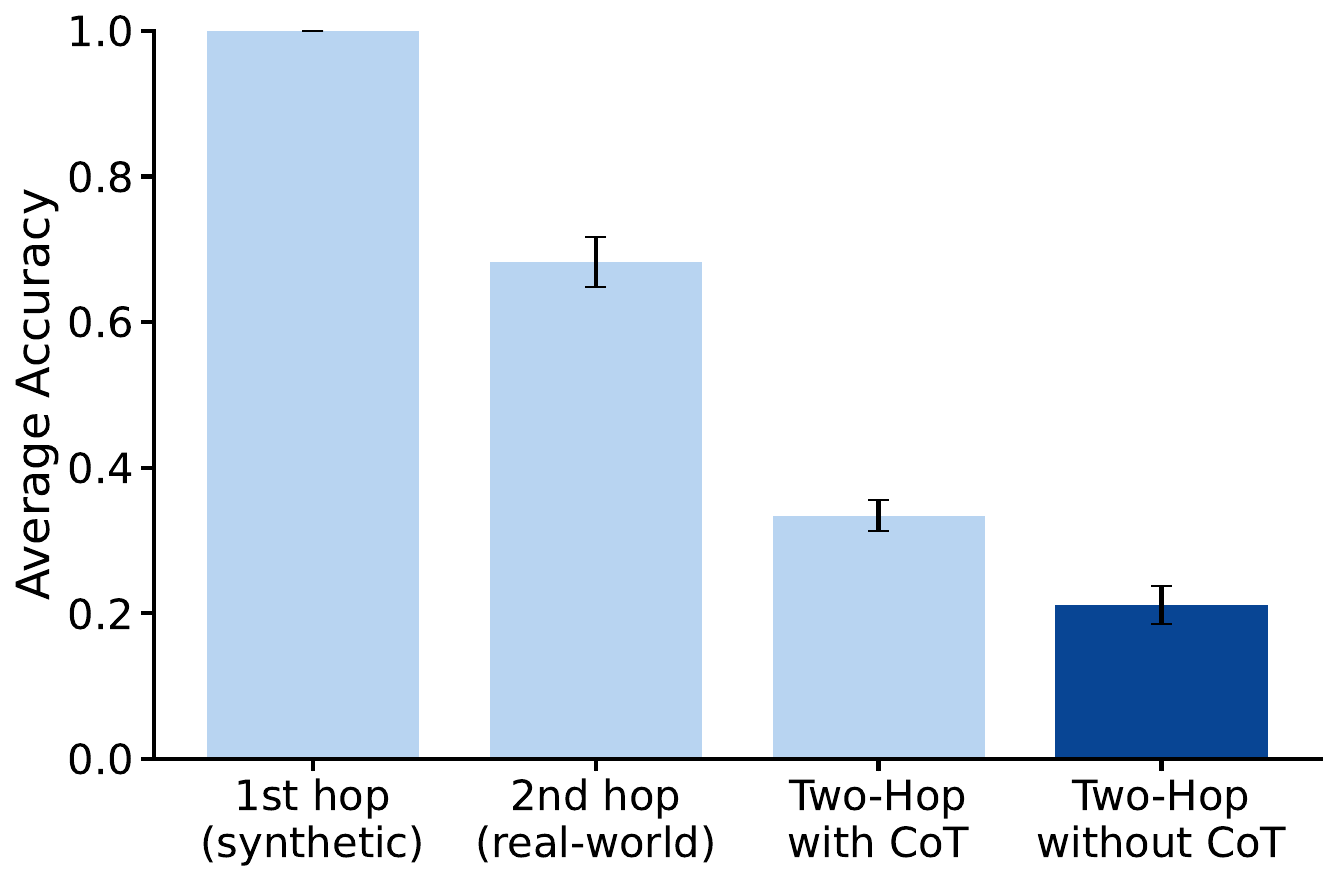}
    \end{subfigure}
    \hfill
    \begin{subfigure}[b]{0.49\textwidth}
        \centering
        \tiny{
            \thinline{dashed,semi_synthetic_no_cot} Loss on random $e_3$s \quad
            \thinline{solid,semi_synthetic_no_cot} Loss on correct $e_3$s
        }
        \includegraphics[width=\textwidth]{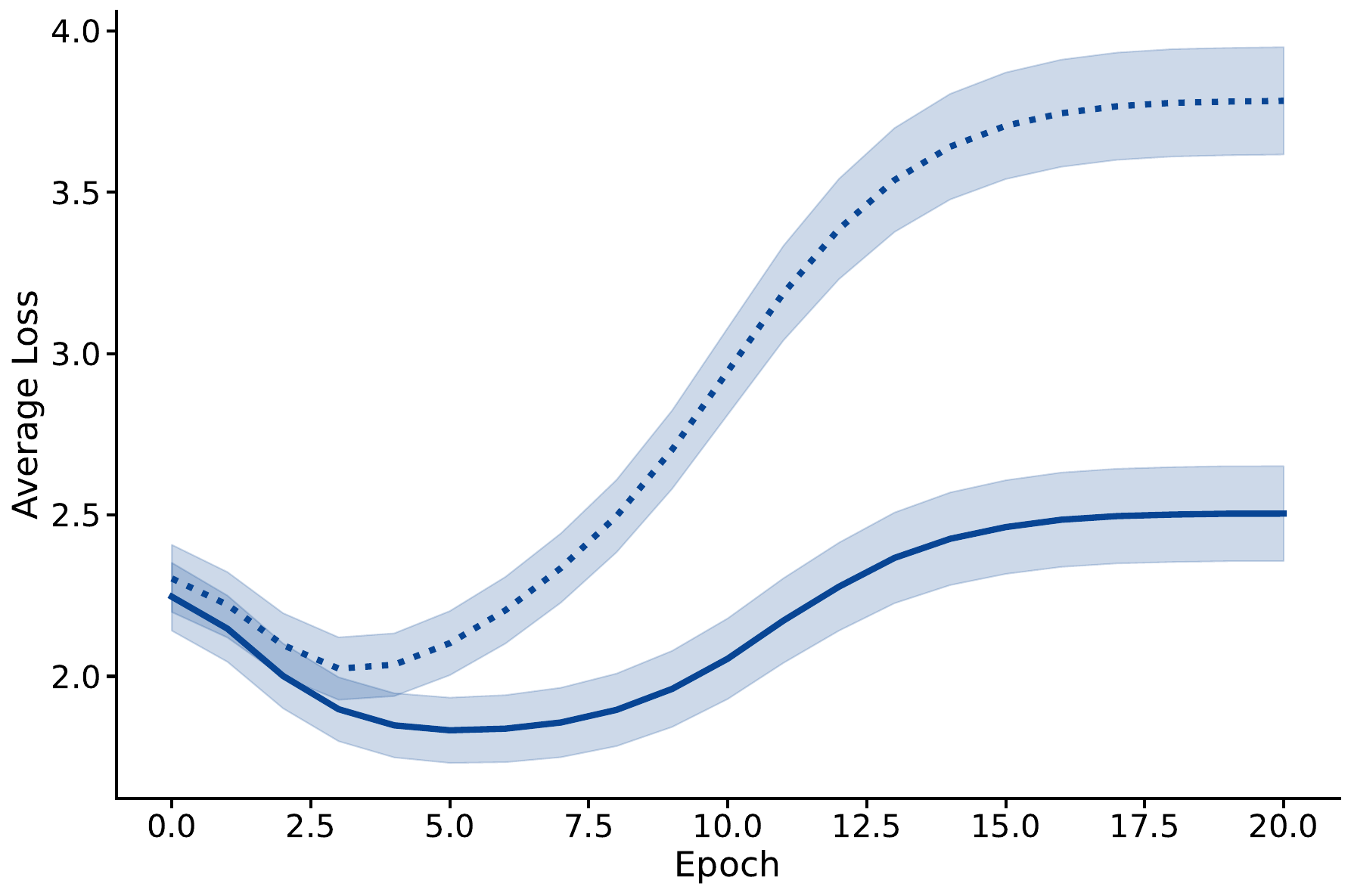}
    \end{subfigure}
    \caption{\textbf{In the \textit{semi-synthetic} fact setup, models \emph{are} capable of latent two-hop reasoning, achieving non-zero accuracy and a significant reduction in loss over a random baseline}. We fine-tune Llama 3 8B Instruct models on 17 different synthetic ``first-hop'' fact datasets, connecting fictional people with real-world entities (e.g. ``Nadia Hassan-Virtanen's favorite programming language is Scala'') and rely on the model's knowledge of the natural ``second-hop'' facts -- attributes of these real-world entities (e.g. ``The creator of Scala is Martin Odersky''). The results are averaged across models, each trained with three seeds on one of synthetic datasets consisting of around 20 `first-hop' facts, and evaluated on questions about two to four attributes per dataset. A more detailed breakdown of the results is included in Appendix~\ref{appendix:additional_semi_synthetic_results}.}
    \label{fig:semi_synthetic}
\end{figure}

\paragraph{Results}

We find definitive evidence that LLMs are capable of latent two-hop reasoning. Across the 17 training dataset types (evaluated on 63 different question signatures total) we obtain (i)~near--perfect accuracy on the newly taught first-hop questions ($>\!99\%$), (ii)~high accuracy on the natural second-hop questions (average $\approx\!65\%$) and, most importantly, (iii)~clear above-chance performance on two-hop questions \textit{without} chain-of-thought.  The average no-CoT accuracy rises to roughly 20\% (Figure~\ref{fig:semi_synthetic},~left), compared with the near-zero performance observed in the fully–synthetic set-up of Section~\ref{sec:exp-fully-synthetic}. Similarly, the no-CoT test loss drops well below the random-response baseline (Figure~\ref{fig:semi_synthetic},~right). Since it is impossible to achieve above-chance performance in this experiment through shortcuts, these results unequivocally show that LLMs are capable of latent two-hop reasoning.

We find it surprising that the semi-synthetic setup gives rise to latent two-hop reasoning. A natural hypothesis for why models are not capable of latent two-hop reasoning when both facts are synthetic is that representations of facts naturally learned during pretraining are importantly different from representations learned through fine-tuning. However, this does not explain why models are able to reason over a mixture of natural and synthetic facts. We leave a detailed investigation of the stark no-CoT performance difference between the fully synthetic and semi-synthetic setups to future work.

\begin{customblockquote}
\label{finding4}
LLMs are capable of composing two separately learned facts, as long as one of the facts is naturally acquired during pretraining (the second fact can be synthetic and through fine-tuning)
\end{customblockquote}

\section{Discussion}

In this section, we speculate about the reason why LLMs often fail to compose facts learned separately without CoT. We also discuss limitations of our experimental setups and possible future directions.

\subsection{Hypothesis for the cause of two-hop reasoning limitations}

\begin{figure}[t]
    \centering
    \includegraphics[width=\textwidth]{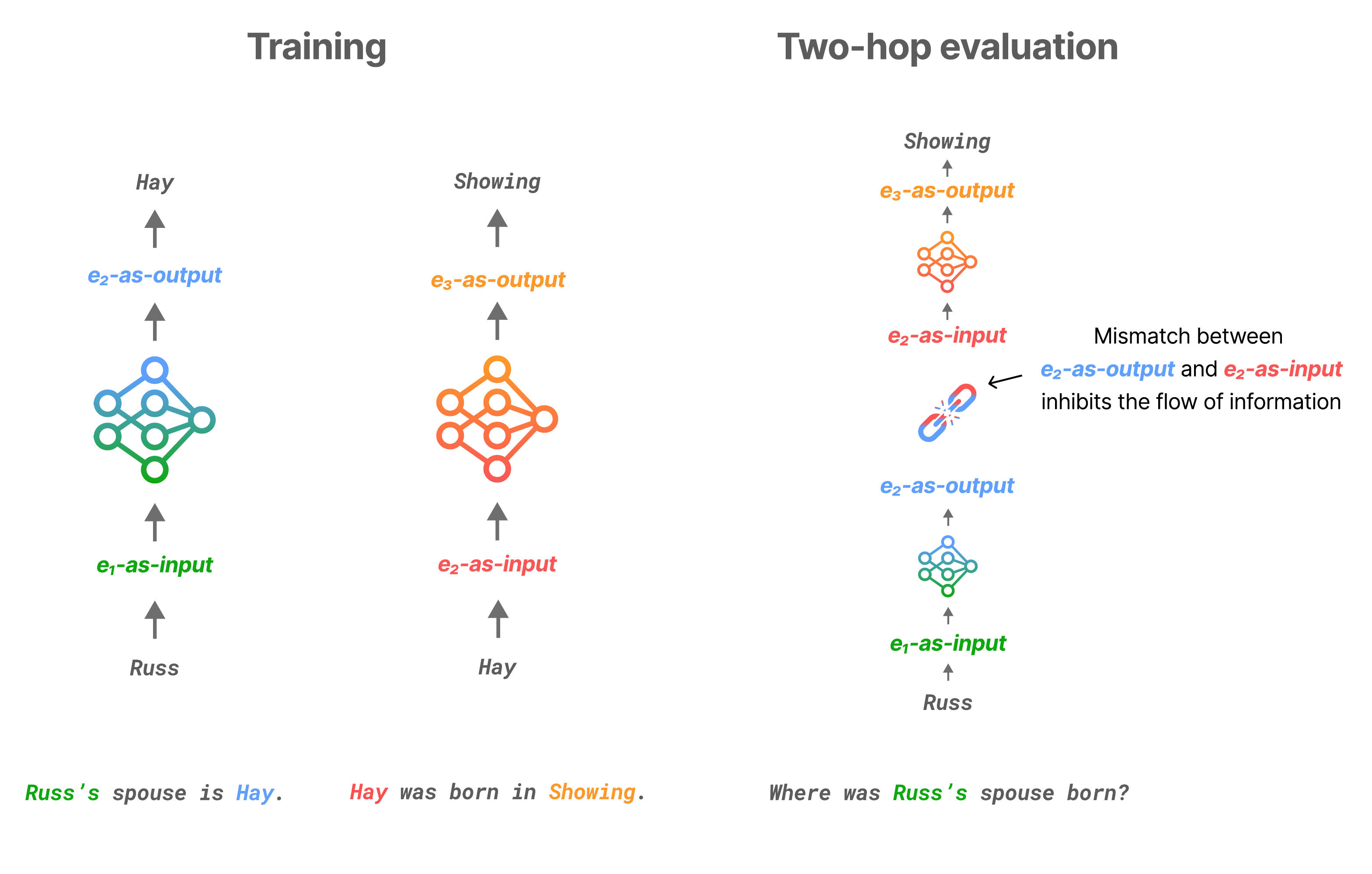}
    \caption{Hypothesized mechanism behind the failures of latent two-hop reasoning. During training (left), the model learns independent mappings between entities. During two-hop evaluation without CoT (right), the model needs to use the bridge entity ($e_2$) both as an output and an input, creating a representational mismatch that inhibits the flow of information.}
    \label{fig:mechanism_explainer}
\end{figure}

We hypothesize that the failure of latent two-hop reasoning in cases where CoT two-hop reasoning suceeds can be accounted for by how feedforward LLMs represent the bridge entity ($e_2$) during training on one-hop facts and during inference on two-hop questions (see Figure~\ref{fig:mechanism_explainer}). During training, facts are learned independently -- when learning ``Russ's spouse is Hay", the model learns to map $e_1$ (Russ) to $e_2$ (Hay), and separately when learning ``Hay was born in Showing", it learns to map $e_2$ to $e_3$ (Showing). However, during two-hop inference without CoT, the model needs to use $e_2$ in two different roles: first as an output of a memory lookup when processing $e_1$, then as an input for looking up $e_3$. We speculate that this creates an information flow bottleneck --- the representation of $e_2$ optimized for being an output (i.e. causing the LLM to produce high logits on the corresponding output token) may not be suitable for being an input for a fact lookup, inhibiting the chain of reasoning. This bottleneck does not exist when facts appear together during training (as the model can learn direct $e_1$ to $e_3$ mappings) or when using CoT (as $e_2$ is explicitly generated as a token that can then be consumed as input).

\subsection{Limitations of our experiments}

In this paper, we try to investigate the capabilities of LLMs in naturalistic settings, while controlling for confounders plaguing prior work. Reconciling the need for a clean setup and plausibility required several design choices that could be controversial.

\paragraph{Fine-tuning vs pretraining} 

In order to have a clean experimental setup, we fine-tune models on fictional facts. However, one might worry that the cleanliness of this setup is fundamentally different from how knowledge is normally acquired by LLMs during pretraining \citep{jain2024mechanisticallyanalyzingeffectsfinetuning}. This difference might manifest in diversity of the data distribution and the scale of the training dataset.

To ensure the diversity of the training distribution, we include multiple (30) paraphrases of each fact, which leads language models to learn the underlying logical facts as opposed to just memorizing the sentences that express them  \citep{berglund2023takencontextmeasuringsituational,berglund2024thereversalcurse}. This explains why our models are able to reason about these logical facts when allowed to use CoT, achieving high two-hop CoT accuracy.

Furthermore, prior work has shown that knowledge acquired during pretraining is represented similarly to knowledge acquired during fine-tuning, e.g. the Reversal Curse has been observed in models pretrained on natural data \citep{grosse2023studyinglargelanguagemodel}, models pretrained on large-scale synthetic data \citep{allenzhu2024physicslanguagemodels32}, and models fine-tuned on synthetic facts \citep{berglund2024thereversalcurse}.

\paragraph{Ratio of two-hop to single-hop facts}

Prior work has shown that a particular ratio of the number of atomic and two-hop facts involving a given entity is crucial for incentivizing two-hop reasoning as opposed to memorizing answers to two-hop questions \citep{wang2024grokkedtransformersimplicitreasoners}. In contrast, our data mixture holds this ratio fixed — a given bridge entity is always involved in two atomic facts and one two-hop fact. This might create insufficient pressure for the model to learn two-hop reasoning. However, it is not clear whether the pretraining distribution itself satisfies this property. Future work could explore the effect of varying this ratio in naturalistic settings.

\paragraph{The strength of activation-level supervision}

Our auxiliary objectives incentivize the model to resolve the bridge entity (first hop) in activation space. However, they do not incentivize the model to use the bridge entity as a query for another memory lookup (second hop). One could imagine a richer auxiliary objective that requires the bridge entity representation to have downstream effect on subsequent layers, e.g. maximizing the gradient of the final answer w.r.t. to the representation of the bridge entity \citep{koh2017}. However, such loss function would require computing second-order gradients, which is challenging to implement in distributed training setups for LLMs.

\paragraph{Two-hop reasoning as a toy model of latent reasoning} Finally, one might ask whether two-hop reasoning is a good toy model for latent reasoning in LLMs. The capability of composing two facts retrieved from parametric memory, without explicitly generating the intermediate fact, is a convinient proxy for a general capability of performing multi-step reasoning in latent space. However, it is different from the kinds of reasoning most relevant in the context of AI safety, e.g. reasoning about the flaws of a reward function during training in order to exploit them \citep{baker2025monitoringreasoningmodelsmisbehavior} or reasoning about a model's evaluation procedure needed for convincingly pretending to be aligned \citep{greenblatt2024alignmentfakinglargelanguage}. For instance, in the cases mentioned above, most relevant facts might be available in context or might be saliently represented in model's parameters which might constrast with how trivia facts are being represented. A more realistic approach to evaluating safety-relevant latent reasoning capaiblities might involve end-to-end evaluations of LLM agents' capability to complete harmful tasks without CoT \citep{emmons2025chainthoughtnecessarylanguage}.

\section{Related work}

\paragraph{Externalized reasoning} Prompting LLMs to externalize their reasoning (or, ``think step by step'')  has long been known to improve their performance on various reasoning tasks \citep{reynolds2021promptprogramminglargelanguage,wei2023chainofthoughtpromptingelicitsreasoning,Kojima2024large}. This prompting strategy is known as ``chain-of-thought'' (CoT). Even though the advantages of CoT are not uniform across tasks \citep[it primarily benefits mathematical and symbolic reasoning;][]{sprague2024cotcotchainofthoughthelps}, giving LLMs the ability to spend a certain amount of tokens on thinking provably extends the complexity class of problems they can tackle \citep{merrill2024expressivepowertransformerschain}.
\cite{openai2024o1} has recently shown how the capability of LLMs to take advantage of CoT reasoning can be further improved with outcome-based reinforcement learning finetuning, leading to state-of-the-art results across multiple benchmarks \citep{math,rein2024gpqa}. Despite those boosts, CoT does not always reliable reflect the causal process that leads an LLM to giving a certain answer \citep{lanham2023measuringfaithfulnesschainofthoughtreasoning,10.5555/3666122.3669397,anwar2024foundational}. Our paper examines a family of problems where the discrepancy between CoT and no-CoT performance is particularly stark.

\paragraph{Two-hop reasoning} Multi-hop question answering is a long-standing problem in natural language processing \citep{yang-etal-2018-hotpotqa}, blending together factual recall and reasoning. \cite{press-etal-2023-measuring} have attempted to single out the reasoning component of two-hop question answering by measuring the \emph{compositionality gap} of an LLM — the fraction of two-hop questions for which the LLM can answer the underlying (single-hop) facts but fails to combine them when answering a two-hop question. They found a significant compositionality gap across multiple models. \cite{berglund2023takencontextmeasuringsituational} and \cite{yang-etal-2024-large-language-models} found inconclusive evidence for two-hop reasoning: LLM performance varied significantly across domains. Following up on the latter paper, \cite{biran2024hoppinglateexploringlimitations} found that in many cases, even if the first hop successfully resolves the bridge entity, this information often does not get consumed by the upstream layers. \cite{binder2024lookinginwardlanguagemodels} speculate that a form of two-hop reasoning (involving only one memory lookup) underlies introspective capabilities of LLMs. \cite{feng2025extractivestructureslearnedpretraining} propose a theoretical framework for studying circuits involved in two-hop reasoning and introduce a semi-synthetic finetuning setup for eliciting two-hop reasoning. \cite{guo2025llmsperformtwohopreasoning} analyze how transformers learn to perform two-hop reasoning over facts provided in context. \cite{lindsey2025biology} analyze circuits involved in two-hop reasoning by computing attribution graphs \citep{ameisen2025circuit} and observe that Claude 3.5 Haiku posseses true two-hop reasoning circuits which nonetheless coexist alongside ``shortcut'' reasoning circuits.

\paragraph{Fundamental limitations of latent reasoning in transformers}

Transformers consist of a sequence of feedforward networks \citep{NIPS2017_3f5ee243} and are subject to strict bounds on the class of problems they can solve (see \citep{strobl-etal-2024-formal} for a survey). \cite{feng2023towards} first proved that transformers without CoT cannot solve certain problems and \cite{merrill-sabharwal-2023-parallelism,merrill2023a} further proved that the problems they can solve without CoT belong to the circuit complexity class $\TC^0$. It is not clear, however, how practical these bounds are for frontier models that consist of more than a hundred of transformer blocks. Fundamental limits to learnability of certain algorithms might impose tighter bounds on LLM reasoning capabilities: \cite{10.5555/3666122.3669203} found that transformer capabilities of solving some compositional problems (such as multi-digit addition or dynamic programming) scale unfavorably with problem complexity. Similarly, \cite{ye2024physicslanguagemodels21} found that transformers can only be trained to solve some mathematical problems when they are sufficiently deep.

\paragraph{Eliciting latent reasoning capabilities via finetuning} 

\cite{wang2024grokkedtransformersimplicitreasoners} show that two-hop reasoning circuits can be learned through grokking (training a low-capacity model for 50 epochs) but those circuits remain brittle (do not generalize to out-of-distribution examples). Moreover, while \citeauthor{wang2024grokkedtransformersimplicitreasoners} focus solely on pretraining toy models on artificial data (each example is three tokens long), we finetune LLMs close to frontier (Llama 3.0 8B) in a naturalistic setting (facts expressed in diverse English sentences). \cite{pfau2024letsthinkdotdot} train models to use meaningless filler tokens (e.g., `...') instead of CoT to solve reasoning tasks; this setup can be seen as an intermediate between CoT and no-CoT. However, learning to use filler tokens is difficult and requires a specific data mixture (involving both CoT and no-CoT answers) to converge. A related line work work focused on distilling CoT reasoning, i.e. training models to zero-shot give answers similar to those they would give after CoT \citep{zelikman2022star,zelikman2024quietstar,hsieh-etal-2023-distilling,mukherjee2023orcaprogressivelearningcomplex,chen-etal-2024-learning-maximize,yu2024distilling21}. A particularly succesful example of this approach involves gradual CoT distillation: progressively discarding steps of arithmetic CoT until only a small fraction of the original CoT remains \citep{deng2024explicitcotimplicitcot}. However, arithmetic problems are not always strictly sequential and can sometimes be solved in parallel \citep{nanda2023progressmeasuresgrokkingmechanistic}. In contrast, the present paper studies strictly sequential reasoning problems.

\section{Conclusion}

Our results highlight how experimental design fundamentally shapes conclusions about LLM reasoning capabilities. Models that completely fail at latent two-hop reasoning when learning synthetic facts separately (0\% accuracy and no loss improvement over random chance) succeed when those same facts co-occur in training documents or when one fact comes from pretraining rather than fine-tuning. Conversely, frontier models' apparent success on real-world two-hop questions may reflect memorization or facts co-occurring in training data rather than genuine compositional reasoning. This sensitivity to experimental conditions cuts both ways: negative results may miss latent capabilities while positive results may overstate them. For researchers studying limitations of LLM reasoning, this suggests that both failure to elicit certain capabilities in toy experiments as well as apparent presence of those capabilities may be artifacts of evaluation methodology. 

\section*{Acknowledgments}

We are grateful to (in alphabetical order)
David Bau,
Nora Belrose,
Joseph Bloom,
Lucius Bushnaq,
Hoagy Cunningham,
Bilal Chughtai,
Leo Gao,
Mor Geva,
Peter Hase,
Marius Hobbhahn,
Benjamin Hilton,
Geoffrey Irving,
Sebastian Jaszczur,
Robert Kirk,
Bruce W. Lee,
David Lindner,
Vlad Mikulik,
Neel Nanda,
Jacob Pfau,
Fabian Roger,
Buck Shlegeris,
Jacques Thibodeau,
Misha Wagner,
Niels Warncke and
Sohee Yang
for helpful conversations and feedback on previous drafts of this paper.

\bibliography{references}
\bibliographystyle{iclr2025_conference}

\appendix

\newpage
\section{Model Versions}
\label{appendix:model_versions}

Tables below list the specific versions of all models used in our experiments.

\begin{table}[h]
\centering
\caption{Model versions used in Experiments 1, 2 and 4 (controlled experiments with synthetic facts)}
\label{tab:model_versions_synthetic}
\begin{tabular}{ll}
    \toprule
    Model Name & Identifier \\
    \midrule
    Llama 3 8B Instruct & \texttt{meta-llama/Meta-Llama-3-8B-Instruct} \\
    Qwen 2.5 7B Instruct & \texttt{Qwen/Qwen2.5-7B-Instruct} \\
    GPT-4o-mini & \texttt{gpt-4o-mini-2024-07-18} \\
    GPT-4o & \texttt{gpt-4o-2024-08-06} \\
    \bottomrule
    \end{tabular}
\end{table}

\begin{table}[h]
\centering
\caption{Model versions used in real-world knowledge evaluation}
\label{tab:model_versions_real}
\begin{tabular}{lll}
    \toprule
    Model Name & Identifier & Provider \\
    \midrule
    Claude 3 Haiku & \texttt{claude-3-haiku-20240307} & Anthropic API \\
    Claude 3 Sonnet & \texttt{claude-3-sonnet-20240229} & Anthropic API \\
    Claude 3 Opus & \texttt{claude-3-opus-20240229} & Anthropic API \\
    Claude 3.5 Haiku & \texttt{claude-3-5-haiku-20241022} & Anthropic API \\
    Claude 3.5 Sonnet & \texttt{claude-3-5-sonnet-20241022} & Anthropic API \\
    Claude 3.7 Sonnet & \texttt{claude-3-7-sonnet-20250219} & Anthropic API \\
    Llama 3.1 8B Instruct & \texttt{meta-llama/Meta-Llama-3.1-8B-Instruct-Turbo} & TogetherAI \\
    Llama 3.1 70B Instruct & \texttt{meta-llama/Meta-Llama-3.1-70B-Instruct-Turbo} & TogetherAI \\
    Llama 3.1 405B Instruct & \texttt{meta-llama/Meta-Llama-3.1-405B-Instruct-Turbo} & TogetherAI \\
    GPT-3.5-turbo & \texttt{gpt-3.5-turbo-0125} & OpenAI API \\
    GPT-4o-mini & \texttt{gpt-4o-mini-2024-07-18} & OpenAI API \\
    GPT-4o & \texttt{gpt-4o-2024-05-13} & OpenAI API \\
    GPT-4.1-nano & \texttt{gpt-4.1-nano-2025-04-14} & OpenAI API \\
    GPT-4.1-mini & \texttt{gpt-4.1-mini-2025-04-14} & OpenAI API \\
    GPT-4.1 & \texttt{gpt-4.1-2025-04-14} & OpenAI API \\
    GPT-4.5-preview & \texttt{gpt-4.5-preview-2025-02-27} & OpenAI API \\
    Qwen2.5-7B & \texttt{Qwen/Qwen2.5-7B-Instruct-Turbo} & TogetherAI \\
    Qwen2.5-72B & \texttt{Qwen/Qwen2.5-72B-Instruct-Turbo} & TogetherAI \\
    \bottomrule
    \end{tabular}
\end{table}

\newpage
\section{Experiments 1-4: Additional Details on Finetuning and Evaluation}

\subsection{Training details}
\label{app:training_details}

We include the training configuration used in fine-tuning experiments.

\begin{table}[h]
    \caption{Training hyperparameters for open-weights models in Experiments 1-3}
    \label{tab:training_hyperparams}
    \begin{tabular}{l ll}
    \toprule
     & Llama 3 8B Instruct & Qwen 2.5 7B Instruct \\
    \midrule
    Learning rate & \textbf{1e-5} & \textbf{2e-5} \\
    Optimizer & AdamW ($\beta_1{=}0.9$, $\beta_2{=}0.999$, $\epsilon{=}10^{-8}$) & AdamW ($\beta_1{=}0.9$, $\beta_2{=}0.999$, $\epsilon{=}10^{-8}$) \\
    Batch size per device & 4 & 4 \\
    Gradient accumulation & 4 & 4 \\
    Effective batch size & 16 & 16 \\
    Training precision & BFloat16 & BFloat16 \\
    Learning rate schedule & Linear decay & Linear decay \\
    Number of epochs & 1 & 1 \\
    Weight decay & 0 & 0 \\
    \bottomrule
    \end{tabular}
\end{table}

\begin{table}[h]
    \centering
    \caption{Training hyperparameters for Experiment 4}
    \label{tab:training_hyperparams_semi}
    \begin{tabular}{l l}
    \toprule
    Parameter & Value \\
    \midrule
    Learning rate & \textbf{7.5e-7} \\
    Optimizer & AdamW ($\beta_1{=}0.9$, $\beta_2{=}0.999$, $\epsilon{=}10^{-8}$) \\
    Batch size per device & \textbf{1} \\
    Gradient accumulation & \textbf{1} \\
    Effective batch size & \textbf{4} \\
    Training precision & BFloat16 \\
    Learning rate schedule & Linear decay \\
    Number of epochs & \textbf{20} \\
    Weight decay & 0 \\
    \bottomrule
    \end{tabular}
\end{table}
All open-weights experiments were conducted using 4xA100 80GB GPUs with PyTorch's fully-sharded data parallel (FSDP) training. We used the Center for AI Safety's compute cluster for training.

\begin{table}[h]
    \centering
    \caption{Training hyperparameters for proprietary models}
    \label{tab:training_hyperparams_proprietary}
    \begin{tabular}{l ll}
    \toprule
    Parameter & GPT-4o-mini & GPT-4o \\
    \midrule
    Learning rate multiplier & 6.0 & 6.0 \\
    Batch size & 45 & 45 \\
    Number of epochs & 1 & 1 \\
    \bottomrule
    \end{tabular}
\end{table}

\subsection{Evaluation Details}
\label{appendix:synthetic_eval}

For all finetuning experiments (Experiments 1-4), we evaluate models using temperature 0 for deterministic generation. For evaluating knowledge of atomic facts, we use the same questions as those seen during training. For two-hop questions, we use questions not seen during training (the undemonstrated set). In both CoT and no-CoT evaluation, we use 20 few-shot examples, randomly sampled from the training dataset. These examples remain fixed throughout evaluation (i.e., the same examples are used for all test samples).

For no-CoT evaluation of open-weights models, we restrict the outputs to the set of test set answer options by only allowing tokens corresponding to valid answers (plus end-of-sequence tokens). If the response contains the correct answer ($e_3$), we additionally check that the response does not contain the intermediate step of reasoning ($e_2$).

\paragraph{System prompts} We use the same system prompts during training and evaluation:

For evaluation \textit{with CoT}, we use the following \textbf{system message}:

\begin{tcolorbox}[breakable,pad at break*=1mm,colback=black!5!white,colframe=black!75!black,fontupper=\small\ttfamily,fonttitle=\bfseries]
\begin{lstlisting}
You will be given questions about fictional characters from the "Spouses" saga.

Answer the following questions step by step.
\end{lstlisting}
\end{tcolorbox}

For evaluation \textit{without CoT}, we use the following \textbf{system message}:

\begin{tcolorbox}[breakable,pad at break*=1mm,colback=black!5!white,colframe=black!75!black,fontupper=\small\ttfamily,fonttitle=\bfseries]
\begin{lstlisting}
You will be given questions about fictional characters from the "Spouses" saga.

Answer the following questions directly, without any other text before or after your answer.
\end{lstlisting}
\end{tcolorbox}

\newpage
\section{Details on Real-World Knowledge Experiments}
\label{appendix:real_world_eval}

To evaluate two-hop reasoning capabilities on real-world knowledge, we use the dataset from \cite{biran2024hoppinglateexploringlimitations}. While positive performance on real-world facts cannot be confidently attributed to two-hop reasoning (due to potential memorization and shortcuts), examining performance patterns across models and question types can provide additional context for understanding two-hop reasoning capabilities.

We evaluate 19 frontier models, including models from the Claude 3 \citep{anthropic2024claude3} and Claude 4 family \citep{anthropic2025claude4systemcard}, the Llama 3.1 family \citep{dubey2024llama3herdmodels}, the GPT family (such as GPT-3.5-turbo, GPT-4o, GPT-4o-mini, and GPT-4.1 variants) \citep{openai2024_4o}, and the Qwen 2.5 family \citep{qwen2025qwen25technicalreport}. We exclude reasoning models (e.g., OpenAI's O-series) from our evaluations because it is not always possible to suppress their CoT. We evaluate our selected models on a dataset of two-hop questions involving real-world facts compiled by \cite{biran2024hoppinglateexploringlimitations}. For each question, we evaluate models both with and without CoT. Because we focus on evaluating two-hop reasoning, we only report performance on questions where models successfully answered single-hop questions about underlying facts.

We exclude examples that clearly do not require two-hop reasoning, following \cite{biran2024hoppinglateexploringlimitations} in filtering out cases where models can answer two-hop questions without performing both hops. Specifically, we evaluate models on partial two-hop questions that omit either $e_1$ or the first relationship, and exclude cases where models succeed on these baseline conditions.

For each model and question type combination, we evaluate on $n=1500$ random samples (of a total of $71569$) from the dataset with temperature set to 0. In the results, we only report performance on those of these samples where a model successfully answered both underlying one-hop questions. This ensures that failures on two-hop questions can be attributed to reasoning limitations rather than lack of knowledge about the underlying facts.

\subsection{Evaluation Prompts}

For the first-hop and second-hop questions, as well as the two-hop questions without CoT, we use the following \textbf{system message}:

\begin{tcolorbox}[breakable,pad at break*=1mm,colback=black!5!white,colframe=black!75!black,fontupper=\small\ttfamily,fonttitle=\bfseries]
\begin{lstlisting}
Answer the following question with a single noun phrase, without any reasoning. There is always an answer. If the answer is ambiguous, use your best guess.
\end{lstlisting} 
\end{tcolorbox}

For the two-hop questions with CoT, we use the following \textbf{system message}:

\begin{tcolorbox}[breakable,pad at break*=1mm,colback=black!5!white,colframe=black!75!black,fontupper=\small\ttfamily,fonttitle=\bfseries]
\begin{lstlisting}
Answer the following question step by step. There is always an answer. If the answer is ambiguous, use your best guess.
\end{lstlisting} 
\end{tcolorbox}

An example \textbf{user message} for evaluation on a two-hop question without CoT is:

\begin{tcolorbox}[breakable,pad at break*=1mm,colback=black!5!white,colframe=black!75!black,fontupper=\small\ttfamily,fonttitle=\bfseries]
\begin{lstlisting}
Who is the founder of the record label of The Centurians?
\end{lstlisting}
\end{tcolorbox}

\subsection{Grading Prompt}

We use UK AISI's Inspect's \citep{UK_AI_Safety_Institute_Inspect_AI_Framework_2024} model-grading to evaluate responses using GPT-4o as a judge. We use the following grading prompt template:

\begin{tcolorbox}[breakable,pad at break*=1mm,colback=black!5!white,colframe=black!75!black,fontupper=\small\ttfamily,fonttitle=\bfseries]
\begin{lstlisting}
You are comparing a submitted answer to an expert answer on a given question. Here is the data:

[BEGIN DATA]
************
[Question]: {question}
************
[Expert]: {criterion}
************
[Submission]: {answer}
************
[END DATA]

Compare the factual content of the submitted answer with the expert answer. Ignore any differences in style, grammar, or punctuation.

Consider the expert answer absolutely true - even if you think it is wrong.

Does the submission contain the content in the expert answer?

{instructions}
\end{lstlisting}
\end{tcolorbox}

\subsection{Dataset Statistics}

\begin{figure}[h]
    \centering
    \includegraphics[width=.7\textwidth]{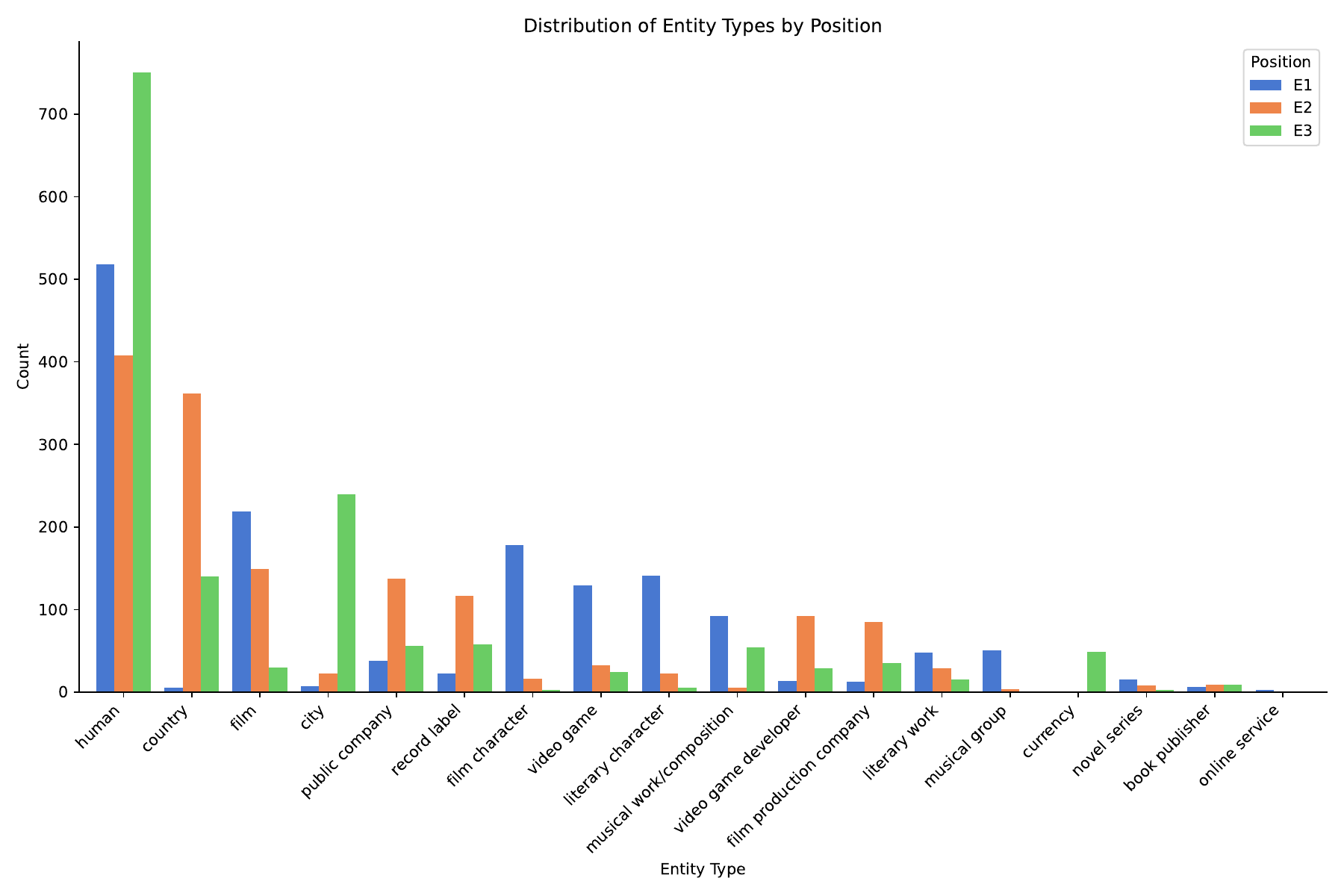}
    \caption{Distribution of entity types by position in the triplet. Note how different positions tend to contain different types of entities: $e_1$ is often a human or creative work, $e_2$ frequently represents organizations or locations, and $e_3$ is dominated by humans and locations.}
    \label{fig:entity_distributions}
\end{figure}

\begin{figure}[h]
    \centering
    \includegraphics[width=.7\textwidth]{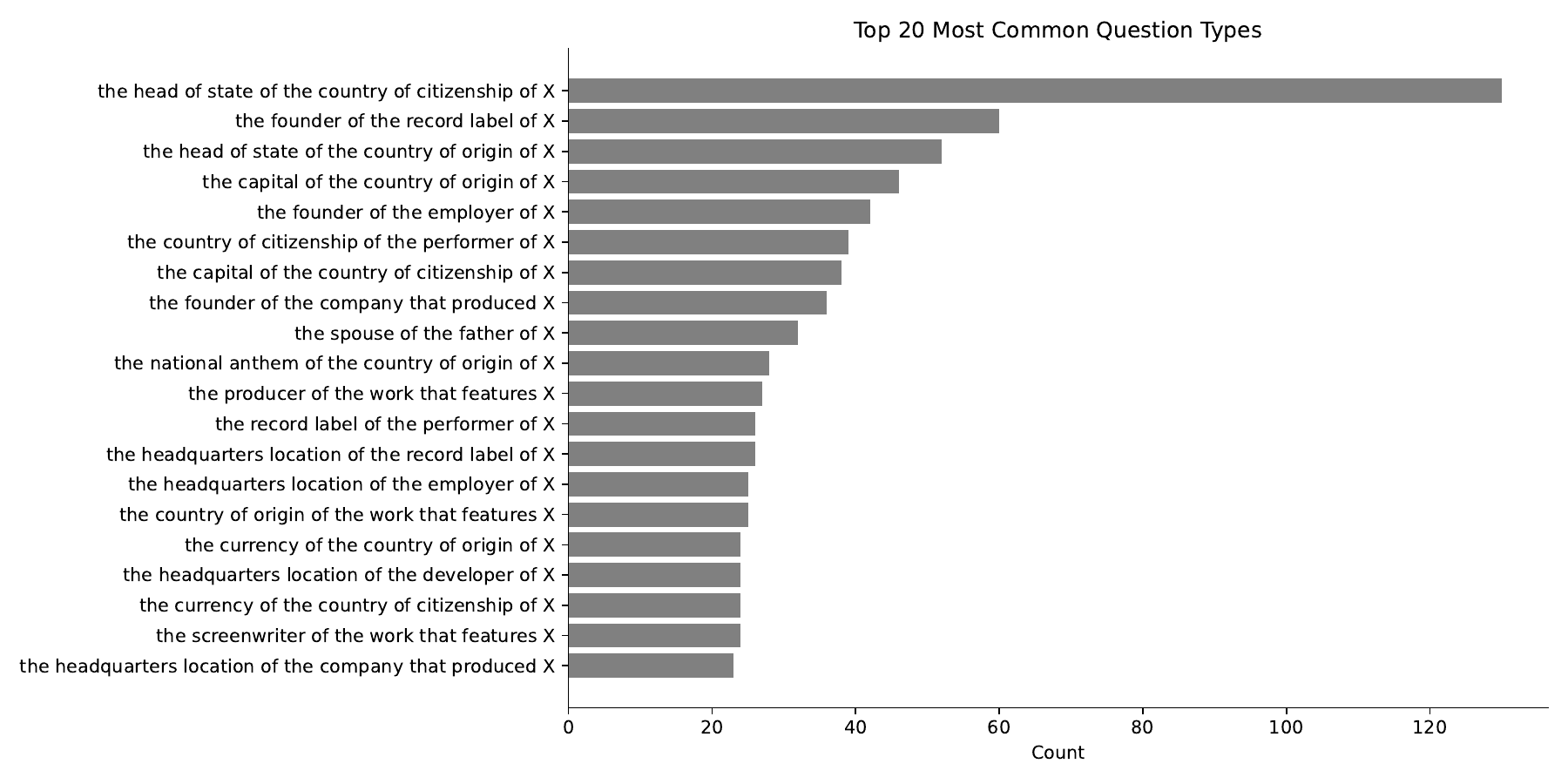}
    \caption{Most common question types in our dataset. Each type represents a specific pattern of two-hop reasoning, combining two relationship templates (e.g., "the country of citizenship of X" followed by "the head of state of Y").}
    \label{fig:question_types}
\end{figure}

Figure~\ref{fig:entity_distributions} shows the distribution of entity types across the three positions in our triplets ($e_1$, $e_2$, $e_3$) across top 20 types of entities. Figure~\ref{fig:question_types} shows the most common question categories.

\newpage
\section{Experiment 1 Ablation: Impact of Training Data Mixture}
\label{sec:ablation_data_mixtures}

To understand how different components of our training data mixture contribute to two-hop reasoning capabilities, we conduct an ablation study with three variants of the training setup:

\begin{enumerate}
    \item \textbf{Full mixture} (our standard setup): Training on atomic facts and both two-hop CoT and no-CoT QA pairs
    \item \textbf{No-CoT mixture}: Training on atomic facts and only two-hop no-CoT QA pairs
    \item \textbf{Atomic-only mixture}: Training exclusively on atomic facts
\end{enumerate}

\begin{figure}[t]
    \begin{subfigure}[c]{1\textwidth}
    \centering
    \small{
        \line{mixture_ablation_main} Full mixture \quad
        \line{mixture_ablation_nocot} No-CoT mixture \quad
        \line{mixture_ablation_atomic} Atomic-only \\
        \vspace{2mm}
        \thinline{solid,gray} Loss on correct responses \quad
        \thinline{dashed,gray} Loss on random responses
        \vspace{1mm}
    }
    \end{subfigure}
    \centering
    \begin{subfigure}[b]{0.42\textwidth}
        \centering
        \includegraphics[width=\textwidth]{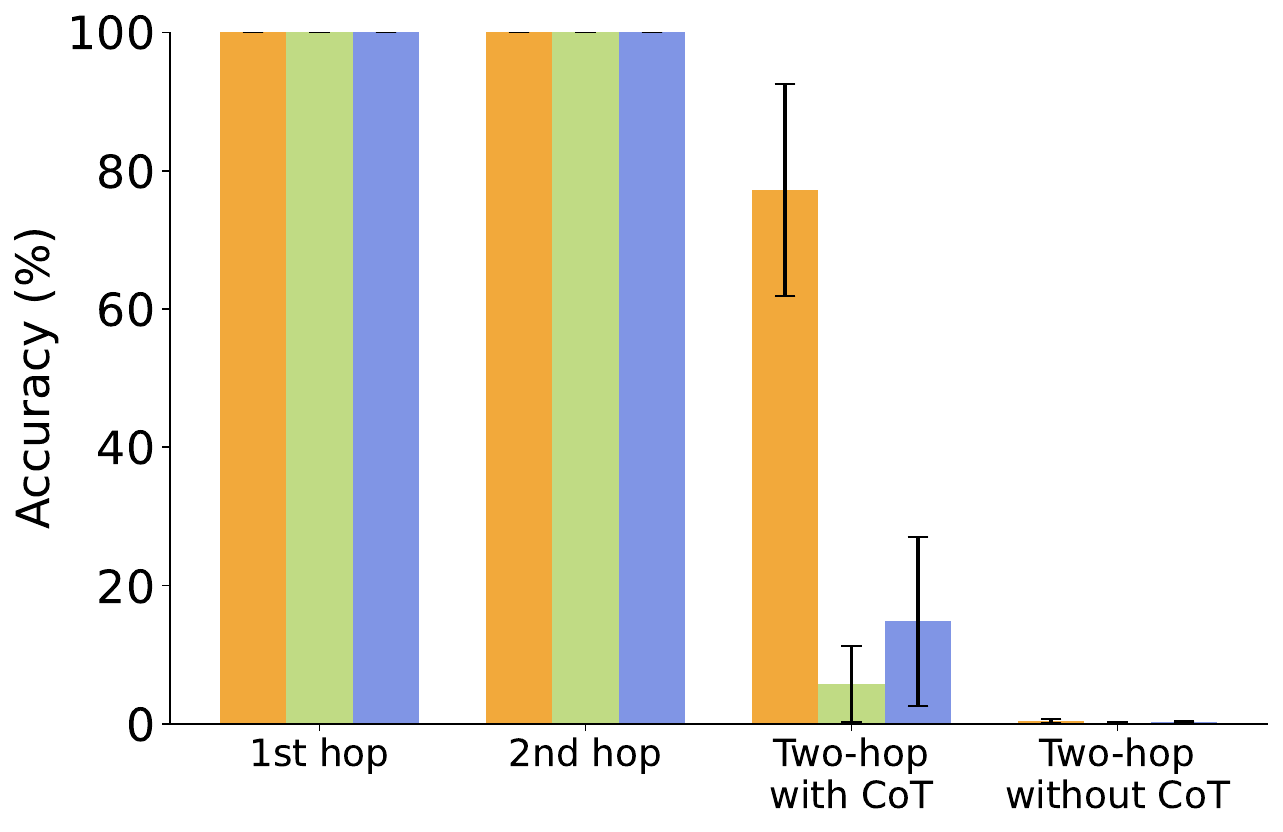}
        \caption{Accuracy across metrics}
    \end{subfigure}
    \begin{subfigure}[b]{0.28\textwidth}
        \centering
        \includegraphics[width=\textwidth]{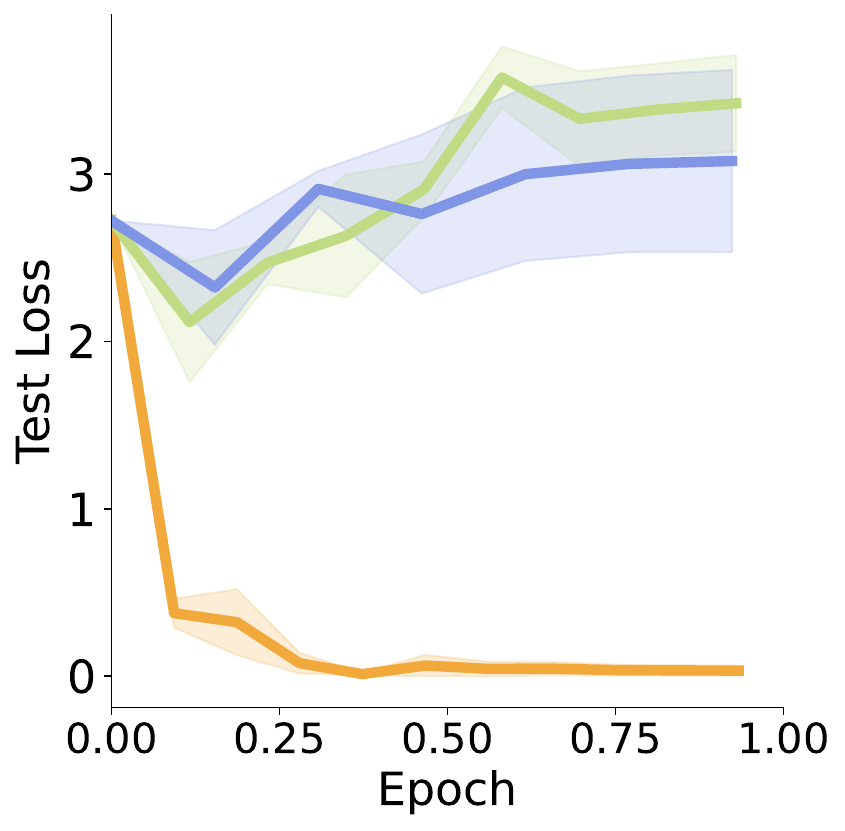}
        \caption{Two-hop CoT loss}
    \end{subfigure}
    \begin{subfigure}[b]{0.28\textwidth}
        \centering
        \includegraphics[width=\textwidth]{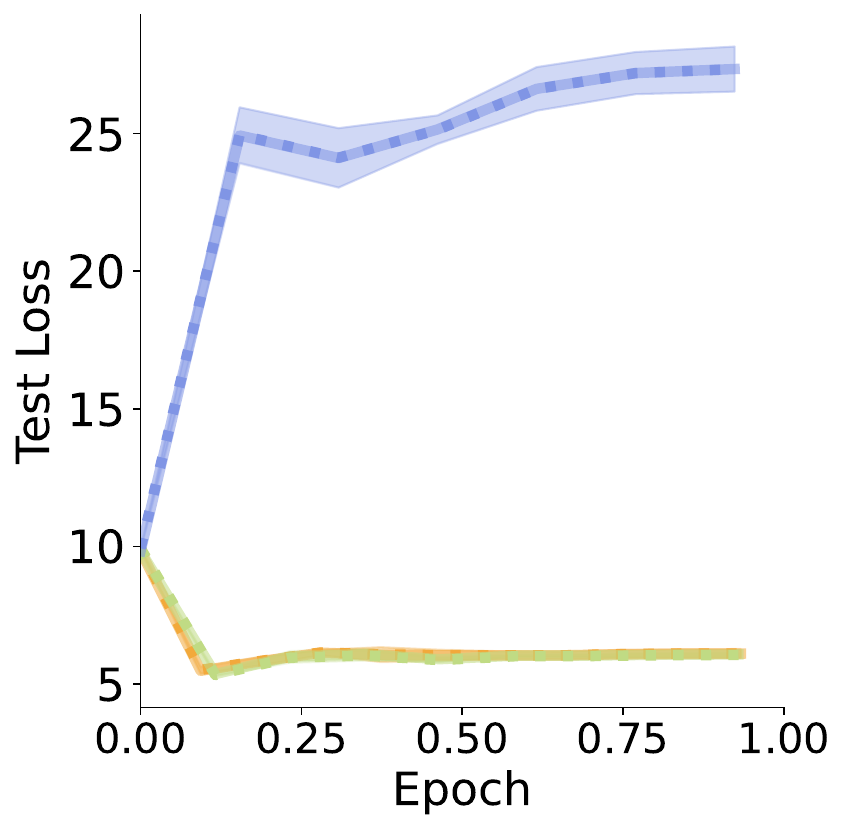}
        \caption{Two-hop no-CoT loss}
    \end{subfigure}
    \caption{Performance comparison of Llama 3 8B Instruct across ablated training data mixtures. \textbf{Left:} Accuracy on one-hop facts (both first and second hop) and two-hop questions (both with and without CoT). \textbf{Middle:} Test loss on two-hop CoT questions. \textbf{Right:} Test loss on two-hop no-CoT questions compared to loss on randomly shuffled responses. The atomic-only and no-CoT mixtures achieve similar performance to the full mixture on one-hop questions but show significantly worse performance on two-hop CoT questions, as evidenced by both accuracy and loss. Two-hop no-CoT performance remains at chance level across all mixtures.}
    \label{fig:ablation_results}
\end{figure}

This ablation shows that removing CoT training examples significantly impairs two-hop CoT performance while maintaining one-hop accuracy. Two-hop no-CoT performance remains at chance level regardless of the training mixture.

\newpage
\section{Experiment 3 Additional Results: Effect of Distractors on Same-Document Two-Hop Reasoning}
\label{sec:appendix_distractors}

In Section~\ref{sec:experiment_3}, we investigate how models perform when the two relevant facts appear in the same document in subsequent sentences. However, in pretraining, facts in the same document might be separated by other content rather than directly following each other. To approximate this, we investigate how models perform when the two relevant facts in the training documents are separated by minimal distractor information. We explore two types of distractors:

\begin{enumerate}
    \item Semantically related but non-essential information about the entities, for example:
    \begin{quote}
    ``Russ shares a marital bond with Hay. \textbf{Russ is 1m 75cm tall and loves bouldering. Hay is slightly higher at 1m 77cm, and they often go climbing together.} The city which saw the birth of Hay is Showing.''
    \end{quote}

    \item Facts about $N$ other random $\langle e_1, e_2, e_3 \rangle$ triplets. For example, given a user message:
    \begin{quote}
    ``Tell me who the following people are married to: Virgin, Russ, View, Just. Then tell me where those spouses were born.''
    \end{quote}
    the assistant message could contain facts about $N=3$ random other triplets:
    \begin{quote}
    ``\textbf{Virgin is wedded to Ha.} Russ calls Hay their spouse. \textbf{View is united with Walking in wedded bliss. Just's marriage partner is Knight. \newline\newline Ha was brought into existence in Active.} Hay entered the world in Showing. \textbf{The beginning of Walking's life was marked by the city of Nobody. The start of Knight's life journey is marked by the city of Crystal.}''
    \end{quote}
\end{enumerate}

The bolded text serves as a distractor between the $\langle e_1, e_2 \rangle$ fact and the $\langle e_2, e_3 \rangle$ fact. The boldness is used here for illustration only.

\begin{figure}[t]
    \begin{subfigure}[c]{1\textwidth}
        \centering
        \scriptsize{
            \thinline{solid,gray} Loss on correct responses \quad
            \thinline{dashed,gray} Loss on random responses
        }
    \end{subfigure}
    \centering
    \begin{subfigure}[c]{0.57\textwidth}
        \centering
        \includegraphics[width=\textwidth]{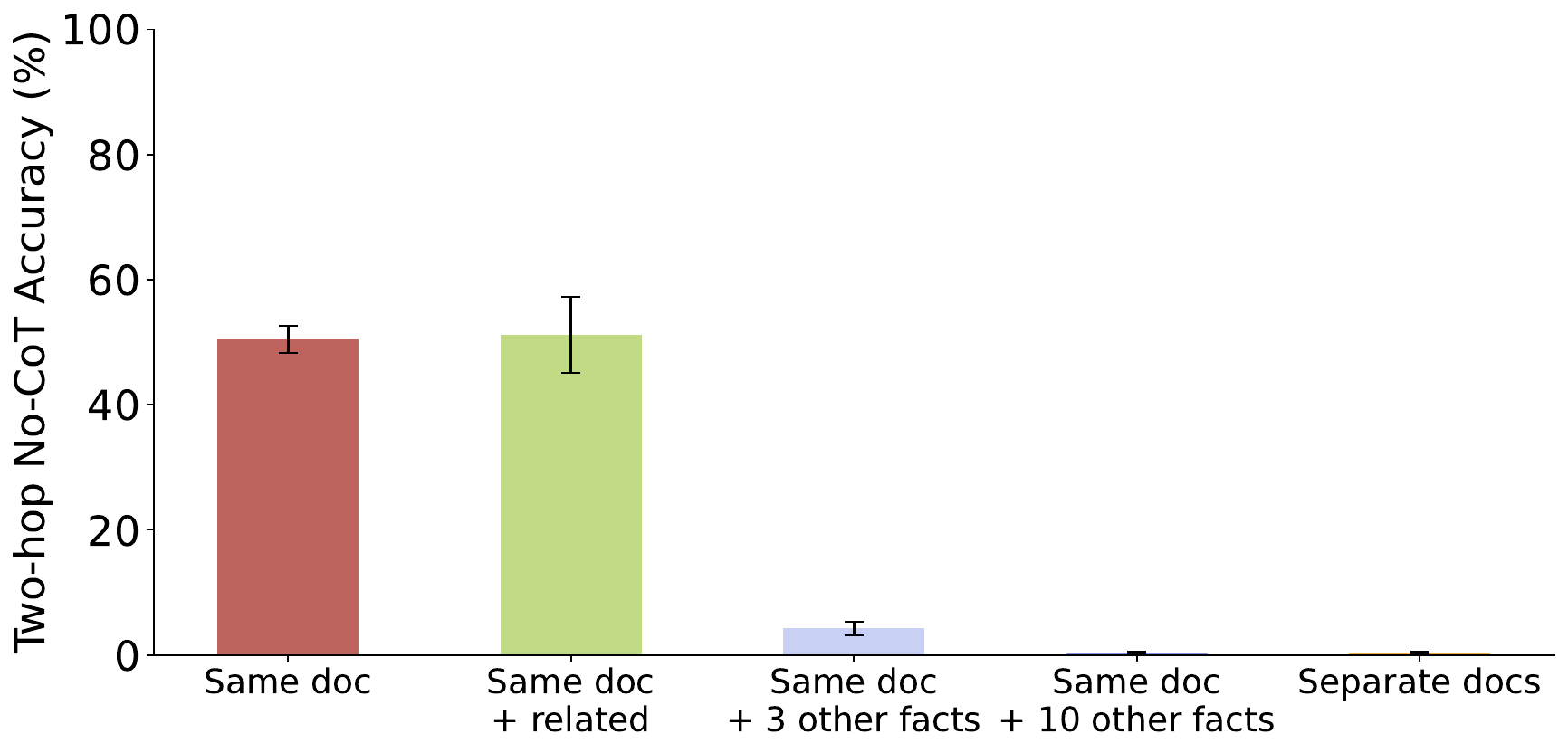}
    \end{subfigure}
    \begin{subfigure}[c]{0.41\textwidth}
        \centering
        \includegraphics[width=\textwidth]{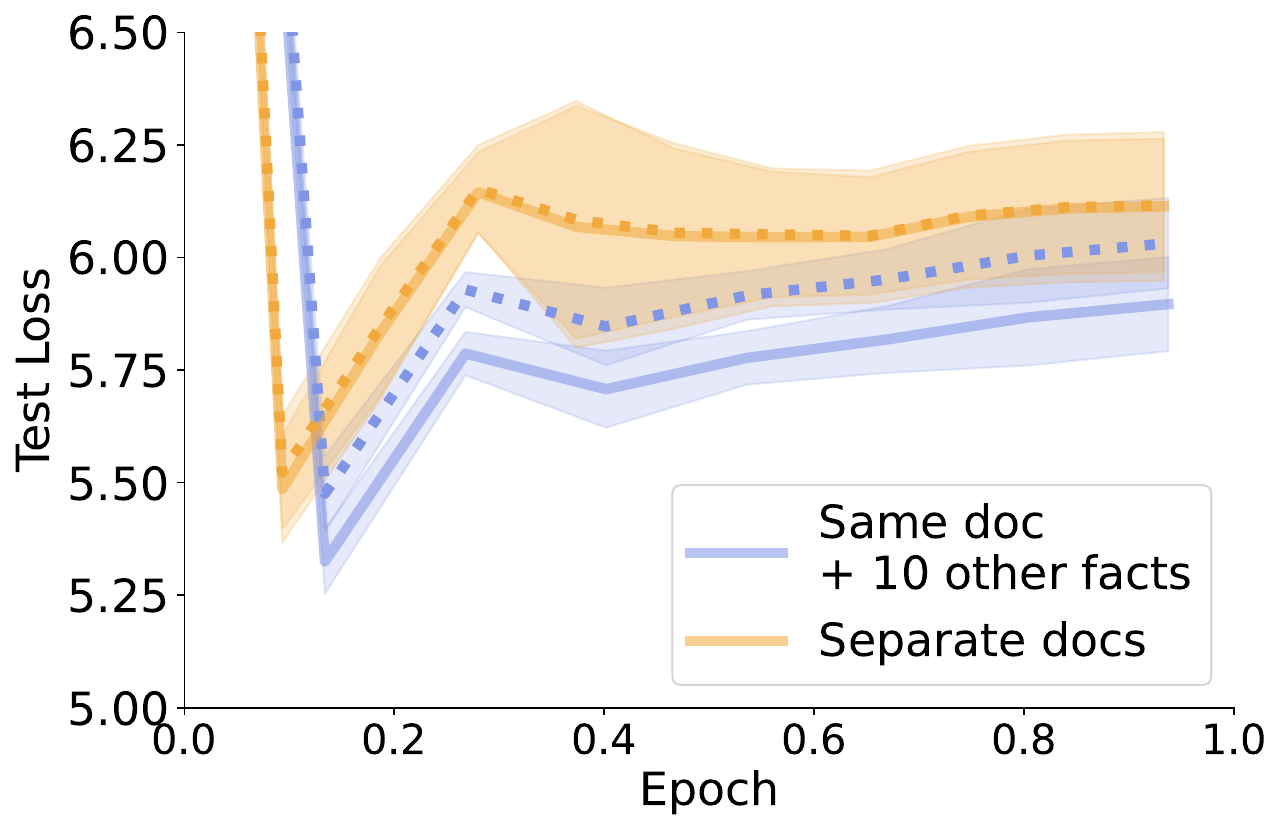}
    \end{subfigure}
    \caption{Latent two-hop reasoning ability of Llama 3 8B Instruct, measured on fictional facts learned in the same document setting, across different distractor settings.  \textbf{Left:} Adding semantically related distractors maintains similar performance to the pure same-document condition, while adding facts about other triplets leads to progressively lower accuracy as the number of distractor triplets increases. \textbf{Right:} Despite chance-level accuracy in the setting with facts about 10 other triplets, Llama still achieves better-than-chance loss, in contrast to the separate-document setting.}
    \label{fig:distractor_results}
\end{figure}

Figure~\ref{fig:distractor_results} shows the performance comparison across different conditions for Llama 3 8B Instruct. Adding semantically related distractors between facts maintains similar performance to the pure same-document condition, while adding facts about other triplets leads to lower but still above-chance accuracy with 3 distractors (see Figure~\ref{fig:distractor_results}). Increasing the number of distractor triplets to 10 reduces accuracy to chance level. However, even with 10 distractors, the test loss remains below that of random predictions, suggesting the model still learns some two-hop reasoning from the training data, in contrast to the separate-document setup. 

While our distractors are minimal compared to the variety of content that might separate facts in pretraining, these results suggest that models can perform latent reasoning even when facts are not directly adjacent in a document.

\newpage
\section{Experiment 4 Additional Results: Semi-synthetic two-hop reasoning results}
\label{appendix:additional_semi_synthetic_results}

\begin{figure}[h]
    \centering
    \begin{subfigure}[b]{0.48\textwidth}
        \centering
        \includegraphics[width=\textwidth]{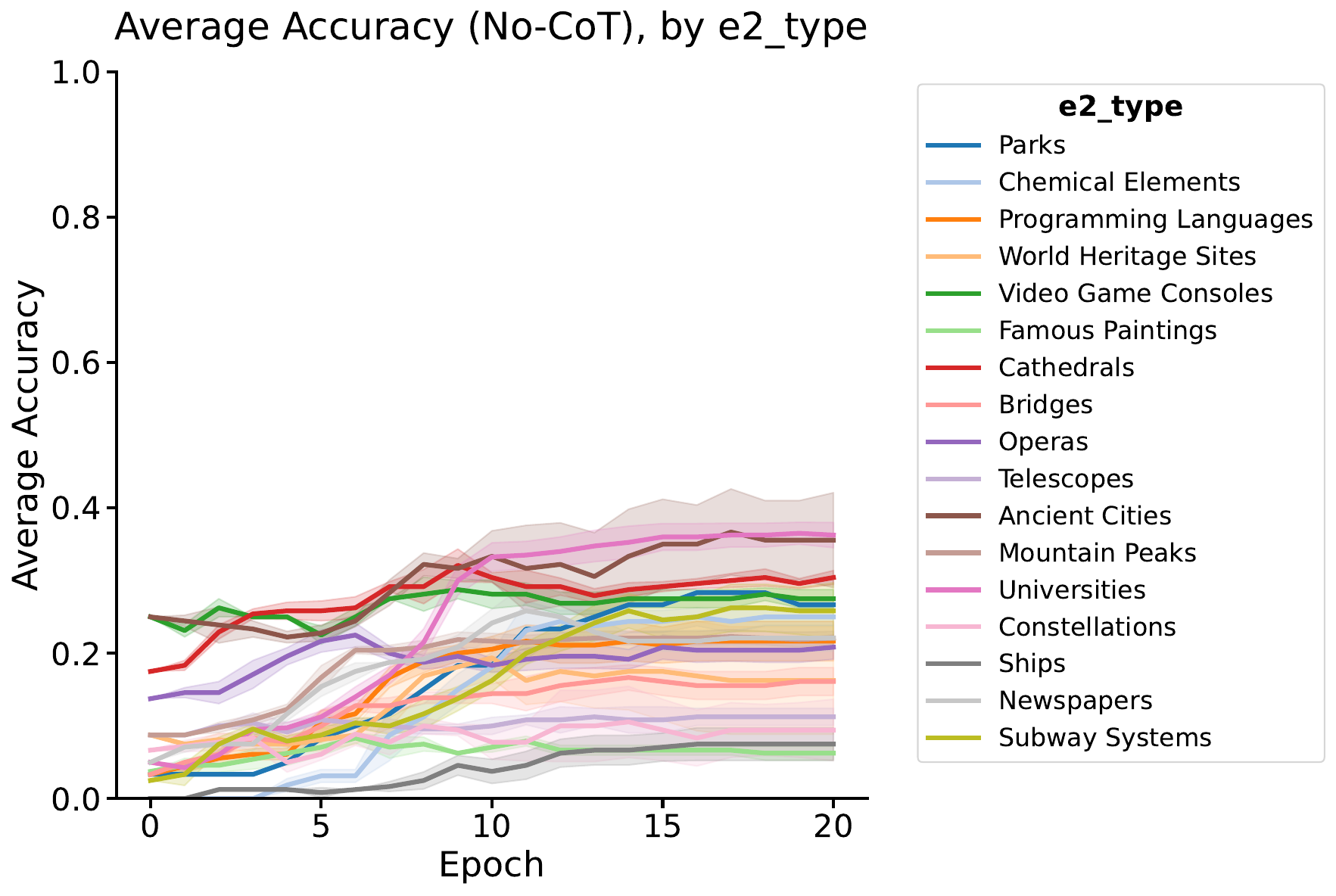}
        \caption{No-CoT accuracy by $e_2$ type}
    \end{subfigure}
    \hfill
    \begin{subfigure}[b]{0.48\textwidth}
        \centering
        \includegraphics[width=\textwidth]{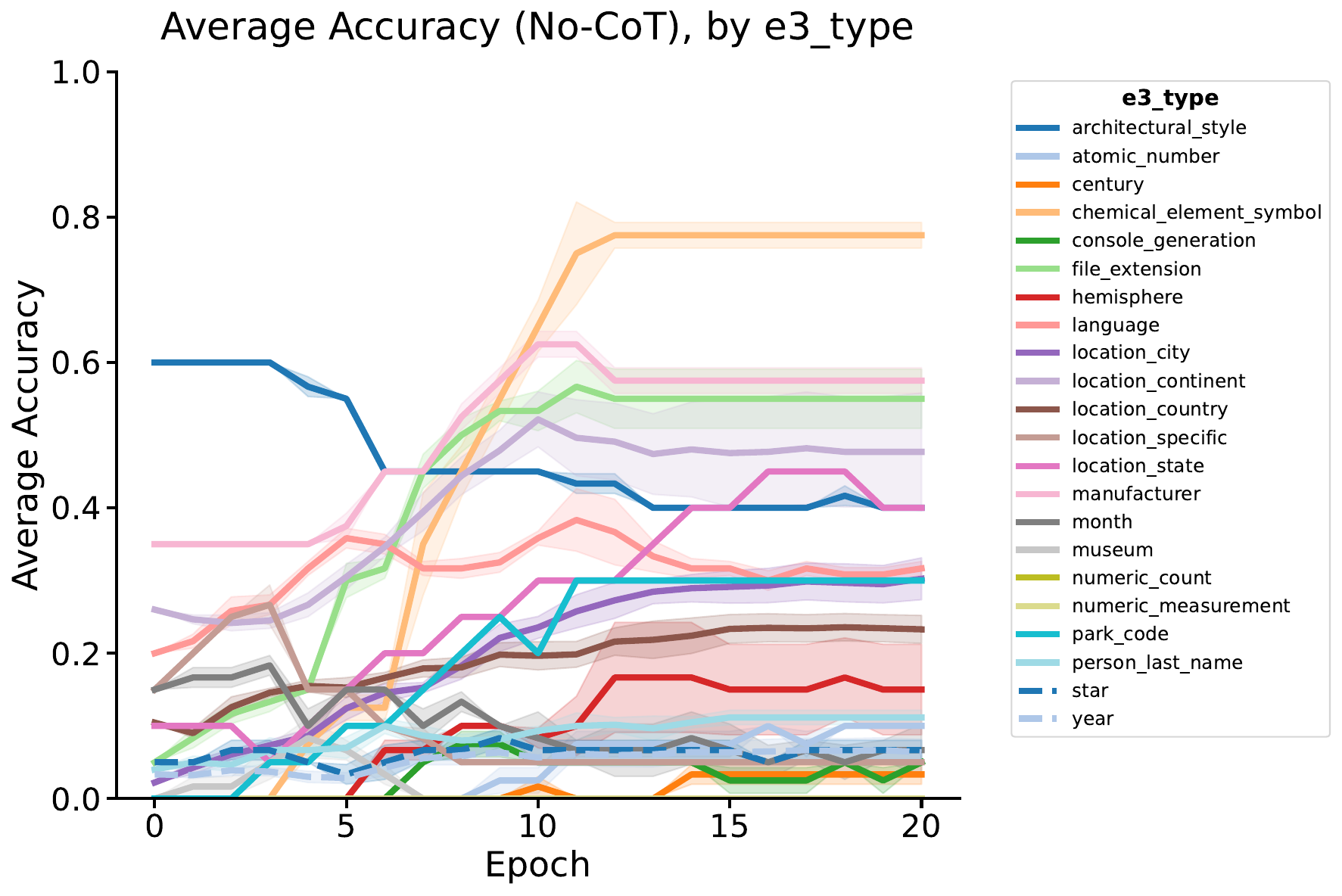}
        \caption{No-CoT accuracy by $e_3$ type}
    \end{subfigure}

    \vspace{0.8em}

    \begin{subfigure}[b]{0.48\textwidth}
        \centering
        \includegraphics[width=\textwidth]{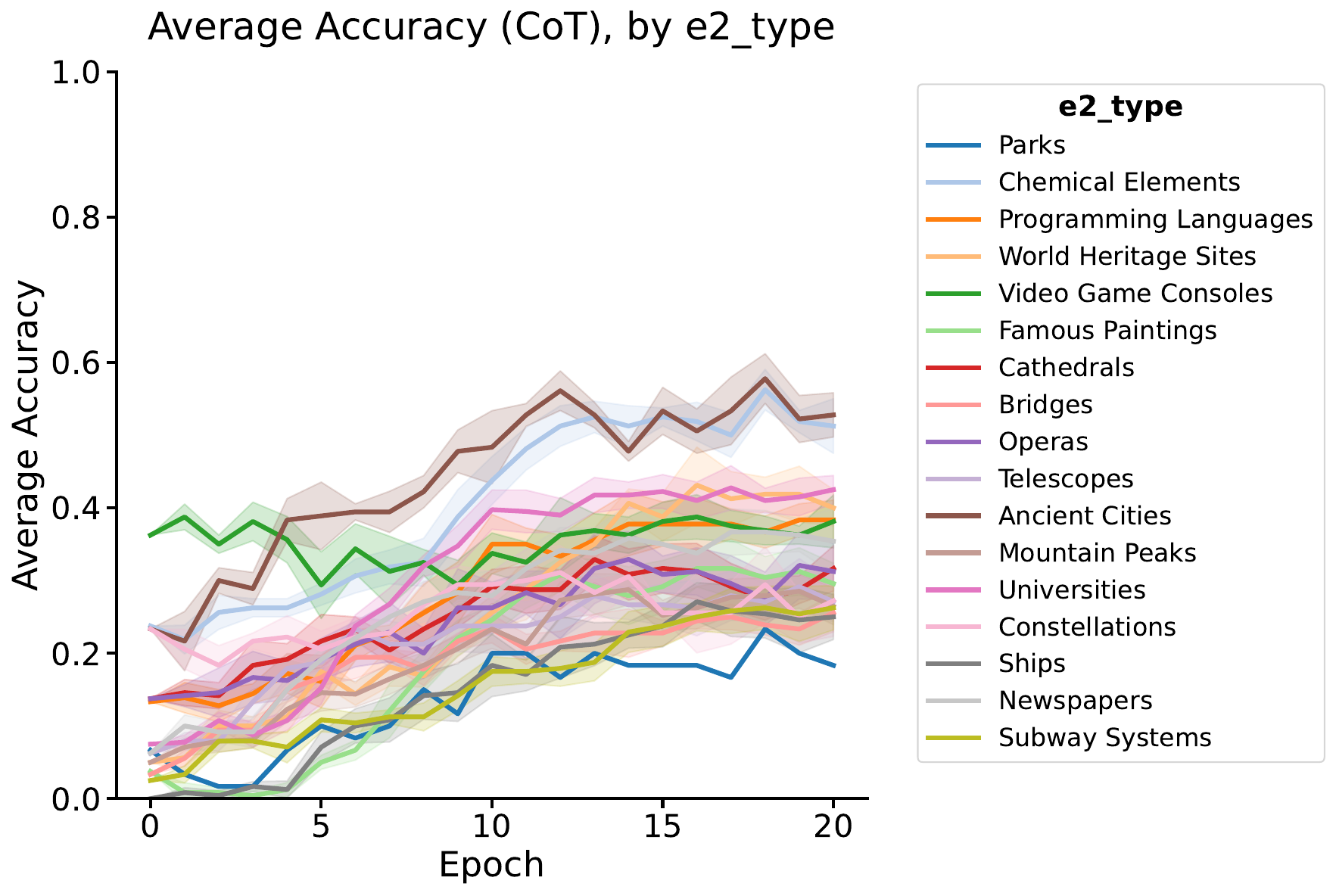}
        \caption{CoT accuracy by $e_2$ type}
    \end{subfigure}
    \hfill
    \begin{subfigure}[b]{0.48\textwidth}
        \centering
        \includegraphics[width=\textwidth]{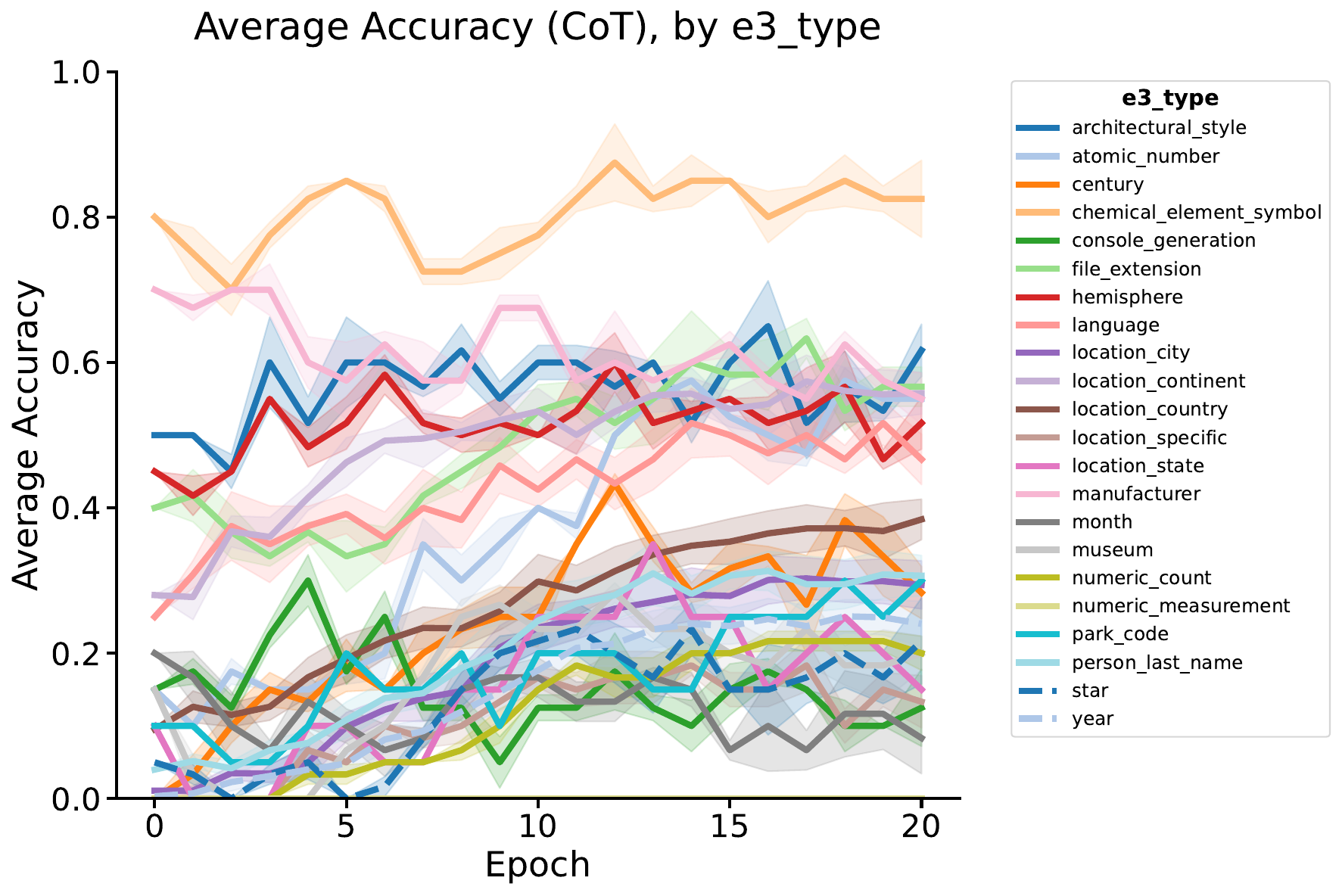}
        \caption{CoT accuracy by $e_3$ type}
    \end{subfigure}

    \vspace{0.8em}

    \begin{subfigure}[b]{0.48\textwidth}
        \centering
        \includegraphics[width=\textwidth]{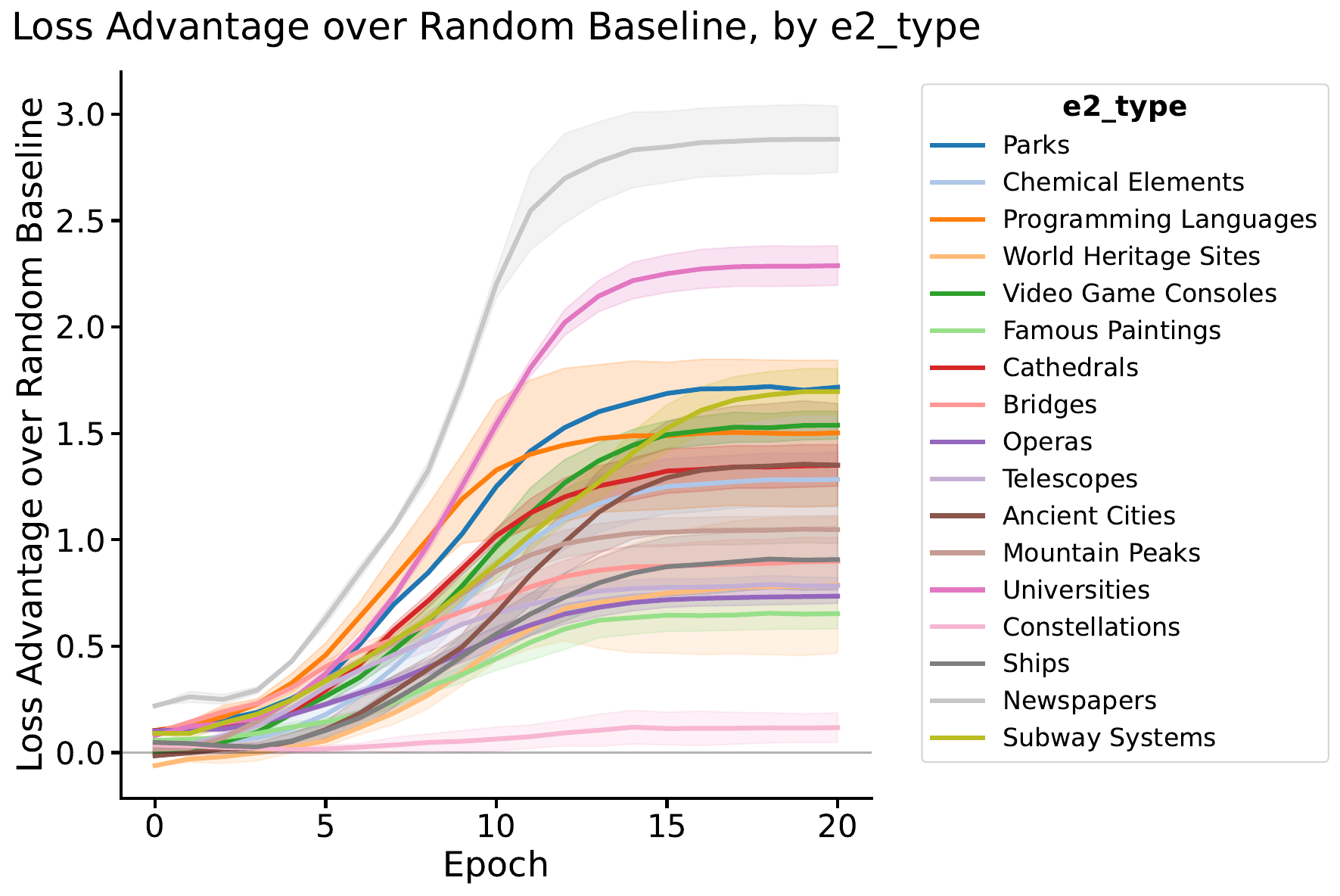}
        \caption{No-CoT loss advantage by $e_2$ type}
    \end{subfigure}
    \hfill
    \begin{subfigure}[b]{0.48\textwidth}
        \centering
        \includegraphics[width=\textwidth]{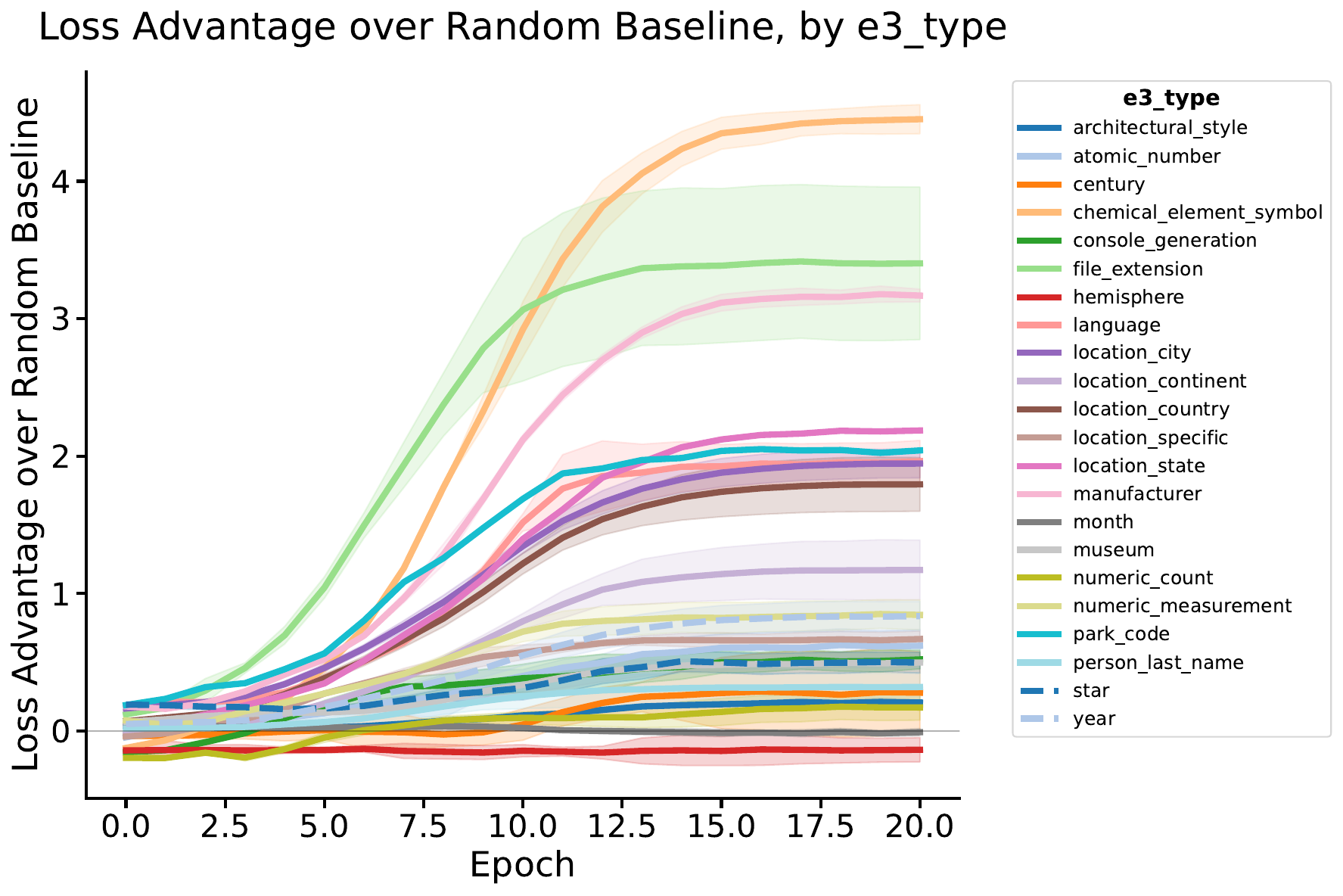}
        \caption{No-CoT loss advantage by $e_3$ type}
    \end{subfigure}

    \caption{\textbf{Accuracy and loss vary substantially across entity categories.} Each subplot groups performance by the type of the bridge entity ($e_2$) or the target entity ($e_3$). $e_3$ types plotted here are aggregates of the actual $e_3$ types used in individual datasets (for example, ``year'' includes ``year of discovery'', ``year of completion'', etc,). ``Loss advantage'' denotes the difference between the model's empirical cross-entropy loss on correct answer tokens and the random-baseline loss (loss on randomly shuffled answers from the same test set); positive values indicate the model is better than chance.}
    \label{fig:semi_synthetic_detailed}
\end{figure}

\newpage
\section{Additional Results: Per-category two-hop results on real facts}
\label{appendix:additional_frequency_plots}

\begin{figure}[h]
    \begin{subfigure}[c]{1\textwidth}
    \centering
    \tiny{
        \line{fig1_twohop_cot} With CoT \quad
        \line{fig1_twohop_nocot} Without CoT
    }
    \end{subfigure}
    \begin{subfigure}[b]{0.23\textwidth}
        \centering
        \includegraphics[width=\textwidth]{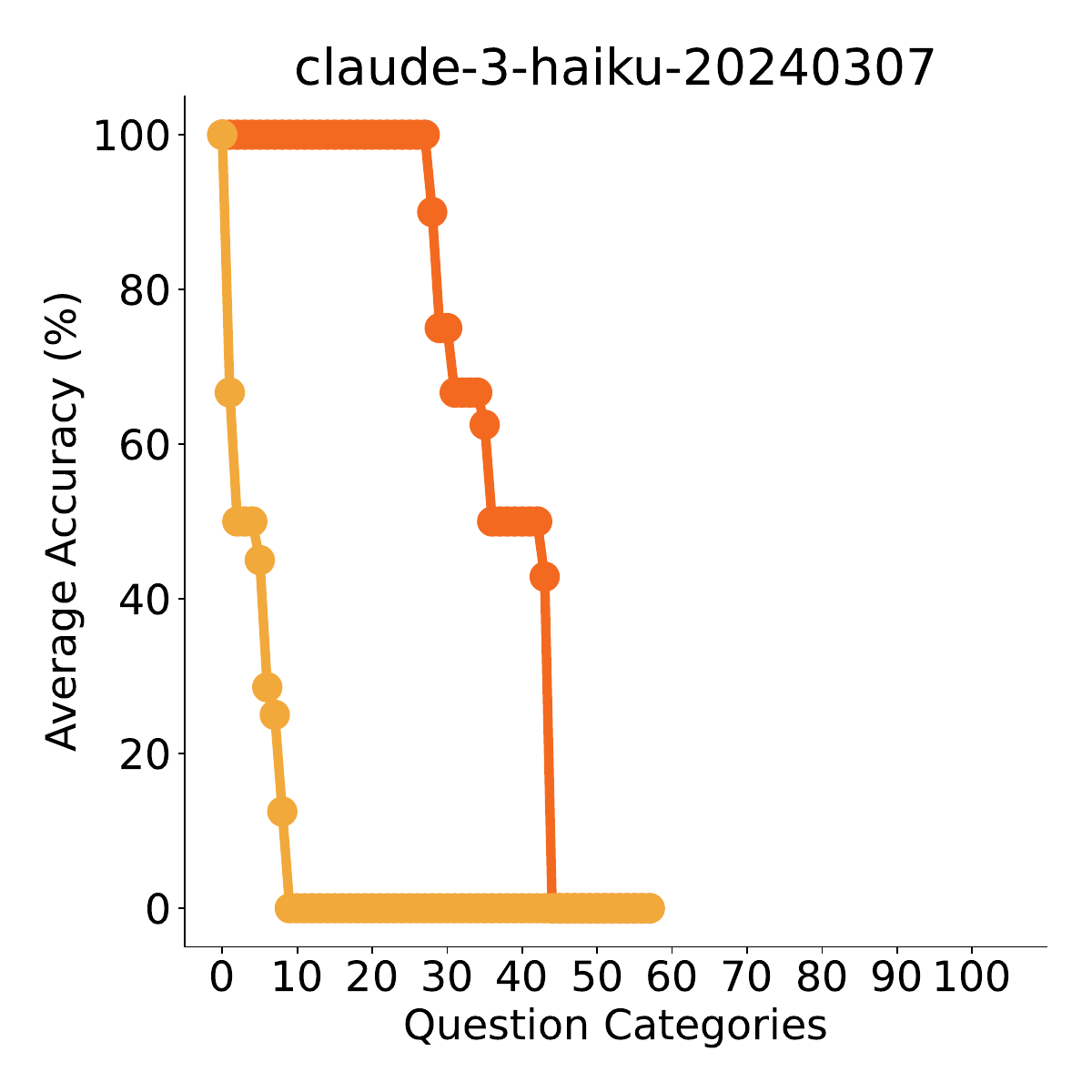}
    \end{subfigure}
    \begin{subfigure}[b]{0.23\textwidth}
        \centering
        \includegraphics[width=\textwidth]{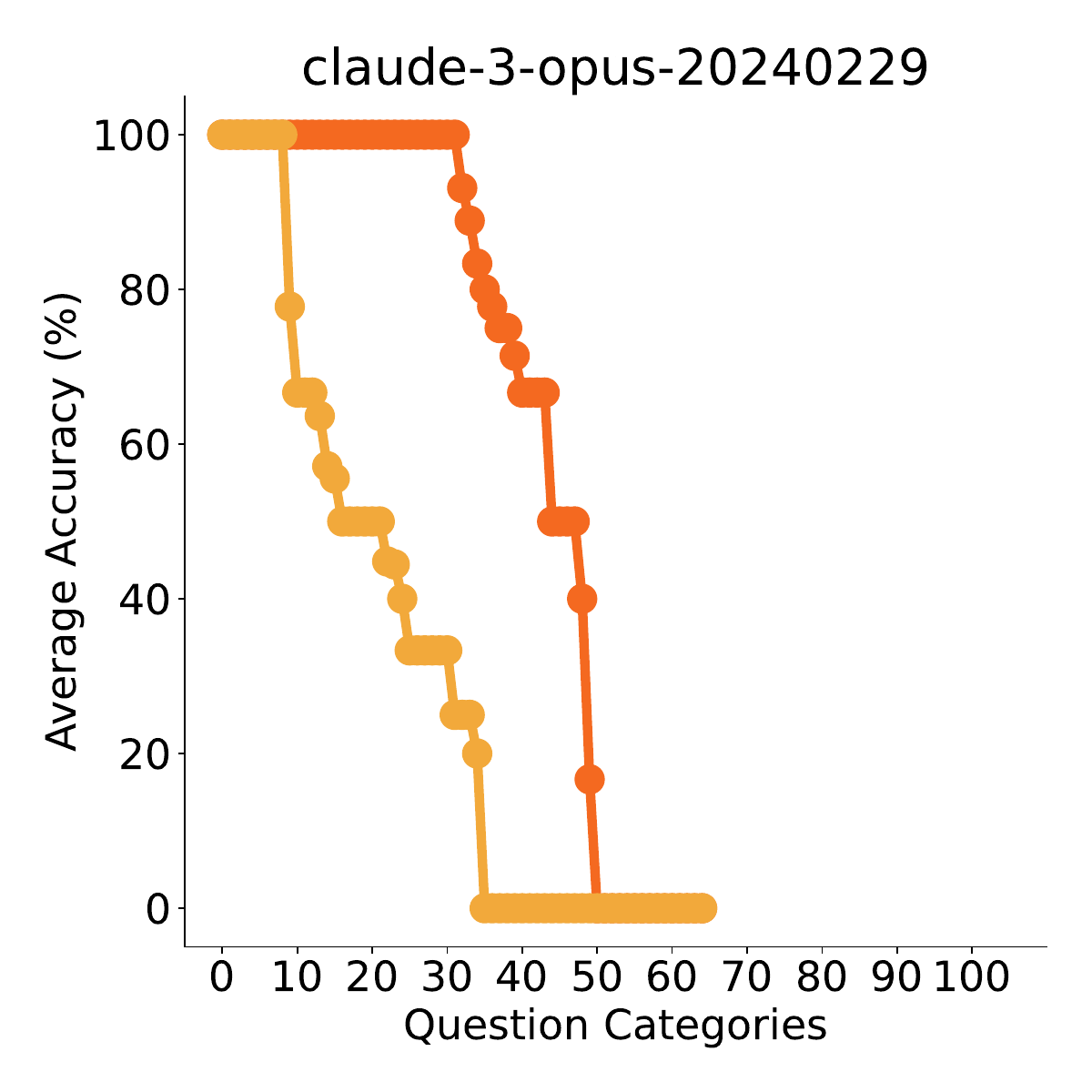}
    \end{subfigure}
    \begin{subfigure}[b]{0.23\textwidth}
        \centering
        \includegraphics[width=\textwidth]{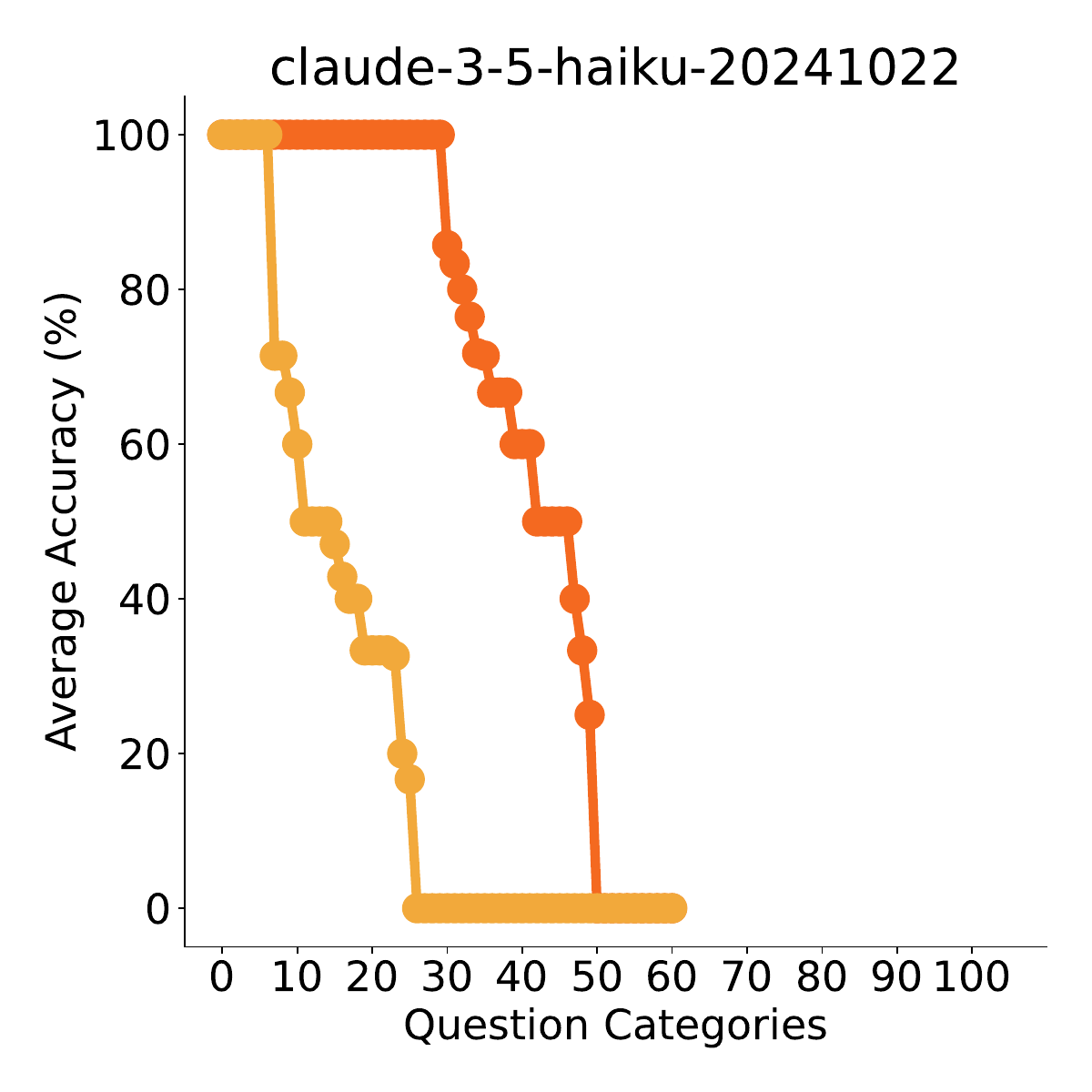}
    \end{subfigure}
    \begin{subfigure}[b]{0.23\textwidth}
        \centering
        \includegraphics[width=\textwidth]{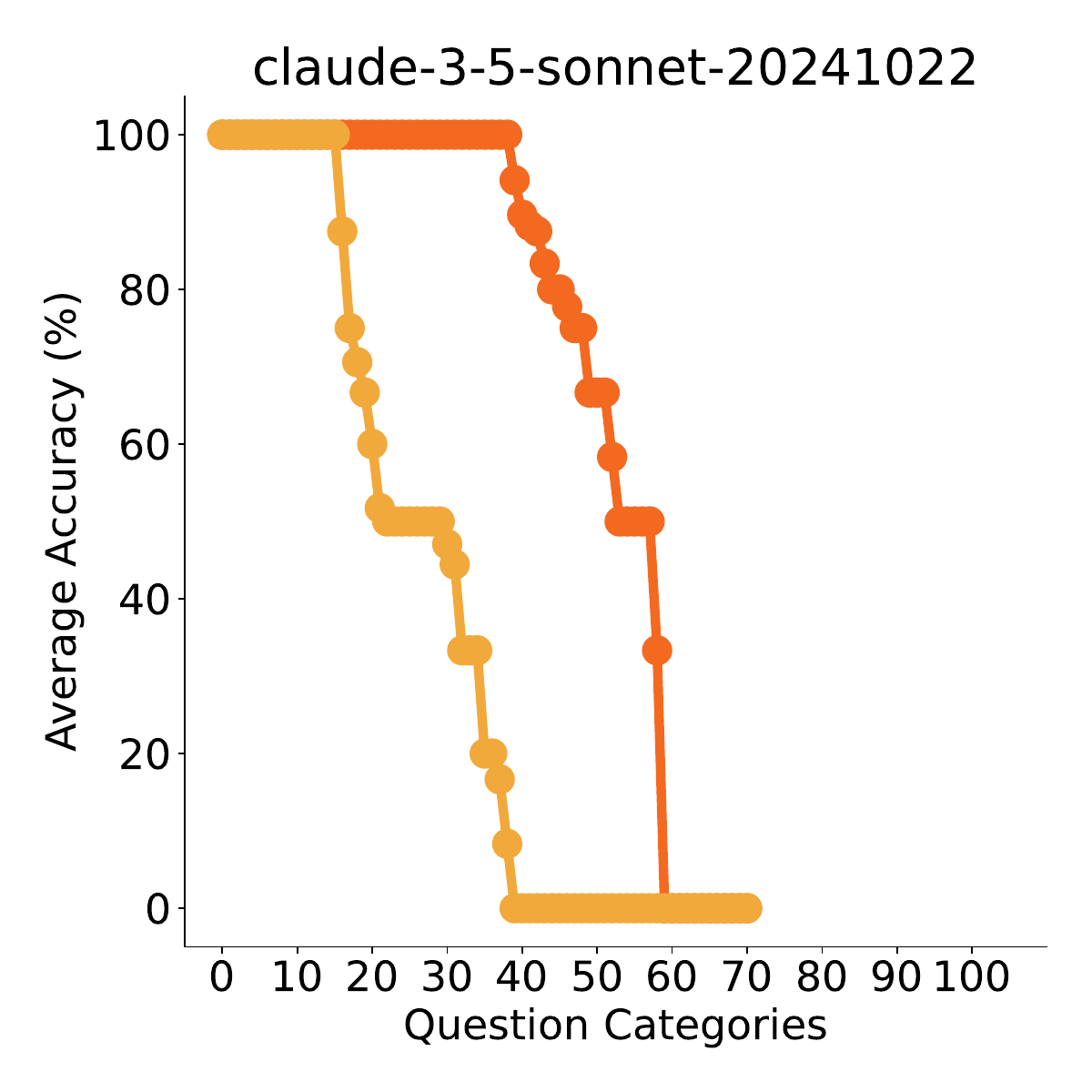}
    \end{subfigure}

    \begin{subfigure}[b]{0.23\textwidth}
        \centering
        \includegraphics[width=\textwidth]{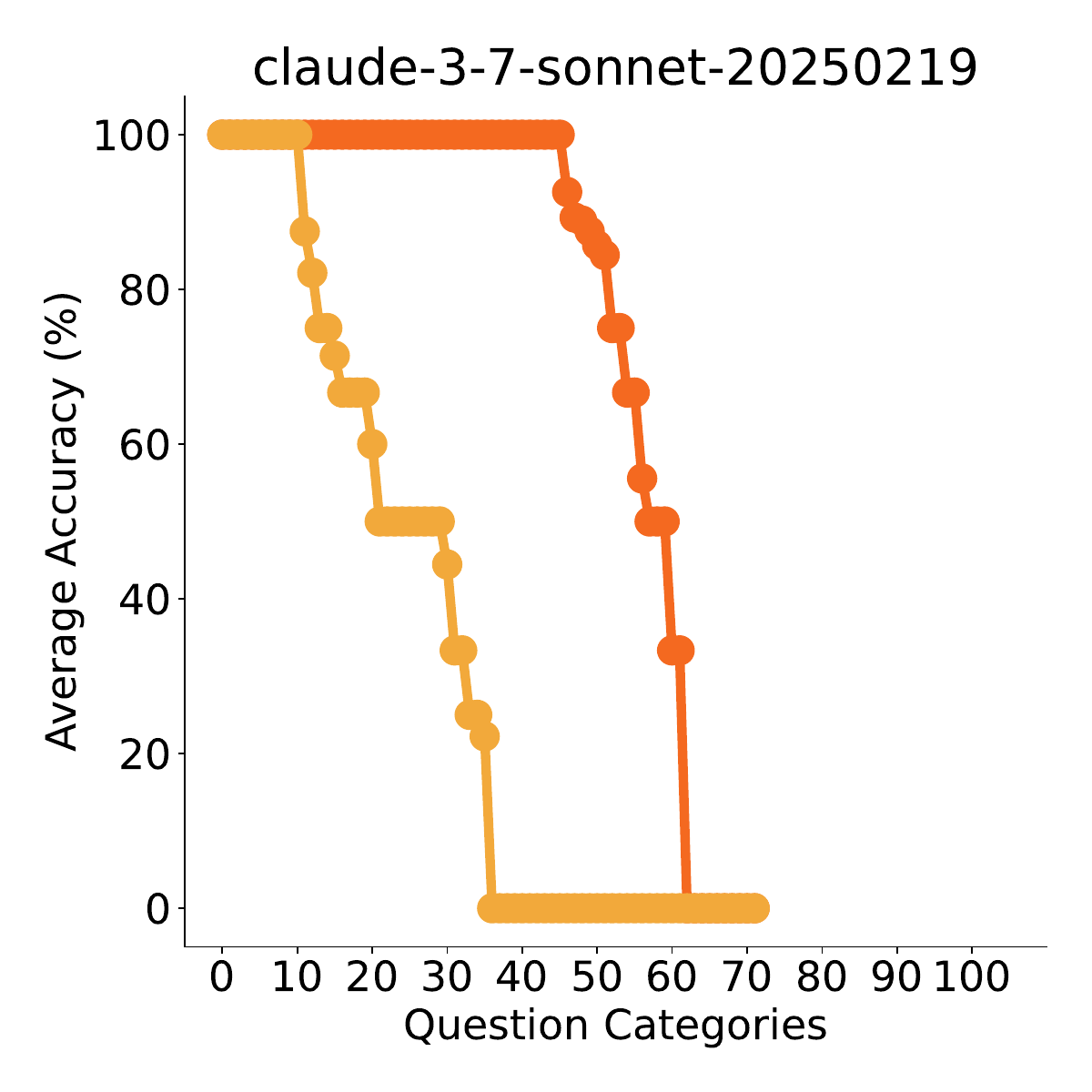}
    \end{subfigure}
    \begin{subfigure}[b]{0.23\textwidth}
        \centering
        \includegraphics[width=\textwidth]{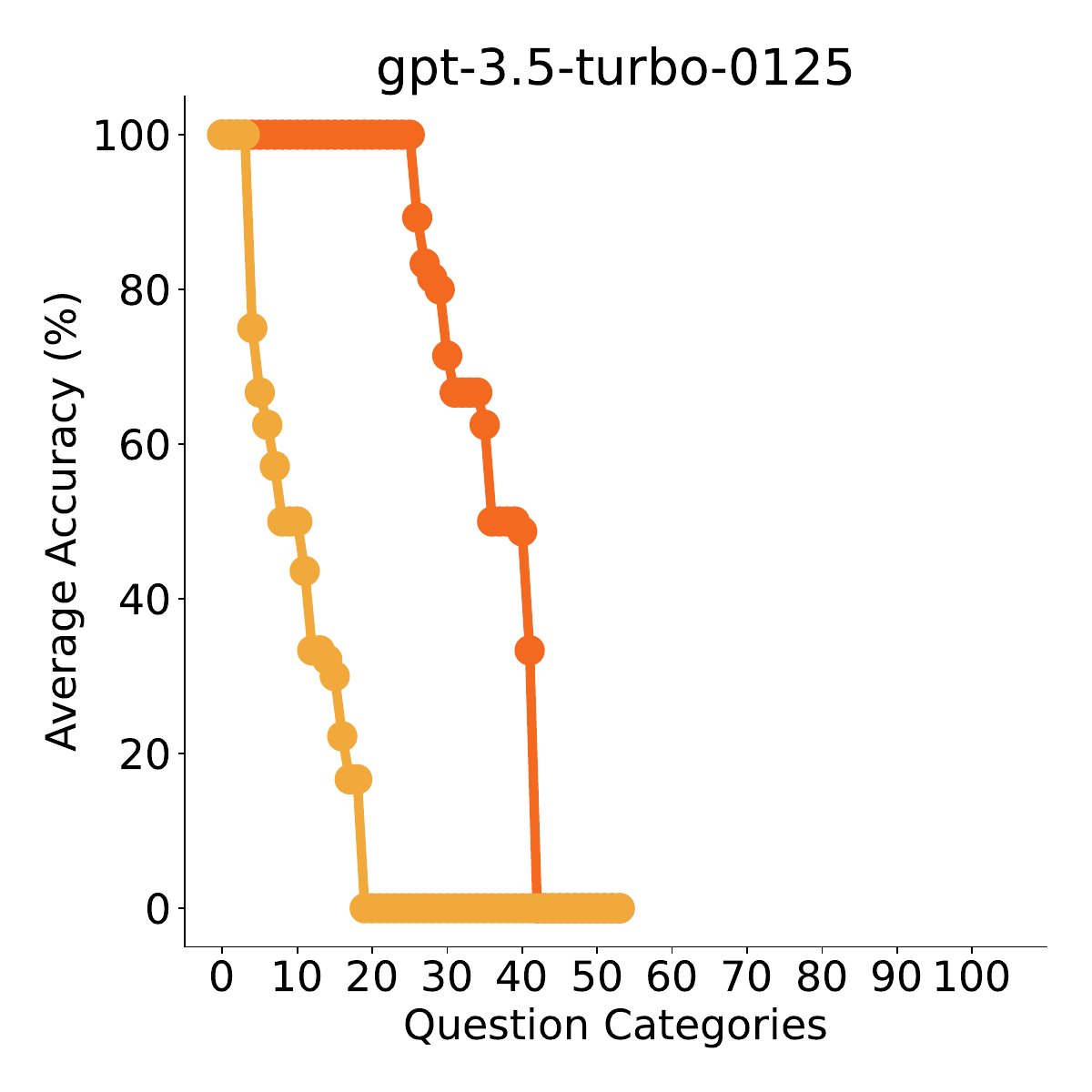}
    \end{subfigure}
    \begin{subfigure}[b]{0.23\textwidth}
        \centering
        \includegraphics[width=\textwidth]{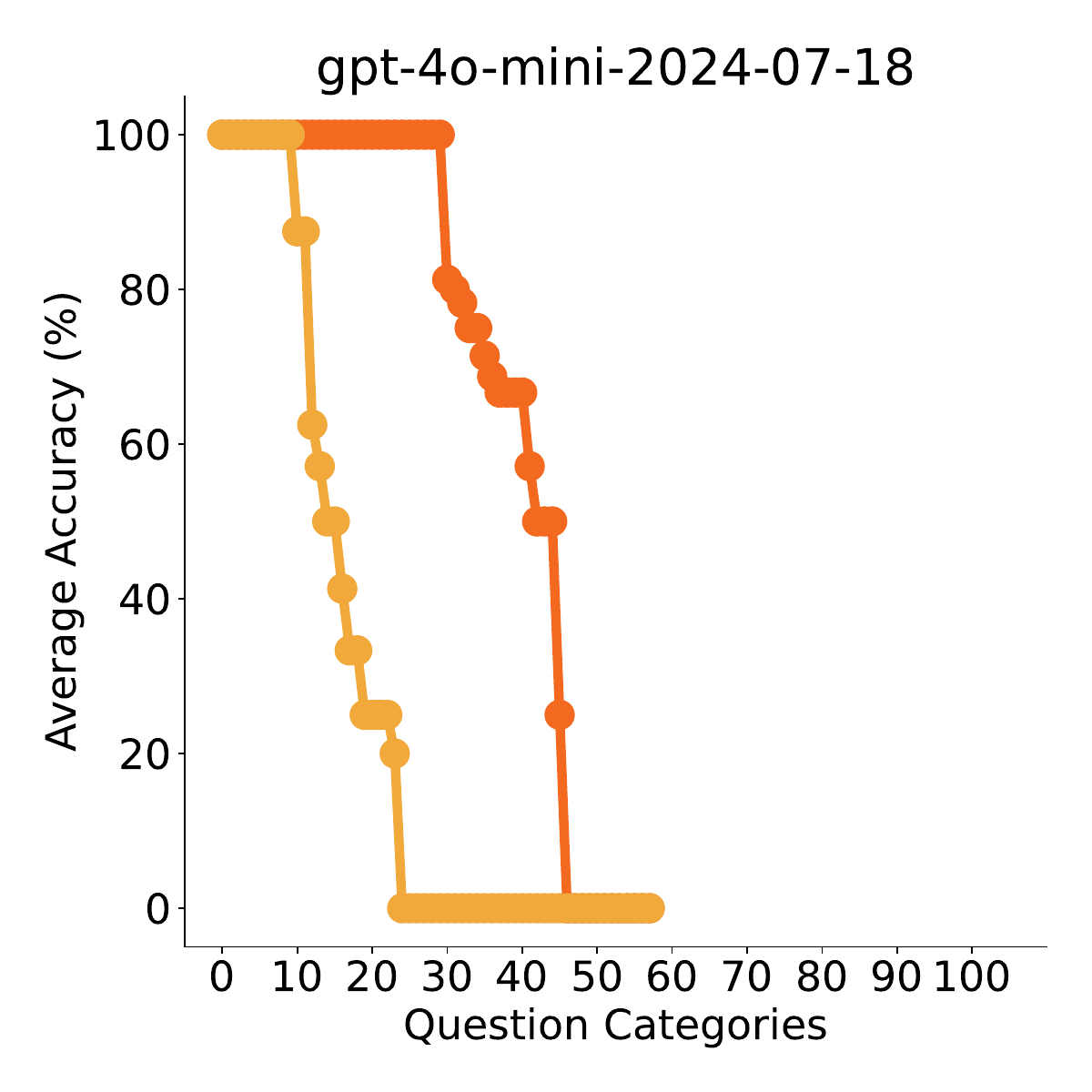}
    \end{subfigure}
    \begin{subfigure}[b]{0.23\textwidth}
        \centering
        \includegraphics[width=\textwidth]{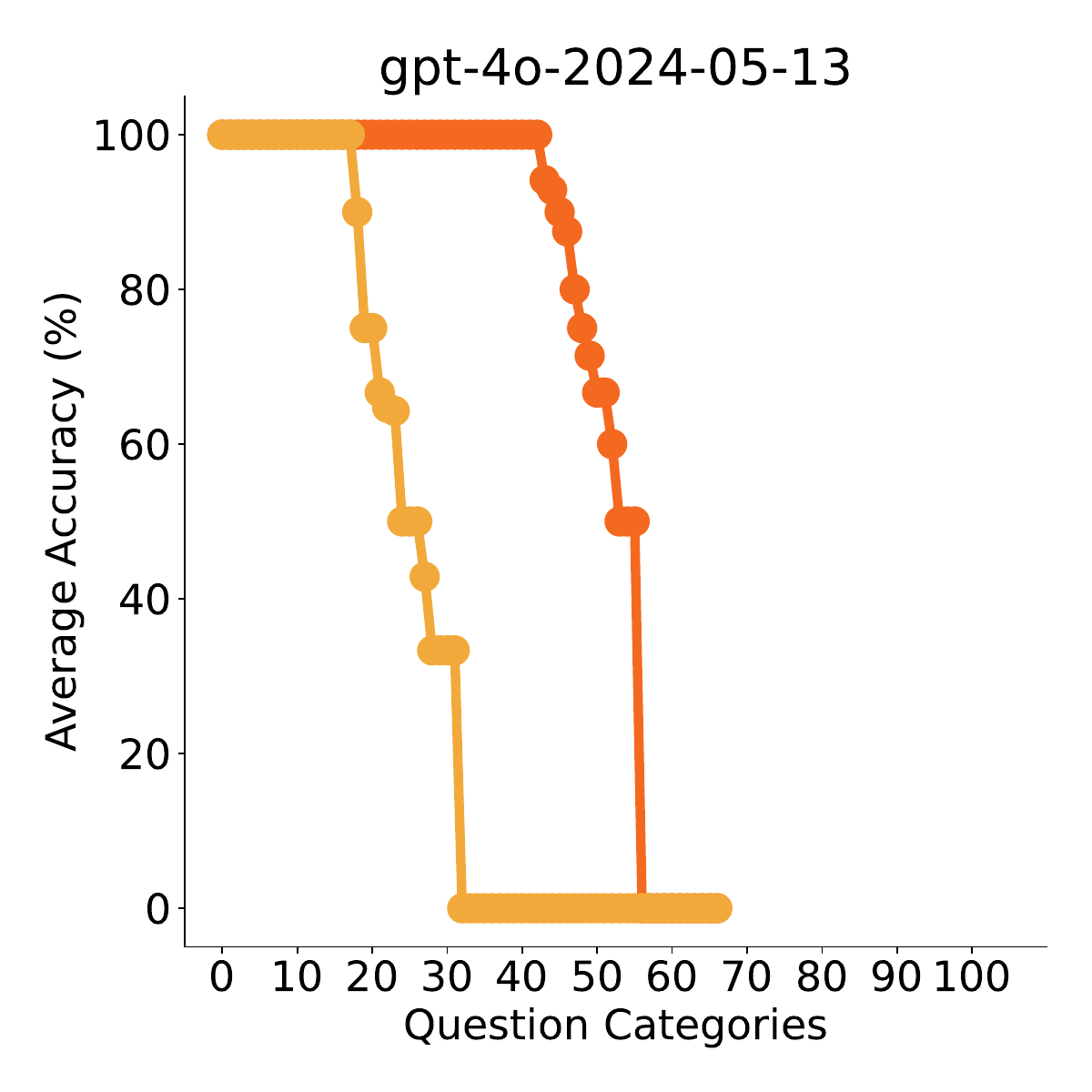}
    \end{subfigure}

    \begin{subfigure}[b]{0.23\textwidth}
        \centering
        \includegraphics[width=\textwidth]{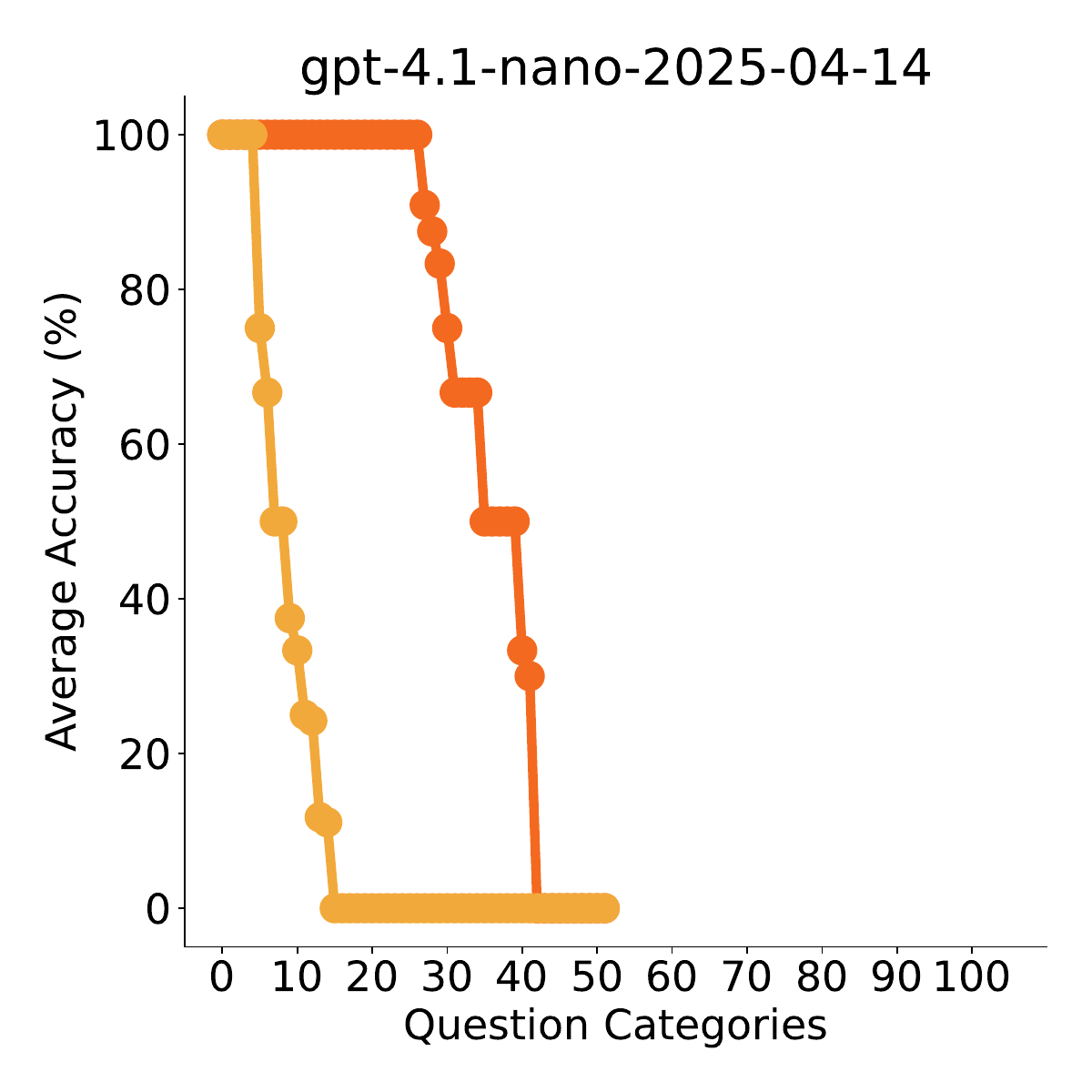}
    \end{subfigure}
    \begin{subfigure}[b]{0.23\textwidth}
        \centering
        \includegraphics[width=\textwidth]{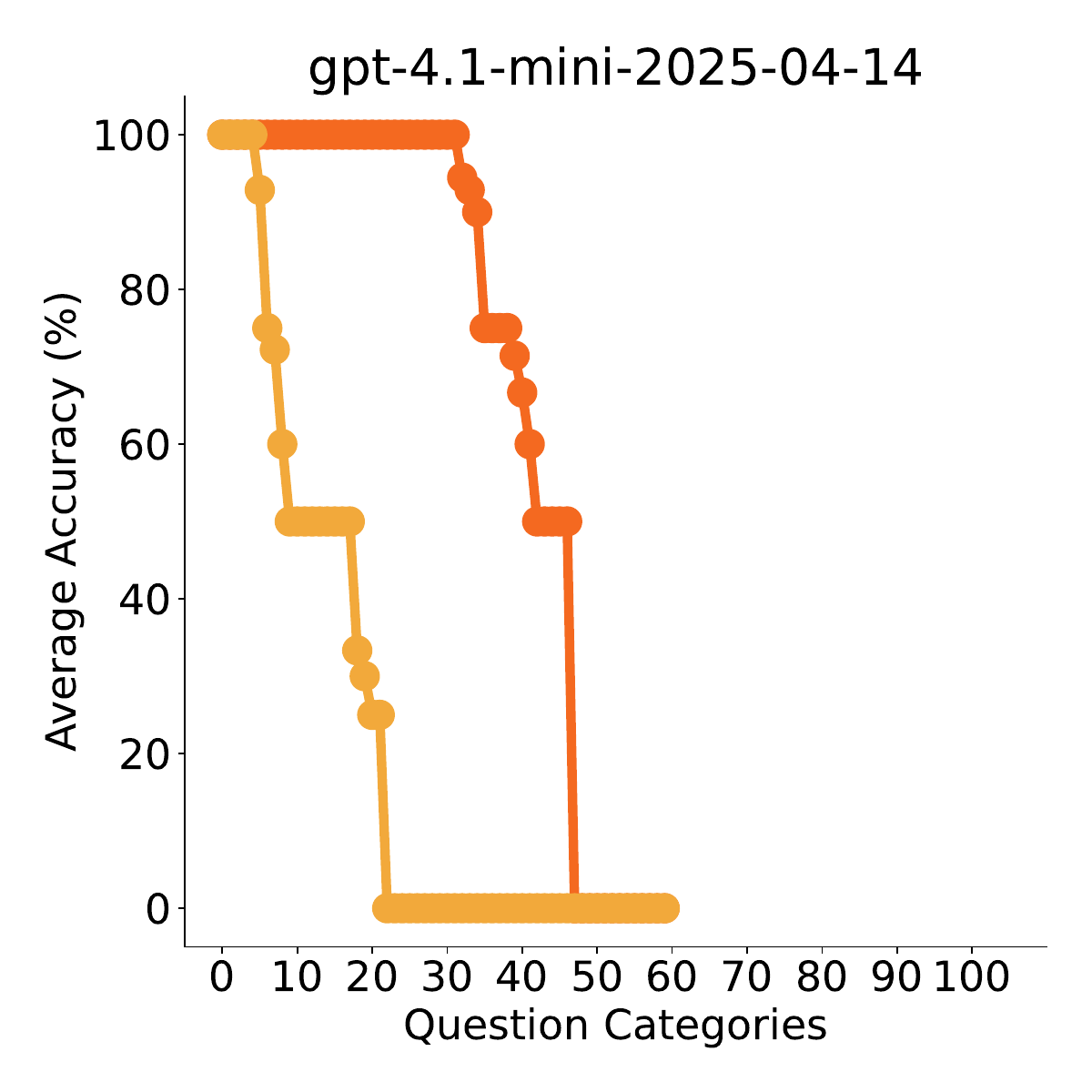}
    \end{subfigure}
    \begin{subfigure}[b]{0.23\textwidth}
        \centering
        \includegraphics[width=\textwidth]{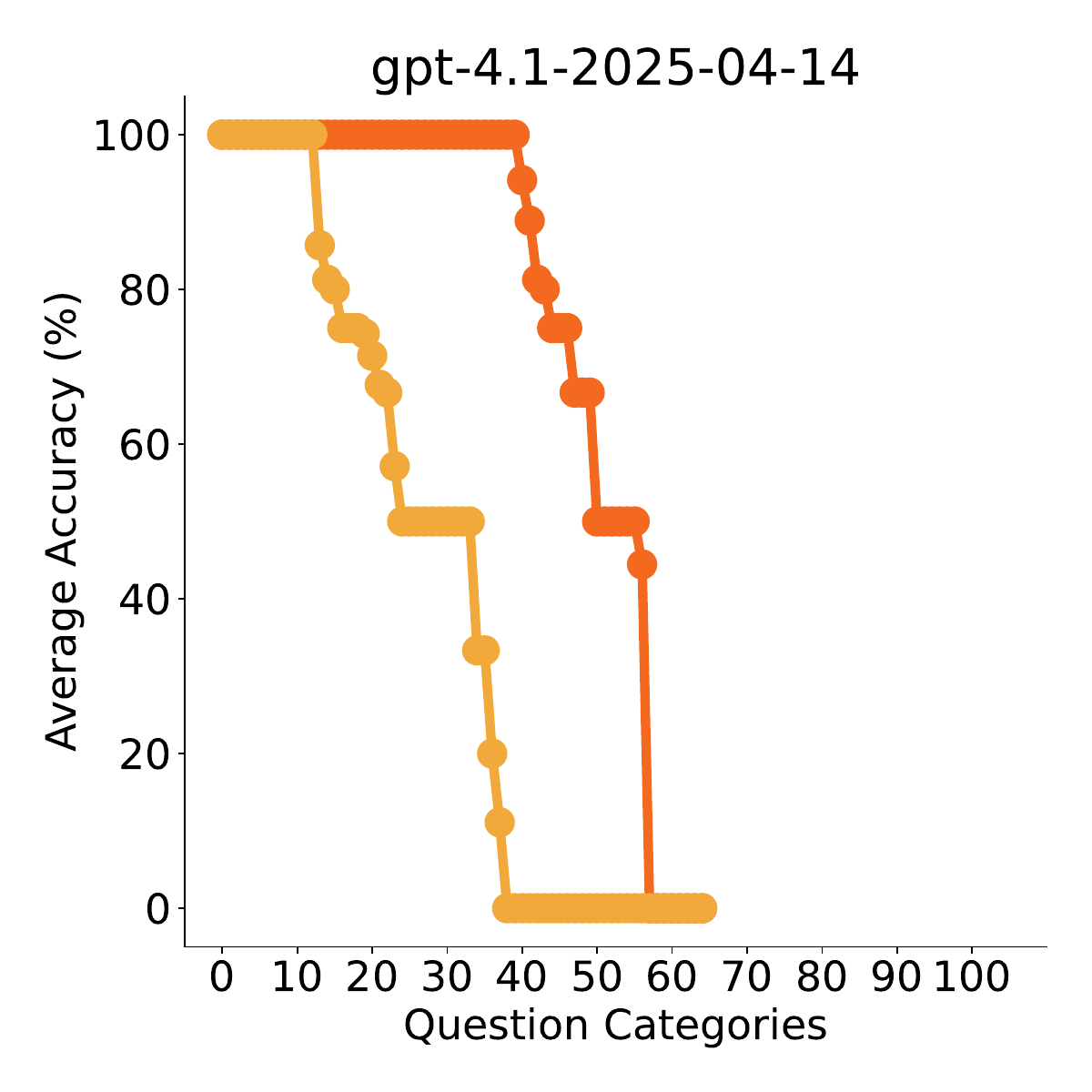}
    \end{subfigure}
    \begin{subfigure}[b]{0.23\textwidth}
        \centering
        \includegraphics[width=\textwidth]{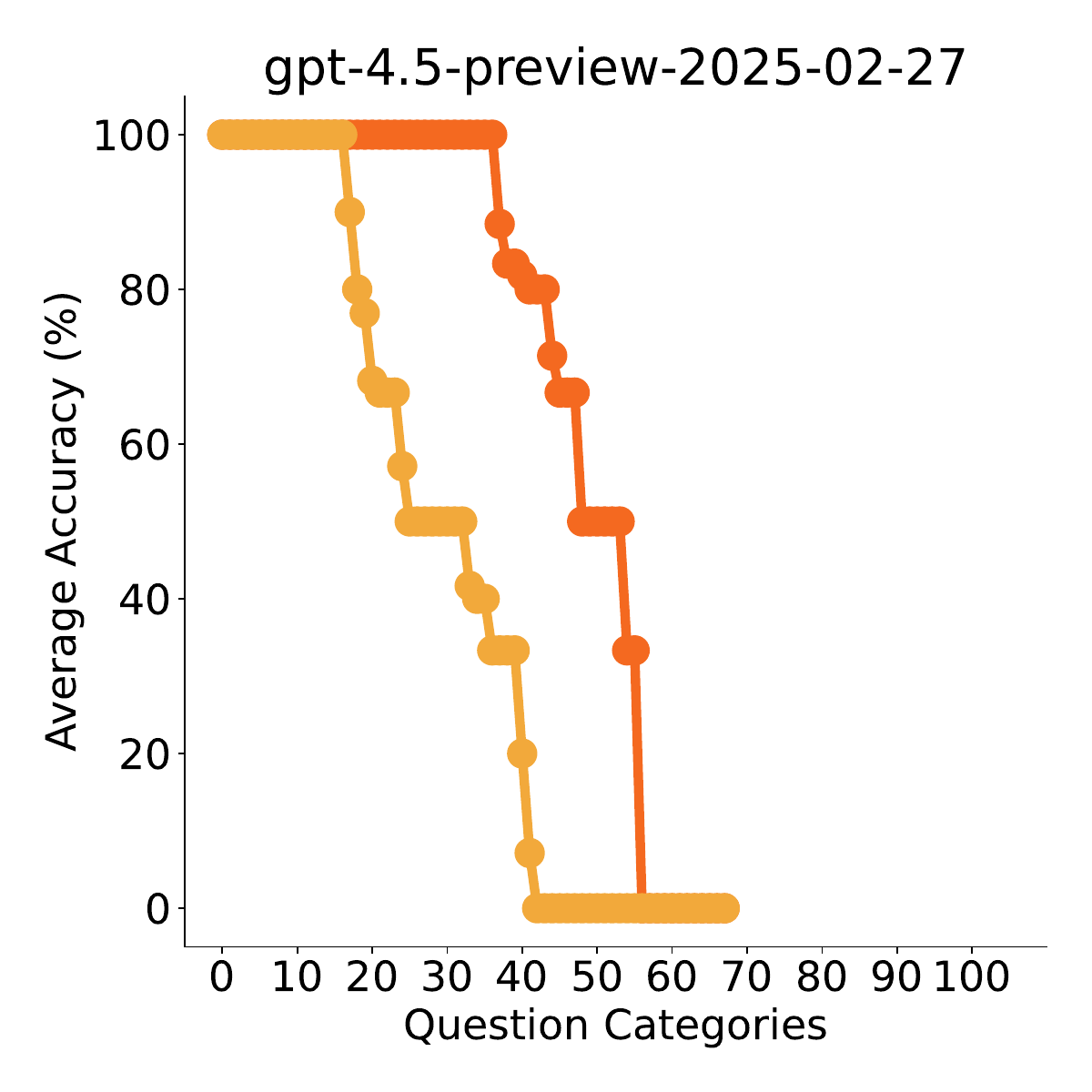}
    \end{subfigure}

    \begin{subfigure}[b]{0.23\textwidth}
        \centering
        \includegraphics[width=\textwidth]{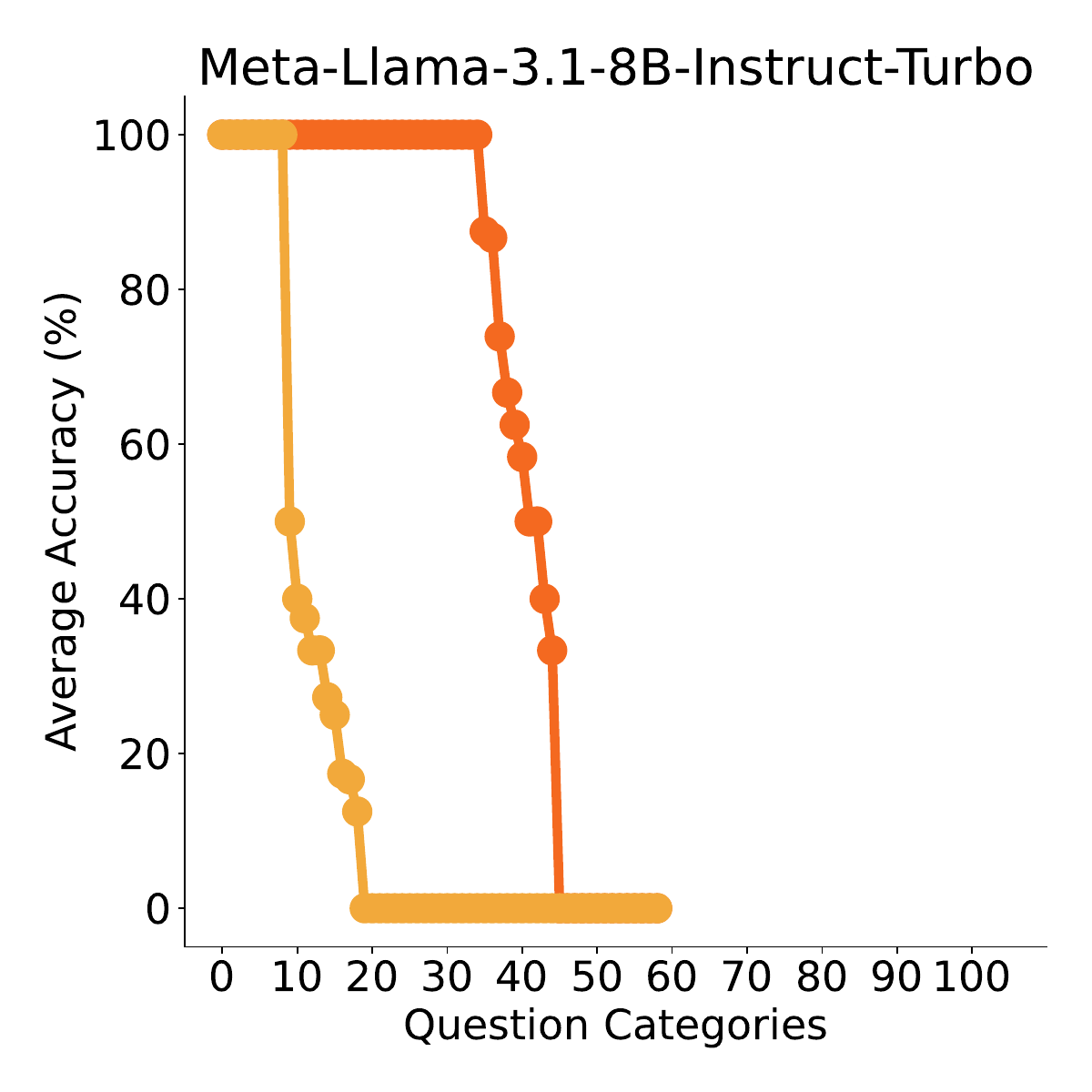}
    \end{subfigure}
    \begin{subfigure}[b]{0.23\textwidth}
        \centering
        \includegraphics[width=\textwidth]{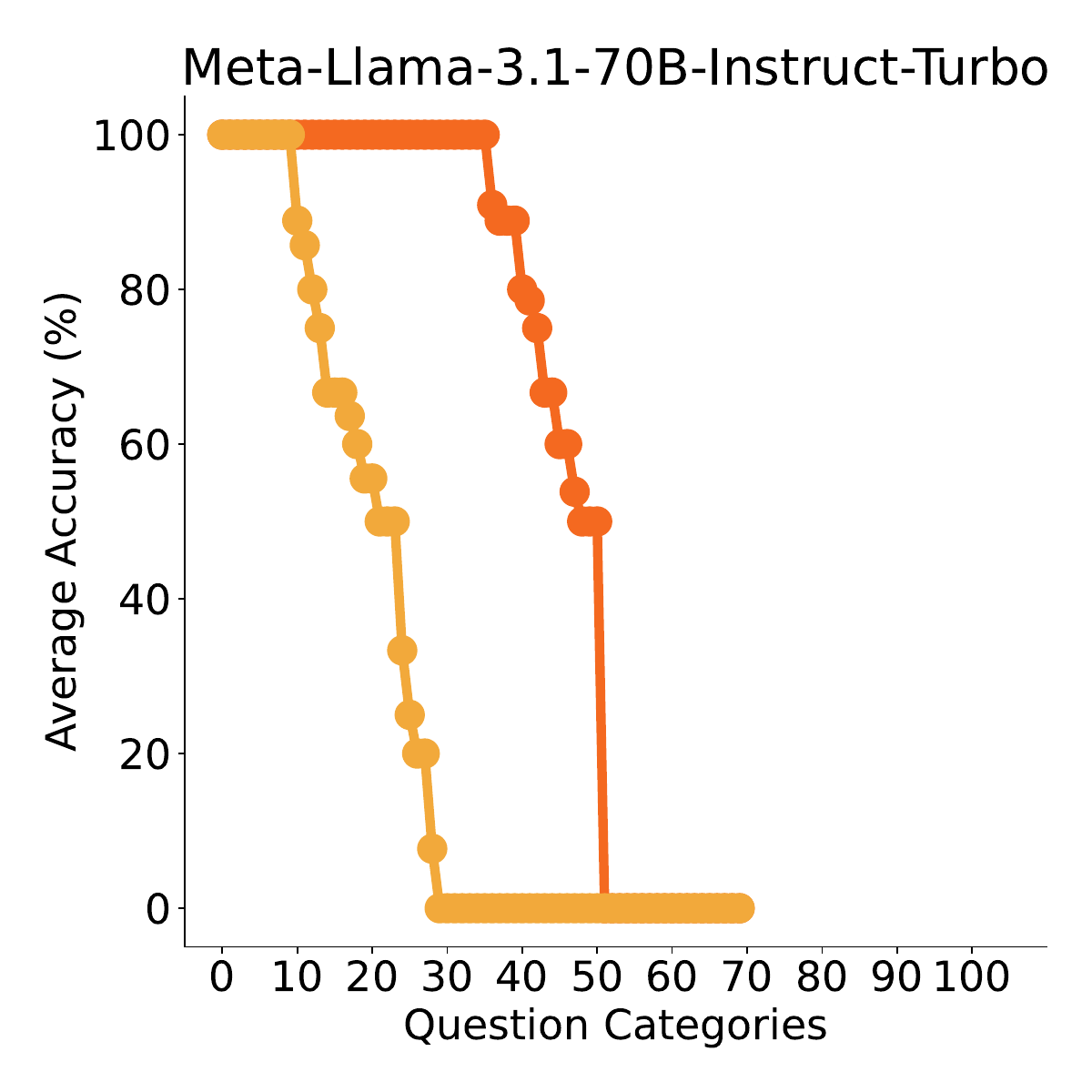}
    \end{subfigure}
    \begin{subfigure}[b]{0.23\textwidth}
        \centering
        \includegraphics[width=\textwidth]{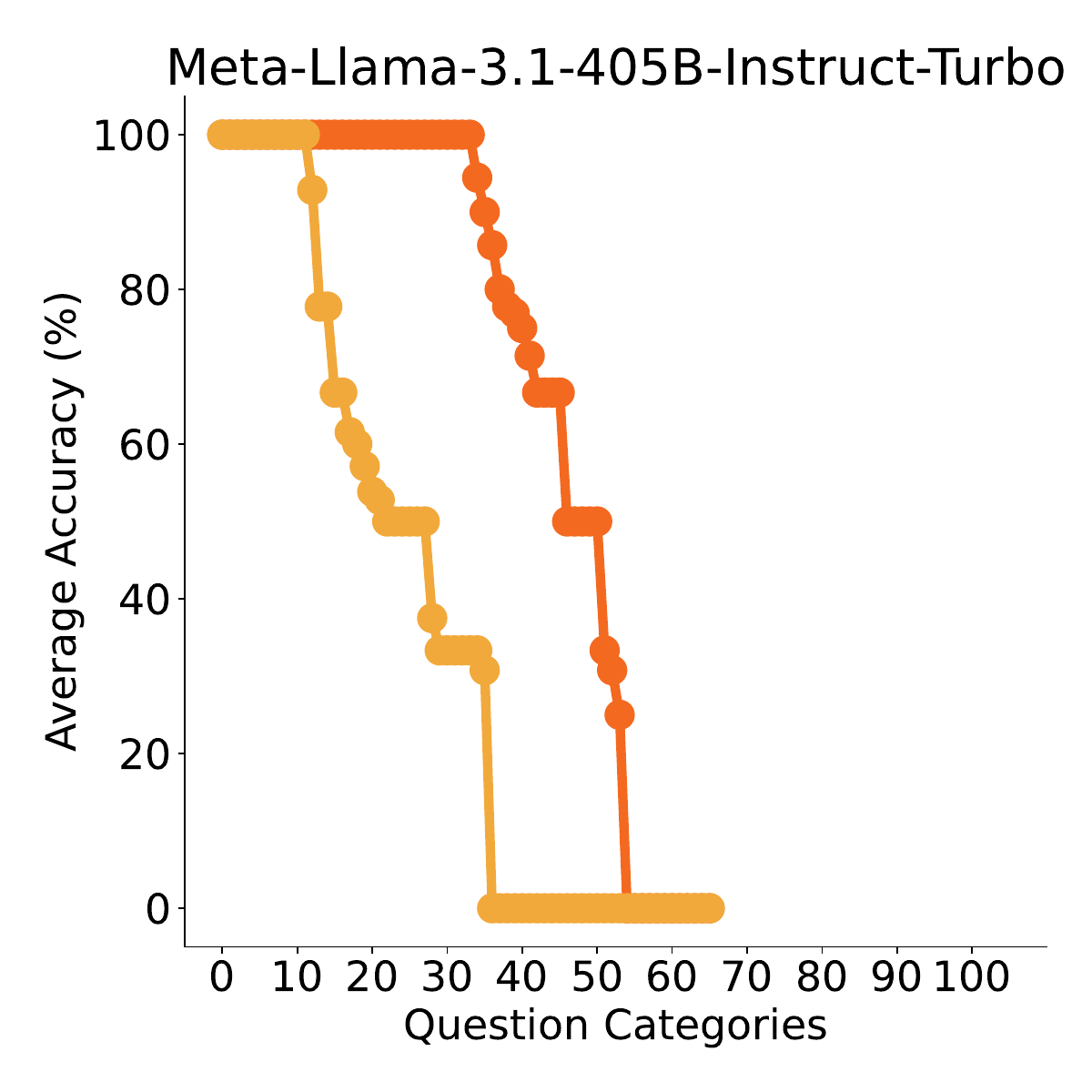}
    \end{subfigure}
    \begin{subfigure}[b]{0.23\textwidth}
        \centering
        \includegraphics[width=\textwidth]{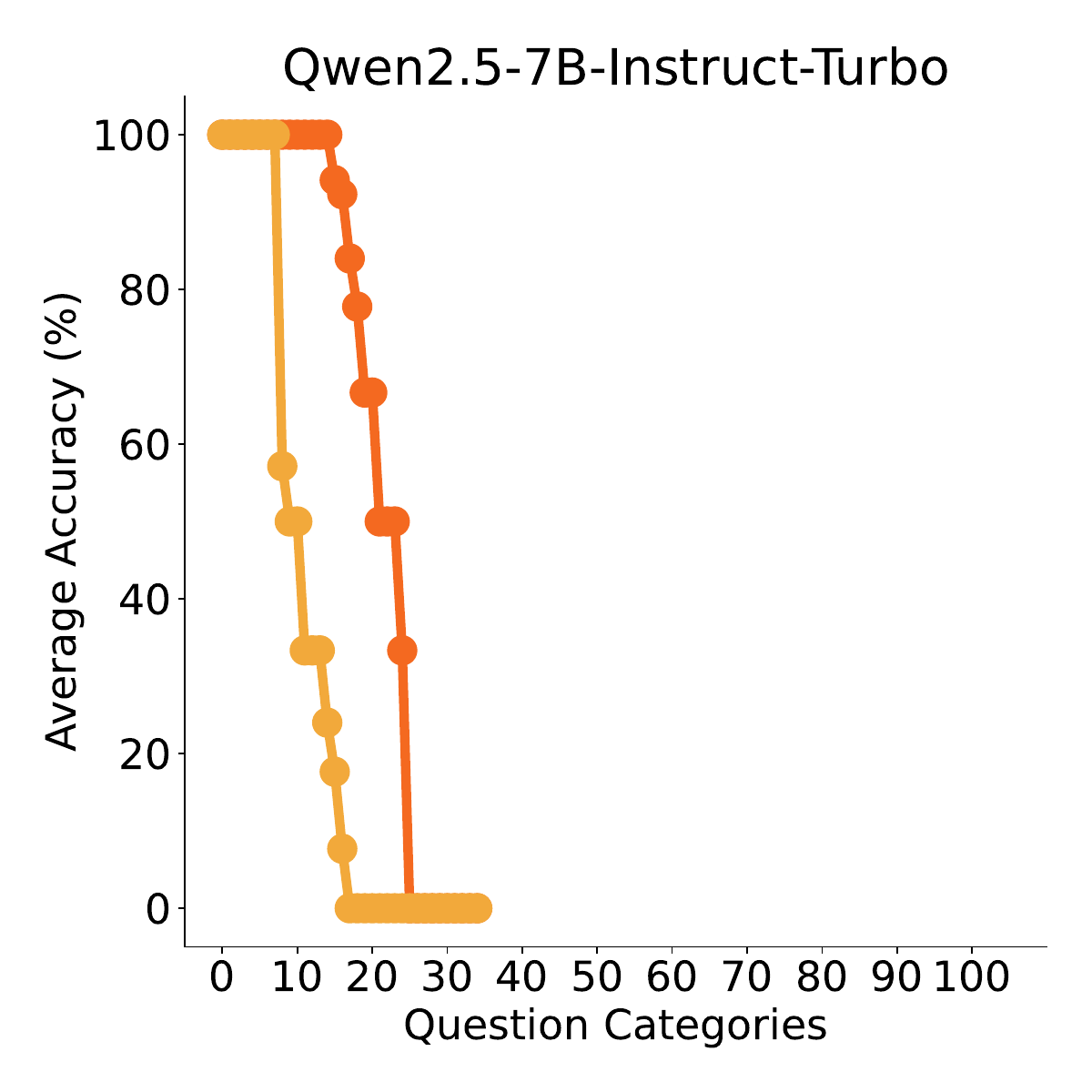}
    \end{subfigure}

    \begin{subfigure}[b]{0.23\textwidth}
        \centering
        \includegraphics[width=\textwidth]{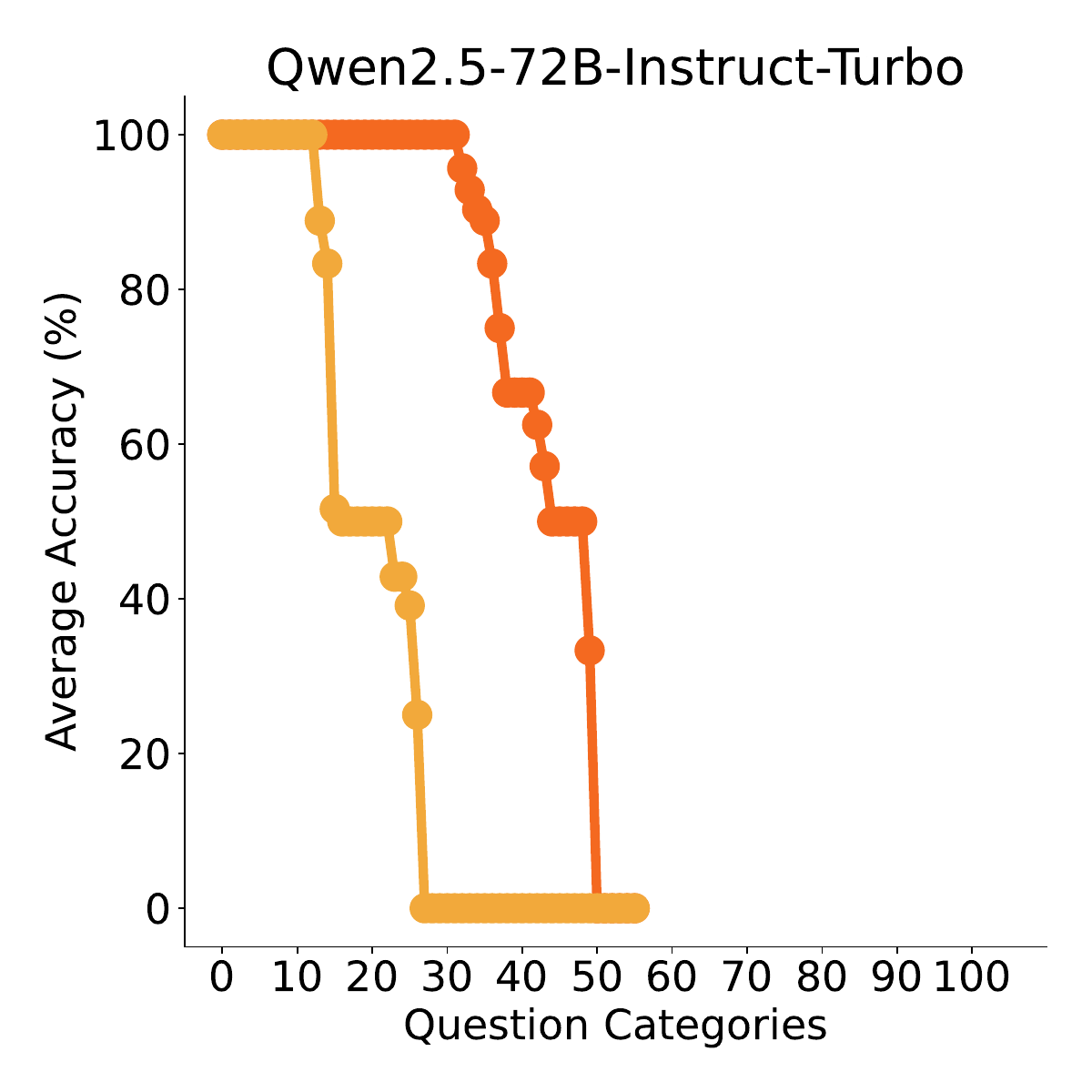}
    \end{subfigure}
    \begin{subfigure}[b]{0.23\textwidth}
        \centering
        \includegraphics[width=\textwidth]{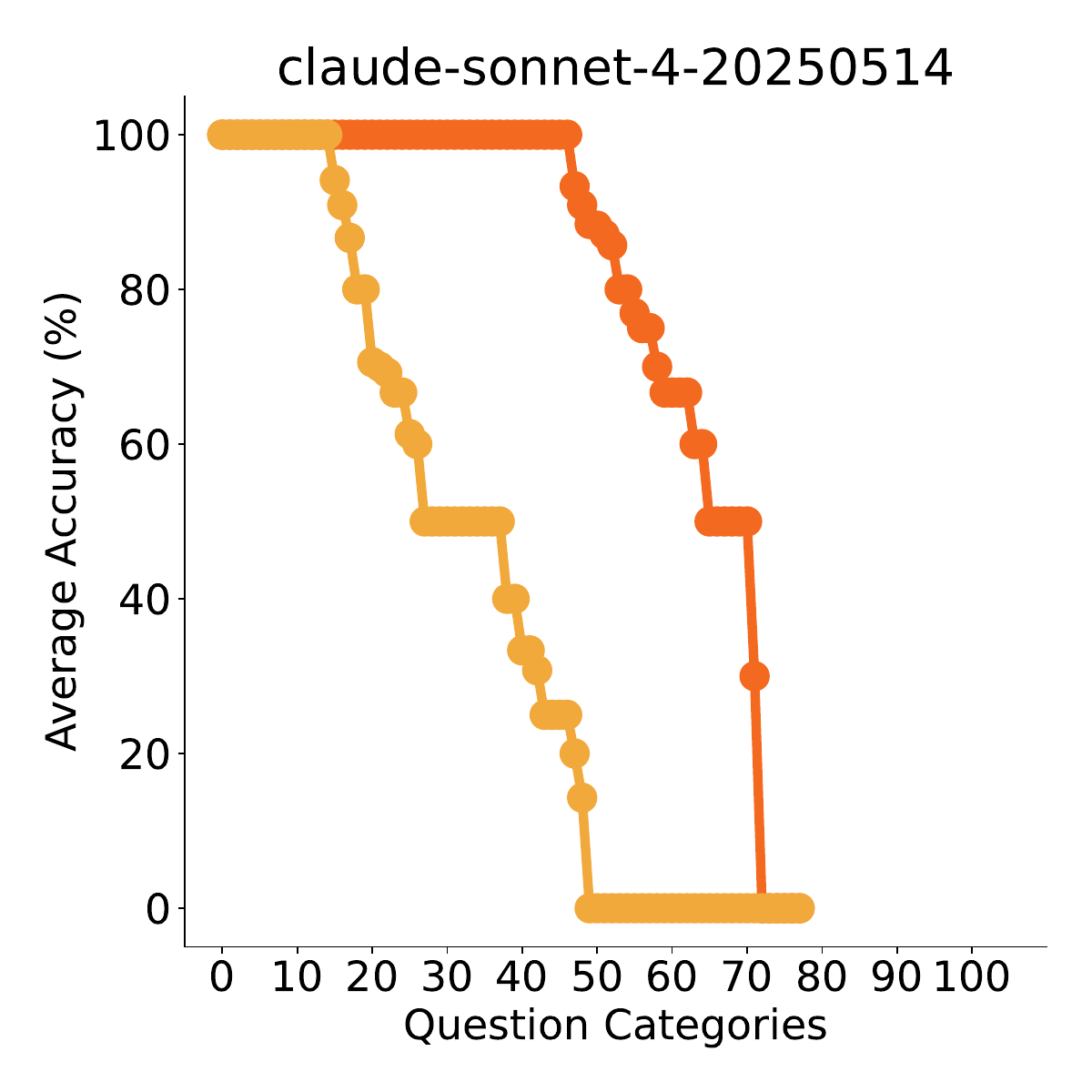}
    \end{subfigure}
    \begin{subfigure}[b]{0.23\textwidth}
        \centering
        \includegraphics[width=\textwidth]{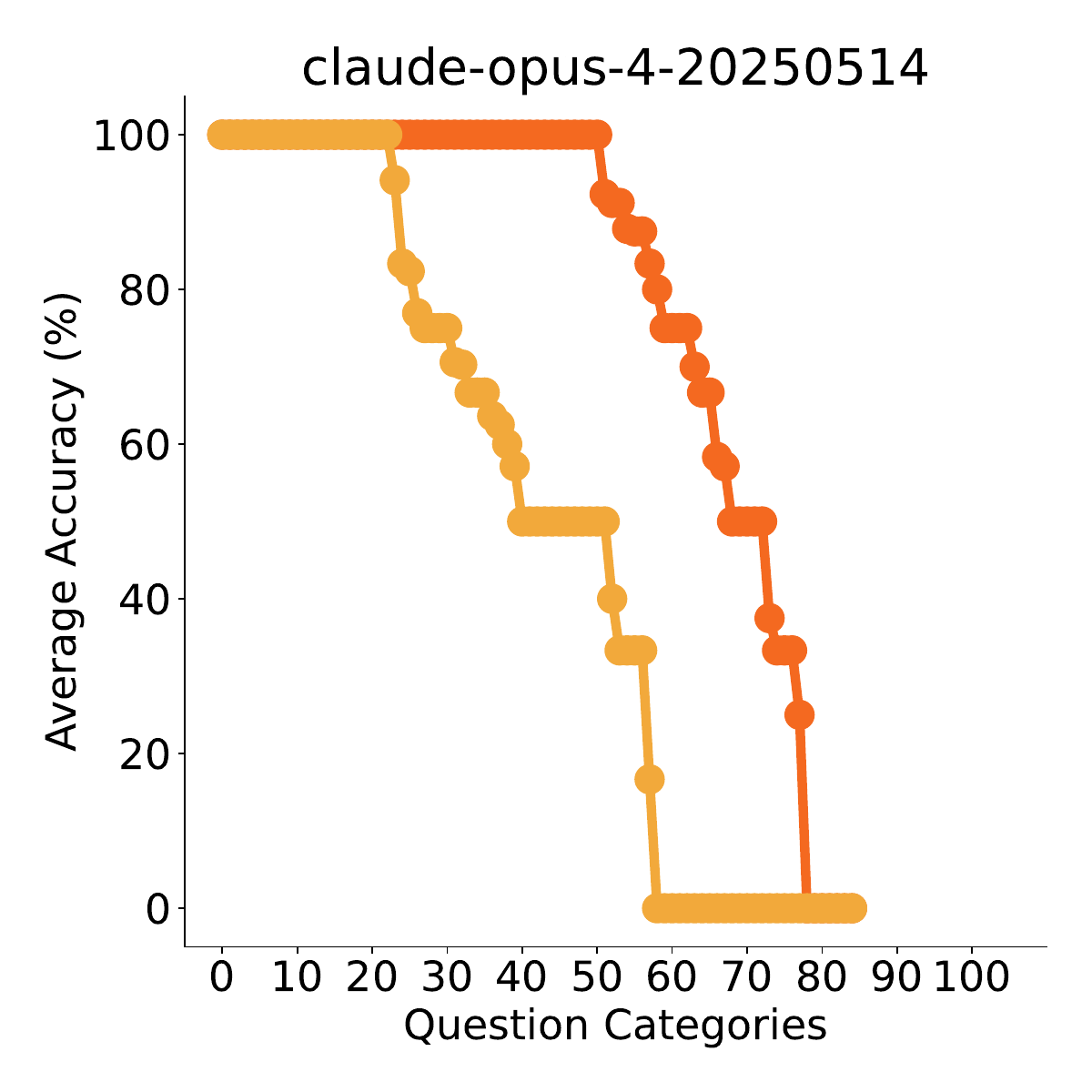}
    \end{subfigure}

    \caption{Per-category performance for all evaluated frontier models. Each subplot shows the distribution of performance across question categories for both CoT and no-CoT reasoning. For many models, we observe that no-CoT performance is zero for most categories. In contrast, for most models (except for Qwen-2.5) their CoT performance maintains partial success for at least half of the categories.}
    \label{fig:all_model_frequencies}
\end{figure}

\end{document}